\documentclass[10pt,twocolumn,letterpaper]{article}

\usepackage[pagenumbers]{cvpr}

\usepackage{cuted} %

\usepackage{currfile} %

\usepackage{caption} %

\usepackage{graphicx}
\usepackage{amsmath,amssymb}
\usepackage{booktabs}
\usepackage{array}
\usepackage{multirow}
\usepackage{makecell}
\usepackage{xspace}
\usepackage{bm}
\usepackage[inline]{enumitem}
\usepackage{xcolor}

\newcommand{\ci}[2]{[#1,\,#2]}
\newcommand{\sdm}{SD3.5-M\xspace}
\newcommand{\sdxl}{SDXL\xspace}
\newcommand{\flux}{FLUX.1-dev\xspace}
\newcommand{\fluxk}{FLUX.2\,[klein]\xspace}
\newcommand{\samt}{SAM\,3\xspace}
\newcommand{\qwen}{Qwen-Image\xspace}
\newcommand{\gdino}{Grounding DINO\xspace}
\newcommand{\owl}{OWLv2\xspace}
\newcommand{\para}[1]{\noindent\textbf{#1}\hspace{0.4em}}
\newcommand{\supref}[1]{supp.~\cref{#1}}

\setlist{nosep,leftmargin=*}

\definecolor{cvprblue}{rgb}{0.21,0.49,0.74}
\usepackage[pagebackref,breaklinks,colorlinks,allcolors=cvprblue]{hyperref}
\hypersetup{pdftitle={Same Layout, Different Count: What Number Words Do Past Ten in Text-to-Image Models},pdfauthor={Liang Yue, Eric A. Vance}}

\newcommand{\NIMAGES}{3{,}088}

\title{Same Layout, Different Count:\\What Number Words Do Past Ten in Text-to-Image Models}

\author{Liang Yue\\
University of Colorado Boulder\\
Boulder, Colorado, USA\\
{\tt\small liyu2195@colorado.edu}
\and
Eric A. Vance\\
University of Colorado Boulder\\
Boulder, Colorado, USA\\
{\tt\small Eric.Vance@colorado.edu}
}

\begin{document}
\maketitle
\begin{strip}
\centering
\begin{minipage}[t]{0.545\linewidth}
\vspace{0pt}\raggedright{\small\textbf{a}}\\[-1pt]
\centering\includegraphics[width=\linewidth]{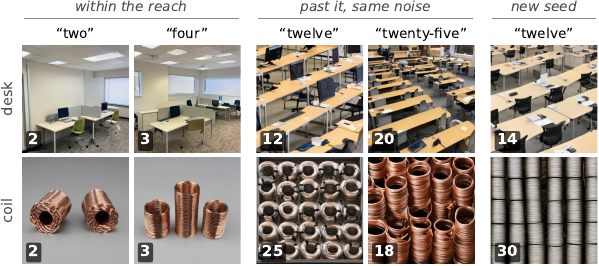}
\end{minipage}\hfill
\begin{minipage}[t]{0.44\linewidth}
\vspace{0pt}\raggedright{\small\textbf{b}}\\[-1pt]
\centering\includegraphics[width=\linewidth]{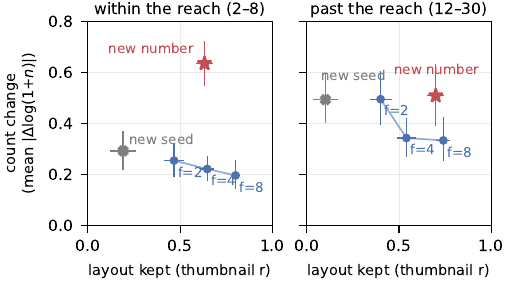}
\end{minipage}
\captionsetup{hypcap=false}
\captionof{figure}{\textbf{Same layout, different count} (\sdm). (a)~From one starting noise per row (people's counts): within the reach, asking for four instead of two adds objects to the same layout; past it, asking for twenty-five instead of twelve keeps the composition of the picture and changes the count, up for the desks and down for the coils; a new seed redraws the picture. Two of the 16 people-counted types, chosen to show both directions. (b)~Over 64 object types, how far the count moves against how much of the layout is kept (\cref{tab:measures}), for a new number from the same noise, a new seed, and a redraw of only the noise finer than $f$ cycles per image. Within the reach (requests 2--8) a new number moves the count far more than any redraw; past it (12--30), a redraw above eight cycles keeps the layout as well as a new number does and moves the count two-thirds as far (0.69 times as far by people's counts of 16 types). Counts by \samt, 95\% intervals over types.}
\captionsetup{hypcap=true}
\label{fig:teaser}\label{fig:bandseed}
\end{strip}

\begin{abstract}
Ask SD3.5-Medium for twelve of an object and then, from the same starting noise, for twenty-five. The picture keeps its layout, but the count changes about as much as it would under a new random seed, and in no consistent direction. We study what number words do once a request passes about ten objects, in five open text-to-image models. Holding the object and the starting noise fixed, we change one thing at a time, and people who never see the request count \NIMAGES{} of the images. Every model follows the number for small crowds, but only up to a point, its reach: SD3.5-M and FLUX.1-dev stop following it soon after ten, while FLUX.2\,[klein] and Qwen-Image follow it loosely up to twenty-five. Beyond SD3.5-M's reach, the coarse part of the starting noise fixes the composition of the picture but not how many objects it holds: redrawing only the noise finer than an eighth of the image keeps the layout as well as a new number does, and already moves the count two-thirds as far. The number word still reaches the denoiser and acts in its first steps, but nothing we tried gives it a clear direction. Shown instead as a noised grid in the starting latent, the count is drawn: people find it in 59\% of images, against 4\% when it is only told. Past ten, an exact-match score barely separates a model that follows the number loosely from one that ignores it, and a relative-error score can even prefer a constant guess of sixteen. A same-noise test with a few requests per object tells them apart.
\end{abstract}

\section{Introduction}
\label{sec:intro}

Ask a text-to-image model for twenty cream pitchers and you will rarely get twenty. Counting is one of the most persistent failures of these models~\cite{kajic2024geckonum,testolin2025visual,t2icountbench2025}, and it worsens with the number: past ten objects, open and closed models are rarely exact~\cite{t2icountbench2025,countloop2025,numbench2026}. That much is known. What is not known is whether the number word still does anything there. A model that follows the number loosely and a model that ignores it are both rarely exact, so an exact-match rate cannot tell them apart (\cref{fig:regimes}). Recent work points to the starting noise, which fixes much of a picture's layout early in sampling~\cite{zhao2025noise,gu2026noise}. Zhao~\etal~\cite{zhao2025noise} find that in a model fine-tuned on synthetic counts, a fixed noise tends to give the same count whatever the prompt asks. If the noise sets the count, what is left for the number word?

We answer with same-noise interventions. Sampling is deterministic, so an object and a starting noise define a picture, and we can change one thing at a time: the number, the seed, the spelling of the number, an unrelated word, or only the fine-scale part of the noise. The test is simple. If the number still steers the crowd, the larger request should give more objects from the same noise, and a new number should move the count further than changes that do not touch it. We run it in five open models, from \sdxl~\cite{podell2024sdxl} to \fluxk~\cite{flux2klein2026} and \qwen~\cite{wu2025qwenimage}, count every image with three detectors, and have \NIMAGES{} of the images counted by people who never see the request.

The answer turns on a split between layout and count (\cref{fig:teaser}). Past the range of requests that \sdm follows, a new number from the same noise keeps the picture's layout, much as Zhao~\etal observed, but not its count: the count changes about as much as under a new random seed, and in no consistent direction. Redrawing only the noise finer than an eighth of the image keeps the layout just as well and already moves a large crowd's count two-thirds as far. Past the reach, the coarse part of the starting noise fixes the composition of the picture, but not how many objects it holds. Four findings build this picture.

\para{Each model follows the number only so far (\cref{sec:reach}).} Every model follows small numbers: from two to eight objects, a larger request gives more objects in 80--99\% of the same-noise pairs whose counts differ. Past ten the models part ways. \sdm and \flux stop following the number soon after ten, while \fluxk and \qwen follow it up to twenty-five about as loosely as a noisy estimator would. We call the range a model follows its \emph{reach}.

\para{Past the reach, the count changes beneath a kept layout (\cref{sec:layout}).} In \sdm, fine-scale noise moves crowds of every size. Within the reach, a new number moves the count further than a redraw of the fine noise, and toward the request. Past it, the redraw moves the count two-thirds as far as the number, by detectors and by people alike, and the number has no direction.

\para{The number arrives, without a direction (\cref{sec:when}).} The large number can be read off \sdm's conditioning in order, and in all three models we timed, number words act in the first denoising steps. Nothing we tried gives the large number a clear direction, from amplifying it to fine-tuning on counted crowds.

\para{A large count can be shown instead (\cref{sec:show}).} Started from a noised grid of its own object, \sdm draws the requested count in 59\% of images by people's counts, against 4\% from the prompt. A planted count survives a conflicting number word, while a different object word repaints the grid. A layout written into the noise can carry the count, as Zhao~\etal's count-aware noise prior does; the random noise's own layout does not.

These results also bear on measurement: past ten, exact-match and relative-error scores barely tell following from ignoring, and a same-noise order can.

\para{Contributions.} (1)~A same-noise protocol, with a fine-scale redraw of the noise and a reference read through people's own counting noise, and people's counts of \NIMAGES{} generated images, which we will release. (2)~The reach of five open models. (3)~The finding that past \sdm's reach the layout and the count come apart: a large crowd's count changes with noise finer than the layout, the number word changes it about as much as a new seed and without a direction, and a count supplied as a layout is drawn far more often. Unlike Zhao~\etal~\cite{zhao2025noise}, who fine-tune a model on synthetic counts, we test off-the-shelf models, separate the noise's layout from its fine detail, and compare the number with controls that leave its value alone.

\section{Related Work}
\label{sec:related}

\para{Counting past ten.}
Counting is a standard category of compositional benchmarks~\cite{ghosh2023geneval,huang2025t2icompbenchpp,kamath2025geneval2}, and dedicated benchmarks show scores falling with the number for open, closed and recent large models~\cite{kajic2024geckonum,t2icountbench2025,numbench2026}. Prompt refinement does not help~\cite{t2icountbench2025}, and a benchmark that grades counts by relative error finds grid layouts best~\cite{numbench2026}. Generated counts up to ten behave like estimates whose errors grow with the number~\cite{testolin2025visual}, and numeral features in FLUX.1 [schnell] order more against fewer objects without sharp values~\cite{mazur2026diffract}. We add a test that tells following from ignoring.

\para{The noise and the layout.}
The starting noise fixes much of the layout early in sampling~\cite{ban2025crystalball,seedsofstructure2025,mao2024lottery,dahary2025decisive}, its low frequencies carry coarse structure~\cite{everaert2024signalleak,wu2024freeinit}, and a seed tends to keep where objects go and how large they are across prompts for the same object~\cite{xu2025goodseed}. Zhao~\etal~\cite{zhao2025noise} find that in \flux fine-tuned on synthetic counts a fixed noise tends to give the same count whatever the prompt asks, that a diffusion classifier built from the model separates counts only coarsely, and that fine-tuning on counted images helps mainly below ten; their count-aware noise prior raises exact counts from 20\% to 85\% on a synthetic set of up to fifty objects. Gu~\etal~\cite{gu2026noise} adjust the noise and the early cross-attention at test time. We hold the noise fixed in off-the-shelf models and measure what the word does on top of it, with a new seed, a respelling, an unrelated edit and a fine-scale redraw as controls, since swapping one word under a fixed noise can reshuffle a picture~\cite{hertz2023prompt}. Our results qualify theirs for these models: the fixed noise keeps the layout, but past the reach it does not keep the count.

\para{When the word acts.}
Text conditioning acts mostly early in sampling~\cite{balaji2022ediffi,patashnik2023mixing,liu2025tgate,yi2024mechanism}, where coarse decisions are made~\cite{biroli2024dynamical,li2024critical,sclocchi2025phase}, and the number of instances is largely fixed early~\cite{countcluster2025}. Steering early along the difference from a prompt without the count improves counts up to ten~\cite{sepehri2026athena}, and attention steering~\cite{boo2025countsteer} and counting guidance~\cite{kang2025countingguidance} use signals of whether a count is right. Prompt switching at chosen steps traces when a concept forms~\cite{gorgun2025temporal}; we time the number word against a respelling. Text encoders represent numbers in order~\cite{wallace2019numeracy}, though CLIP counts poorly~\cite{paiss2023clipcount}.

\para{Supplying the count.}
Past ten, the methods that help hand the model a layout: planned boxes~\cite{lian2024lmd,li2023gligen,xiang2025instanceassemble}, corrected layouts~\cite{binyamin2025makeitcount,countloop2025}, composite or noise starts~\cite{meng2022sdedit,composeanything2025,zhao2025noise,gu2026noise} or dense-layout training~\cite{ngo2026dense}. Count tokens can be optimized~\cite{zafar2026ioco,yolocount2025,sun2026quota} and models fine-tuned on verified counts or rewards~\cite{li2025seeds,mosaic2026,liu2025flowgrpo,zheng2025diffusionnft}.

\section{Same-Noise Tests}
\label{sec:setup}

\begin{figure}[t]
\centering
\includegraphics[width=\linewidth]{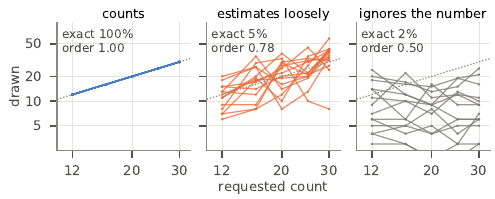}
\caption{\textbf{Why exact rates are not enough} (simulated; one line per starting noise, asked for 12--30 objects). A model that counts is always exact. One that estimates the request loosely (log-normal scatter of 0.4) and one that ignores it are both rarely exact, but only the first orders its crowds (\cref{eq:order}).}
\label{fig:regimes}
\end{figure}

\begin{figure*}[t]
\centering
\includegraphics[width=0.84\linewidth]{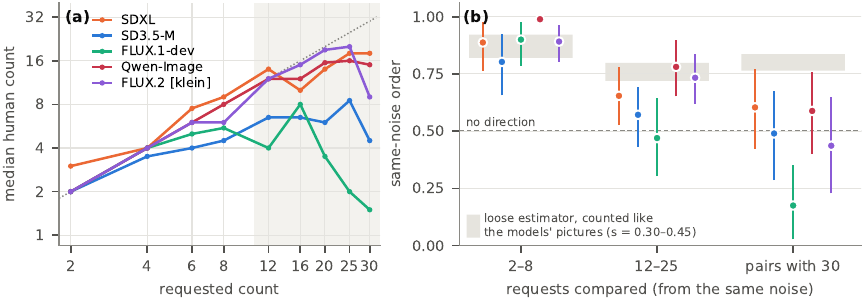}
\caption{\textbf{How far each model follows the number} (people's counts; the same 16 object types for every model, and within a model the same starting noise for every request). (a) Median count by request; dotted: count = request; shaded: past ten. (b) Same-noise order (\cref{eq:order}) for requests of 2--8, for 12--25, and for the pairs that include thirty, with 95\% intervals over types; grey: a loose estimator read through people's counting noise ($s$ = 0.30--0.45); dashed: no direction.}
\label{fig:reach}\label{fig:ceiling}
\end{figure*}

\begin{table}[t]
\centering
\footnotesize
\setlength{\tabcolsep}{3pt}
\begin{tabular}{@{}>{\raggedright\arraybackslash}p{0.17\linewidth}>{\raggedright\arraybackslash}p{0.79\linewidth}@{}}
\toprule
measure & what it is, and how it reads \\
\midrule
order & share of differing same-noise pairs in which the larger request gives more objects: 1 always, .5 no direction; a loose estimator .72--.80 at 12--25 \\
change & mean $|\Delta\log(1{+}n)|$ between two images from the same noise; a picture counted twice by people: .12 \\
layout kept & correlation of $24{\times}24$ thumbnails; past \sdm's reach, a new seed .10, a new number .70 \\
effective scatter & the loose estimator's $s$ that orders a range as the model does; the reach: at most 0.45 \\
\bottomrule
\end{tabular}
\caption{\textbf{The measures at a glance.}}
\label{tab:measures}
\end{table}

Every comparison within a model keeps the object and the starting noise and changes exactly one thing.

\para{Models and prompts.}
We test \sdm~\cite{esser2024sd3,sd35} (a rectified-flow transformer~\cite{liu2023rectified,lipman2023flow} with CLIP and T5 encoders, at $768^2$), \sdxl~\cite{podell2024sdxl} (CLIP only, $1024^2$), \flux~\cite{flux2024} (T5 and a pooled CLIP vector, $768^2$), \fluxk base 4B~\cite{flux2klein2026} (the language model Qwen3-4B~\cite{qwen3} as encoder, $1024^2$) and \qwen~\cite{wu2025qwenimage} (20B, with the vision-language model Qwen2.5-VL-7B~\cite{bai2025qwen25vl} as encoder, $1024^2$, transformer in 4-bit), each with its own sampler and guidance~\cite{ho2022cfg} at 28 to 50 steps (\supref{sec:supp_tests}). The prompt is ``a photo of $c$ $X$s'' with the number $c$ in words, and $X$ is one of 64 everyday object types drawn from the LVIS names~\cite{gupta2019lvis} without looking at any image. A type and a seed fix the starting noise, so the images of all requests for a type share it. People counted requests from 2 to 30; \sdm's tests on 64 types ask for 2, 3, 4, 6 and 8 and for nine requests from 12 to 30.

\para{Counting.}
People on CloudResearch Connect counted \NIMAGES{} generated images in eight studies, clicking on each object without seeing the request or the model (\supref{sec:supp_human}). Two people count each image in separate rounds, and a third decides where they differ. Composites of known count, mixed into the tasks, were counted exactly in 83--100\% of cases. Generated crowds are harder: the first two counts agree on 50--82\% of pictures. Yet two final counts of the same crowd of 12 to 30, made in different studies, differ by a mean $|\Delta\log(1{+}n)|$ of only 0.12 (\sdm, 48 pictures; \supref{sec:supp_noise}), so changes much larger than that are changes in the crowd, not in the counting. The annotators were paid crowdworkers (\$0.10 per image, about \$12--38 per hour) who counted objects in generated images. The tasks recorded counts of objects, not information about the workers, so we treated them as annotation rather than research on human subjects and did not submit them for IRB review; for the same reason, no consent was sought beyond each worker's acceptance of the task on the platform. Every image is also counted by \samt~\cite{carion2026sam3}, \gdino~\cite{liu2024gdino} and \owl~\cite{minderer2023owlv2}. Detectors miss many objects in dense crowds, so we compare models by people's counts and run the tests within a model on \samt, checked against the other two detectors and against people. Unless we say otherwise, a test within \sdm is counted by \samt on all 64 types. Brackets give 95\% bootstrap intervals over object types.

\para{Does the crowd follow the request?}
Let $n_{js}(c)$ be the count drawn from the noise of type $j$ and seed $s$ when the prompt asks for $c$. Over pairs of requests $c<c'$ in a range,
\begin{equation}
\text{order} = \Pr\big[\,n_{js}(c') > n_{js}(c) \;\big|\; n_{js}(c')\neq n_{js}(c)\,\big],
\label{eq:order}
\end{equation}
the share of same-noise pairs whose counts differ in which the larger request gives more objects ($(1{+}\gamma)/2$ for Goodman and Kruskal's $\gamma$~\cite{goodman1954measures}). Because both images share the noise, the order ignores whatever the noise does to every request alike, such as a type that is always drawn large. \Cref{fig:regimes} shows why it is needed: a loose estimator and a model that ignores the number are both rarely exact, and the order tells them apart.

\para{A loose estimator as reference.}
How well should a model that follows the number only loosely order its crowds? We simulate a \emph{loose estimator}, whose count is the request times a log-normal factor with a log standard deviation $s$ of 0.30 to 0.45. It is exact in only 5--7\% of images at 12--30, as rarely as most of the models (\cref{tab:word}). We read its counts through a model of people's counting noise, fitted to 5{,}360 single counts of 2{,}240 pictures, so that it is judged exactly as the models are. The noise model lets a picture be easy or hard to count, more often hard when the crowd is large, and a single count of a hard picture be off by a few objects or, rarely, zero; it reproduces where a third count falls when the first two differ, and how often recounts agree (\supref{sec:supp_refmodels,sec:supp_noise}). So read, it orders crowds of 12 to 25 at 0.72--0.80. A model's \emph{effective scatter} on a range of requests is the $s$ at which the estimator would order that range as well as the model does; once the order falls to one half, it has no finite value. Past ten, a model's \emph{reach} covers a range when its effective scatter there is at most 0.45: the model then orders the range at least as well as an estimator that is exact as rarely as the models are.

\para{How far does a change move the crowd?}
For two images from the same noise that differ in one thing, we take the mean $|\Delta\log(1{+}n)|$ over types. We compare a new number with four changes that leave the number's value alone: a new seed, the same number in digits (``12'' for ``twelve''), an unrelated edit (``a photograph of'' for ``a photo of''), and a \emph{fine-scale redraw}, which keeps the coarse part of the starting noise and takes only its fine part from the type's second seed. With $F_1$ and $F_2$ the 2D Fourier transforms of the two noises over the spatial axes and $r$ the spatial frequency in cycles per image,
\begin{equation}
z_f=\mathcal{F}^{-1}\big[\,\mathbf{1}[r\le f]\,F_1+\mathbf{1}[r>f]\,F_2\,\big]
\label{eq:redraw}
\end{equation}
is again white Gaussian noise, equal to the first seed's noise in all structure coarser than about $1/f$ of the image (\supref{sec:supp_brs}). Above eight cycles, the redraw replaces 98\% of the variance of \sdm's starting noise, yet keeps its coarse structure. How much of the layout a change keeps is the correlation of $24{\times}24$ grayscale thumbnails of the two images: it measures where the large light and dark regions fall, the picture's coarse composition. The order of the number change says whether it moves the crowd toward the request.

\section{Each Model Follows the Number Only So Far}
\label{sec:reach}\label{sec:say}

\begin{table}[t]
\centering
\footnotesize
\setlength{\tabcolsep}{2.2pt}
\resizebox{\linewidth}{!}{%
\begin{tabular}{@{}lccccc@{}}
\toprule
& \multicolumn{3}{c}{same-noise order} & effective & exact \\
\cmidrule(lr){2-4}
model (text encoder) & 2--8 & 12--25 & with 30 & scatter$^a$ & 12--30 \\
\midrule
\flux (T5, CLIP) & .90 & .47 \ci{.30}{.64} & .17 & $>$3 \ci{0.74}{$>$3} & 0\% \\
\sdm (CLIP, T5) & .80 & .57 \ci{.43}{.69} & .49 & 1.55 \ci{0.53}{$>$3} & 4\% \\
\sdxl (CLIP) & .89 & .65 \ci{.53}{.78} & .60 & 0.69 \ci{0.32}{$>$3} & 3\% \\
\fluxk (Qwen3) & .89 & .73 \ci{.62}{.84} & .44 & 0.42 \ci{0.24}{0.89} & 13\% \\
\qwen (Qwen2.5-VL) & .99 & .78 \ci{.65}{.90} & .59 & 0.33 \ci{0.17}{0.68} & 21\% \\
\midrule
\sdxl, screened types & .91 & .78 \ci{.68}{.87} & .75 & 0.32 \ci{0.18}{0.58} & 4\% \\
loose estimator, $s$ = .45 / .30 & .85 / .92 & .72 / .80 & & .45 / .30 & 5--7\% \\
\bottomrule
\end{tabular}}
\caption{\textbf{Within and past ten} (people's counts of 16 types, one seed; ``with 30'': the four pairs that include thirty). $^a$The loose estimator's scatter at which it would order crowds of 12--25 as the model does, with a 95\% interval; the reach covers 12--25 when it is at most 0.45 (\cref{sec:setup}).}
\label{tab:word}
\end{table}

\para{Every model follows small numbers.}
From two to eight objects, the larger request gives more objects in 80--99\% of the same-noise pairs whose counts differ (\cref{tab:word}), and the median crowd grows with the request (\cref{fig:reach}a).

\para{Past ten, two groups.}
From twelve to twenty-five (\cref{fig:reach}b), \sdm and \flux stop following the number and order their crowds worse than either reference estimator would (0.57 and 0.47). \sdm's median crowd stays between 4.5 and 8.5 objects whatever the request, and \flux's even shrinks, to one or two objects at thirty. \fluxk and \qwen keep following the number up to twenty-five about as loosely as the reference does (0.73 and 0.78), and their crowds grow to 16--20 objects. Paired on the same types, each of these two orders its crowds better than \sdm and \flux do (\supref{sec:supp_pairs}). \sdxl lies in between on our unscreened objects (0.65), but on types screened for isolated objects its crowd keeps growing up to thirty as a loose estimator's would. How far a number reaches depends on the objects as well as on the model.

\para{The trouble with thirty.}
The pairs that include thirty are ordered clearly worse than those up to twenty-five in \flux, \fluxk and \qwen, and in five of the seven tests that asked for both, the median crowd at thirty is smaller than at twenty-five (\supref{sec:supp_pairs}). We can place the end of \fluxk's and \qwen's reach only somewhere past twenty-five.

\para{What the usual scores miss.}
Past ten, the usual scores cannot say whether a model follows the number. Exact rates of 0--21\% (\cref{tab:word}) rank the two groups, but a model that misses may still follow the number loosely. \sdm, which does not follow it past ten, is exact in 4\% of images, about as often as a loose estimator that does (5--7\%); their orders, 0.57 against 0.72--0.80, tell them apart. From twenty to thirty, a loose estimator draws the requested count only 1--3 percentage points more often than a neighbouring count, and on 16 types a lift that small cannot be told from that of a model that ignores the number. A graded score is no better: a model that always draws sixteen objects has a mean relative error past ten of 0.27, lower than the looser of our two reference estimators (0.38) and close to the tighter one (0.25). Detectors add a bias of their own, since they undercount dense crowds and can reverse a model's direction: \gdino reads fewer objects in \fluxk's images when twenty-five are asked for than when twelve are, while people find the crowd growing (\supref{sec:supp_refmodels,sec:supp_noise,sec:supp_wordcount,sec:supp_brs,sec:supp_xtime}). Past ten, a model can miss nearly every count and still follow the number; its order against a reference counted the same way shows whether it does.

\section{The Count Changes Beneath a Kept Layout}
\label{sec:layout}\label{sec:reseed}

\begin{figure}[t]
\centering
\includegraphics[width=\linewidth]{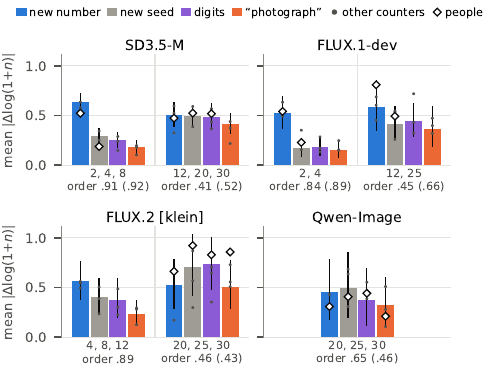}
\caption{\textbf{Four changes from the same noise}, within each model's reach and past it (for \fluxk and \qwen, at its edge, 20--30): mean change of $\log(1{+}n)$ when the number changes, when the seed changes, when the same number is written in digits, and when ``a photo of'' becomes ``a photograph of'' (bars: \samt on 64, 32, 32 and 16 types, 95\% intervals; dots: \gdino, \owl and their consensus; diamonds: people, 16 types). Below each group: the number change's order by \samt (by people).}
\label{fig:fourarm}
\end{figure}

\begin{figure*}[t]
\centering
\includegraphics[width=0.84\linewidth]{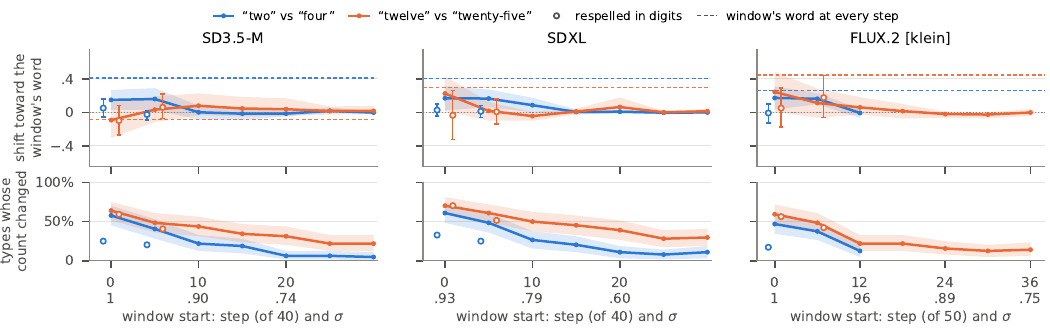}
\caption{\textbf{When the number word acts}, in three models (64 types, \samt). From the noise of the base request's image, the prompt asks for the other count of the pair during one window of steps (five of 40; six of 50 in \fluxk); half of the types go up, half down. Top: shift of the count toward the window's word (95\% intervals); open circles: the base number respelled in digits. Bottom: share of types whose count changed. Dashed: the window's word at every step. Ticks give each window's first step and noise level $\sigma$. Other counters, people's counts and longer windows: \supref{sec:supp_timing,sec:supp_xtime}.}
\label{fig:timing}
\end{figure*}

\begin{table}[t]
\centering
\footnotesize
\setlength{\tabcolsep}{2.6pt}
\resizebox{\linewidth}{!}{%
\begin{tabular}{@{}lccccc@{}}
\toprule
& layout & \multicolumn{2}{c}{\samt, 64 types} & \multicolumn{2}{c}{people, 16 types} \\
\cmidrule(lr){3-4}\cmidrule(l){5-6}
change from the same noise & kept & 2, 4, 8 & 12, 20, 30 & 4, 8 & 12, 20, 30 \\
\midrule
a new number & .70 & .63 & .51 & .37 & .43 \\
a new seed & .10 & .29 & .49 & .25 & .53 \\
the number in digits & .76 & .25 & .48 & & .52 \\
``photograph'' for ``photo'' & & .18 & .42 & & \\
redraw above 2 cycles & .40 & .25 & .49 & & \\
redraw above 4 cycles & .54 & .22 & .34 & .29 & .32 \\
redraw above 8 cycles & .74 & .20 & .33 & .19 & .30 \\
redraw above 16 cycles, 16 types$^a$ & .86 & .25 & .42 & & \\
\midrule
the same picture, counted again & & & & .12 & .12 \\
\bottomrule
\end{tabular}}
\caption{\textbf{How far each change moves \sdm's count}, within the reach and past it (mean change of $\log(1{+}n)$ over object types), and how much of the layout it keeps past the reach (correlation of $24{\times}24$ thumbnails with the original image). A new number: pairs of requests on seed 1; its order is 0.91 within the reach and 0.41 past it (\samt). Last row: the same picture counted by people in two different studies. $^a$\samt, 16 people-counted types; past the reach a new number moves the count by .59, a new seed by .69. Intervals and sources: \supref{sec:supp_fourarm,sec:supp_reseedpeople,sec:supp_brs,sec:supp_layout,sec:supp_noise}.}
\label{tab:changes}
\end{table}

\begin{figure}[t]
\centering
\includegraphics[width=0.94\linewidth]{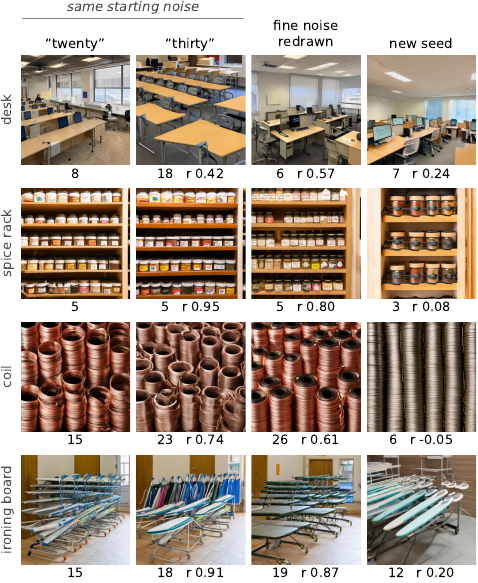}
\caption{\textbf{Fine-scale redraws} next to a new number and a new seed (\sdm; four of the 16 people-counted types, drawn at random). From the noise of the image asked for twenty: thirty asked from the same noise; the noise above eight cycles per image redrawn from the type's second seed, still asked for twenty; twenty from the second seed. People's counts, and $r$, the correlation of $24{\times}24$ thumbnails with the first image. In these four, thirty gives as many objects as twenty or more; across all types, the larger request gives more objects in only about half of the pairs (\cref{sec:layout}).}
\label{fig:redraw}
\end{figure}

If the starting noise fixed a large crowd's count, a new number from the same noise would leave the count where it was. In \sdm it does not, but it does not steer the count either.

\para{The layout stays.}
A new number keeps most of the picture (\cref{fig:teaser}a, \cref{tab:changes}). Past the reach, maps of the objects' boxes correlate at 0.58 between two images from the same noise under a new number, against 0.32 under a new seed (\supref{sec:supp_layout}). The noise decides where things go, as Zhao~\etal found~\cite{zhao2025noise}.

\para{The count moves as under a new seed.}
Past the reach, a new number moves \sdm's count about as far as a new seed or the same number written in digits, and only 1.2--1.5 times as far as an unrelated edit (\cref{tab:changes,fig:fourarm}). It does not push the count toward the request: over all nine requests past ten and both seeds, the larger request gives more objects in only half of the pairs whose counts differ, and the number moves the crowd 0.89 times as far as a seed. People's counts of 16 types agree. A loose estimator counted the same way would move its crowd 1.4--1.7 times as far as a seed and order it at 0.80--0.88; the model's order is below 0.54 with 95\% confidence, one-sided (\supref{sec:supp_noise,sec:supp_pairs,sec:supp_reseed}). Within the reach, by contrast, a new number moves the count further than a new seed, and toward the request. Past it, the number moves the crowd but no longer steers it.

\para{A seed that keeps the layout.}
The count of a large crowd depends on details finer than the layout. A new seed redraws the whole picture, so it is a generous yardstick; a stricter one keeps the coarse noise and redraws only the noise finer than $f$ cycles per image (\cref{eq:redraw}). Redrawn above eight cycles, the picture keeps its layout as well as under a new number (\cref{tab:changes,fig:redraw}). Yet past the reach its count moves two-thirds as far as under a new number. By people's counts of 16 types the ratio is 0.69 \ci{0.38}{1.20}, and the change is two and a half times the difference between two countings of the same picture. Even redrawn above sixteen cycles, where the thumbnails correlate at 0.86, a large crowd's count moves 0.71 times as far as under a new number (16 types; \supref{sec:supp_brs}).

\para{What changes at the reach is the number.}
It is not that noise starts to matter: fine-scale noise moves small crowds too, two-thirds to nine-tenths as far as a new seed does. Within the reach, a new number moves the count 2.5 to 3.2 times as far as a redraw above two to eight cycles, and toward the request. Past it, only 1.0 to 1.5 times as far, and in no direction. The number loses its direction and most of its margin over noise.

\para{Other models.}
The redraws behave the same way in the two models that follow the number furthest. By people's counts of 16 types (\samt reads many of \fluxk's crowds as empty), a redraw above four or eight cycles moves \fluxk's count about half as far as a new number within its reach and about four-fifths as far at its edge. In \qwen, by \samt, the redraw moves the count less than half as far as a new number at four and twelve, and four-fifths as far at twenty and thirty (\supref{sec:supp_brs}). In both models, the number's margin over fine-scale noise shrinks toward the edge of the reach. In \flux, past its reach, a new number moves the crowd further than a new seed, without a reliable direction (\supref{sec:supp_fluxreseed}).

\section{The Number Arrives, Without a Direction}
\label{sec:when}\label{sec:heard}

Past the reach, is the large number lost before the denoiser, too weak, too late, or simply without a direction? We follow it through \sdm, timing it also in \sdxl and \fluxk.

\para{It is encoded.}
From \sdm's pooled CLIP vector, a ridge regression trained on the other number words puts a held-out large number in the right order in 0.89 of same-object pairs, nearly as often as a small one (0.98; \supref{sec:supp_mech}).

\para{It acts early, where small numbers act.}
Swapping the other request of a pair into the prompt for a window of five or six steps shows when a number word acts (\cref{fig:timing}). In all three models we timed, small number words move the count toward themselves only in the first windows, while the same number respelled in digits changes fewer counts and in no direction. The large word acts in the first window too, but without a clear direction in \sdm: it changes 64\% of the counts there, about as many as its respelling does, and only 46\% of those changes go its way. Later windows and people's counts of the swaps are in \supref{sec:supp_timing,sec:supp_xtime}.

\begin{table}[t]
\centering
\footnotesize
\setlength{\tabcolsep}{2.5pt}
\resizebox{\linewidth}{!}{%
\begin{tabular}{@{}lcc@{}}
\toprule
\sdm, same noise (\samt order) & 2--8 & 12--30 \\
\midrule
plain prompt (two seeds) & .86 & .50 \ci{.45}{.55} \\
word amplified, $L{=}10$ & .88 & .41 \ci{.28}{.54} (.15) \\
number in CLIP only / in T5 only & .85 / .85 & .47 / .55 \\
reward-tuned: Flow-GRPO / DiffusionNFT$^a$ & .95 / .92 & .38 / .32 \\
count LoRA / shuffled-count LoRA & .92 / .91 & .58 / .44 \\
count LoRA, people's counts & .98 & .43 \ci{.26}{.61} \\
``tiny'' & .85 & .65 \ci{.57}{.72} (.46) \\
\midrule
add $v(\text{four}){-}v(\text{two})$, $L{=}8$ & \multicolumn{2}{c}{crowd for twelve shrinks: median 8$\to$4} \\
\bottomrule
\end{tabular}}
\caption{\textbf{What we tried, to make \sdm follow a large number.} Same-noise order at 2--8 and at 12--30 (in parentheses: by \gdino); for the amplified word and the streams, two against four and twelve against twenty-five. Baselines: the amplified word's own unsteered pairs order .94 and .52; for the LoRAs and the cue, seed 1's plain prompt orders .89 and .46. By people's counts the count LoRA orders large crowds no better than the shuffled one ($+$0.08 \ci{$-$0.12}{0.27}; by \samt $+$0.14 \ci{0.01}{0.27}). $^a$$512^2$, screened types, base model .88 and .49. Details: \supref{sec:supp_steer,sec:supp_borrow,sec:supp_stream,sec:supp_ceiltrain,sec:supp_room,sec:supp_reward}.}
\label{tab:attempts}
\end{table}

\para{Nothing we tried gives it a clear direction.}
We pushed on the word, the text streams, the weights and the prompt, all in \sdm (\cref{tab:attempts}). Amplifying the difference between the two requests' velocities in the first ten steps~\cite{wu2024contrastive,sepehri2026athena} keeps small pairs ordered and leaves large pairs unordered. Adding the velocity difference between ``four'' and ``two'' to a request for twelve shrinks the crowd instead of growing it. Given different numbers in its CLIP and T5 streams, \sdm follows either stream's small number and neither stream's large one, and checkpoints rewarded on small counts do not order large crowds either. A LoRA trained on 1{,}653 of \sdm's own counted crowds of 12 to 30 learns counts up to about twelve, its most common training count. Past that, it orders crowds somewhat better than the same LoRA trained on shuffled counts by \samt, but not by people's counts. The shuffled LoRA draws 33 to 36 objects at every request from twelve to twenty-five by people's counts, so the model can draw large crowds; what it lacks is a mapping from a large number to one. The one cue that moves the order partway is the word ``tiny'', under \samt and \owl but not under \gdino. The large number reaches the denoiser in time, but nothing we tried ties it to a change of count.\looseness=-1

\section{A Large Count Can Be Shown Instead}
\label{sec:show}\label{sec:layoutstart}

\begin{figure}[t]
\centering
\includegraphics[width=0.96\linewidth]{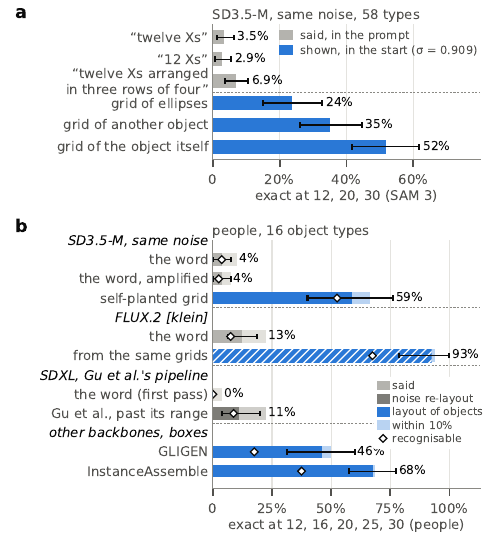}
\caption{\textbf{Saying against showing.} (a) \sdm, same noise, 58 types, exact at 12, 20 and 30 by \samt: three ways of saying the count, and three contents of a noised grid in the starting latent ($\sigma{=}0.909$), which the sentence describes. (b) People's counts, 16 types, at 12, 16, 20, 25 and 30: exact (bars, 95\% intervals), within 10\% (light) and exact with a recognisable object (diamonds), for prompts, \sdxl with Gu~\etal's noise re-layout, the self-planted start, \fluxk from the same grids (hatched: its render keeps much of the supplied composite), and two layout models given the grid as boxes.}
\label{fig:saysow}
\end{figure}

If the number word has no direction past the reach, can the model draw a large crowd at all? We give it the count in another form: as a layout in the starting latent.

\para{Saying against showing, from the same noise.}
In \sdm, from the noise of each type's image, we either say the count (as a word, in digits, or in a sentence naming the rows and columns of a grid, ``twelve $X$s arranged in three rows of four'') or show it. To show it, we start from a grid of the requested count, noised to $\sigma{=}0.909$, and let the unchanged model run its last 31 of 40 steps with the plain prompt (\supref{sec:supp_content}). Every way of saying the count is exact in at most 6.9\% of images (\samt, on the 58 types with a cut-out). Even a grid of plain ellipses in the object's mean colour is exact in 24\%. A grid of the object itself, cut out of the model's own image of two to four objects, is exact in 52\%, and a grid of another object lies in between (\cref{fig:saysow}a): the more the grid looks like the object, the more it helps.

\para{By people's counts.}
People find the asked count in 4\% of \sdm's images from the prompt and in 59\% from a grid of its own object (\cref{fig:saysow}b). Layout models given the grid as boxes, GLIGEN~\cite{li2023gligen} and InstanceAssemble~\cite{xiang2025instanceassemble}, reach 46\% and 68\%.

\para{A planted count holds against the number word, not the object word.}
Started from a noised grid of $N$ objects and asked for a different count $M$, \sdm, \sdxl and \flux draw $M$ in only 0.7--3.5\% of images and keep $N$ in 50--64\% net of chance (screened types; \supref{sec:supp_conflict}). In \sdm these starts come at the end of the steps in which number words act, so the planted count survives the rest of sampling rather than beating the word head on; started earlier, from a crowd between the two requests, a large word still barely pulls the count (\supref{sec:supp_straddle}). Asked instead for twenty of another object of similar shape, \sdm repaints most of the planted grid as the named object in 75\% of images (screened types). The layout carries the count; the word decides what is drawn.

\para{Relation to count-aware noise.}
The planted start is SDEdit~\cite{meng2022sdedit} from a tiled composite of the model's own object~\cite{composeanything2025}, close to the count-aware noise prior of Zhao~\etal~\cite{zhao2025noise}, who showed that such a prior raises exact counts. We use it as a probe of where a large count can enter the model, not as a method: it costs image quality (PickScore~\cite{kirstain2023pickscore} 0.57--0.85 lower) and is no more accurate than InstanceAssemble (\supref{sec:supp_saysow,sec:supp_method,sec:supp_gligen}). Together with the redraws of \cref{sec:layout}, it shows where a large count can come from: a grid written into the noise carries it, while the random noise's own layout does not.

\section{Discussion}
\label{sec:discussion}

\begin{table}[t]
\centering
\footnotesize
\setlength{\tabcolsep}{3pt}
\begin{tabular}{@{}>{\raggedright\arraybackslash}p{0.035\linewidth}>{\raggedright\arraybackslash}p{0.93\linewidth}@{}}
\toprule
1. & For a few dozen object types, draw four or five widely spaced requests past ten from one seed per type. \\
2. & Count the images with people, or with a counter checked against people. \\
3. & Report the same-noise order next to a loose estimator read through the same counting noise. \\
4. & From the same noise, add a new seed and an unrelated edit: a number that steers moves the crowd further than they do, and toward the request. \\
\bottomrule
\end{tabular}
\caption{\textbf{A same-noise test for counting past ten.}}
\label{tab:recipe}
\end{table}

\para{Layout and count.}
In \sdm the layout of a picture is set early, by the coarse part of the noise, while the count of a large crowd depends on finer details. The number word acts while the layout forms, and within the reach it points the count toward the request. Past the reach it still arrives in time, but we found nothing that ties its value to a count, and the count is left to the finer details of the noise.

\para{For benchmarks.}
Past ten, we suggest a same-noise test instead of an exact rate (\cref{tab:recipe}). It needs no access to a model's internals, only to its starting noise. We will release the code, the images, people's counts and their answers.

\para{For control.}
Past the reach, digits, a sentence naming the grid, the native resolution, other guidance scales and the best of four samples are each exact in at most 6.2\% of \sdm's images (64 types), and steering the word and small-count rewards do not order its large crowds (\supref{sec:supp_wording,sec:supp_reward,sec:supp_steer}), while a layout in the latent or as boxes draws the count far more often. For these models, a large count is better given as a layout than as a word, at some cost in image quality.

\section*{Limitations}
\addcontentsline{toc}{section}{Limitations}
\para{Scope.} Five open models, single-object prompts asking for up to thirty objects (up to forty in two side tests), and one prompt template. \qwen ran with a 4-bit transformer at $1024^2$, below its model card's $1328^2$, and \sdm and \flux at $768^2$, below their native size; closed models were not tested, though benchmarks find them failing past ten as well~\cite{t2icountbench2025,numbench2026}.
\para{Evidence.} The equivalence of a new number and a new seed is clean in \sdm; in \flux the number moves the crowd further than a seed, and in \fluxk and \qwen we could test it only at the edge of their reach, where it separates little. People counted the redraws of 16 types; on these types the number's margin over the redraws within the reach is smaller than on all 64, and some of \fluxk's large crowds are ambiguous to people.
The analysis inside the model is in \sdm (timing also in \sdxl and \fluxk). People counted 16 types per model and agree only moderately on dense crowds, so most rankings between models remain unresolved, and requests stop at thirty, before the two longest reaches end. The count LoRA is only one training recipe.

\section*{LLM Usage Statement}
We used Claude (Anthropic), an LLM-based assistant, as a coding and writing assistant. It helped write and run code for image generation, counting, analysis and figures; helped plan and run experiments, including the crowdsourced counting studies; helped search related work and check numbers and references; and helped draft and revise the text. The images we analyse are outputs of the models under study, and every count attributed to people comes from human annotators. The authors directed the work, made all research decisions and take full responsibility for the paper.

{
    \small
    \bibliographystyle{ieeenat_fullname}
    \bibliography{main}

\begin{thebibliography}{69}
\providecommand{\natexlab}[1]{#1}
\providecommand{\url}[1]{\texttt{#1}}
\expandafter\ifx\csname urlstyle\endcsname\relax
  \providecommand{\doi}[1]{doi: #1}\else
  \providecommand{\doi}{doi: \begingroup \urlstyle{rm}\Url}\fi

\bibitem[Bai et~al.(2025)]{bai2025qwen25vl}
Shuai Bai et~al.
\newblock {Qwen2.5-VL} technical report.
\newblock \emph{arXiv preprint arXiv:2502.13923}, 2025.

\bibitem[Balaji et~al.(2022)Balaji, Nah, Huang, Vahdat, Song, Zhang, Kreis,
  Aittala, et~al.]{balaji2022ediffi}
Yogesh Balaji, Seungjun Nah, Xun Huang, Arash Vahdat, Jiaming Song, Qinsheng
  Zhang, Karsten Kreis, Miika Aittala, et~al.
\newblock {eDiff-I}: Text-to-image diffusion models with an ensemble of expert
  denoisers.
\newblock \emph{arXiv preprint arXiv:2211.01324}, 2022.

\bibitem[Ban et~al.(2025)Ban, Wang, Zhou, Gong, Hsieh, and
  Cheng]{ban2025crystalball}
Yuanhao Ban, Ruochen Wang, Tianyi Zhou, Boqing Gong, Cho-Jui Hsieh, and Minhao
  Cheng.
\newblock The {Crystal Ball} hypothesis in diffusion models: Anticipating
  object positions from initial noise.
\newblock In \emph{International Conference on Learning Representations
  (ICLR)}, 2025.

\bibitem[Binyamin et~al.(2025)Binyamin, Tewel, Segev, Hirsch, Rassin, and
  Chechik]{binyamin2025makeitcount}
Lital Binyamin, Yoad Tewel, Hilit Segev, Eran Hirsch, Royi Rassin, and Gal
  Chechik.
\newblock {Make It Count}: Text-to-image generation with an accurate number of
  objects.
\newblock In \emph{Proceedings of the IEEE/CVF Conference on Computer Vision
  and Pattern Recognition (CVPR)}, pages 13242--13251, 2025.

\bibitem[Biroli et~al.(2024)Biroli, Bonnaire, de~Bortoli, and
  M{\'e}zard]{biroli2024dynamical}
Giulio Biroli, Tony Bonnaire, Valentin de Bortoli, and Marc M{\'e}zard.
\newblock Dynamical regimes of diffusion models.
\newblock \emph{Nature Communications}, 15:\penalty0 9957, 2024.

\bibitem[{Black Forest Labs}(2024)]{flux2024}
{Black Forest Labs}.
\newblock {FLUX.1 [dev]}.
\newblock \url{https://huggingface.co/black-forest-labs/FLUX.1-dev}, 2024.

\bibitem[{Black Forest Labs}(2026)]{flux2klein2026}
{Black Forest Labs}.
\newblock {FLUX.2 [klein]}.
\newblock \url{https://huggingface.co/black-forest-labs/FLUX.2-klein-base-4B},
  2026.

\bibitem[Boo et~al.(2025)Boo, Kim, Lee, Lee, Lee, Choi, and
  Cho]{boo2025countsteer}
Hyemin Boo, Hyoryung Kim, Myungjin Lee, Seunghyeon Lee, Jiyoung Lee, Jang-Hwan
  Choi, and Hyunsoo Cho.
\newblock {CountSteer}: Steering attention for object counting in diffusion
  models.
\newblock \emph{arXiv preprint arXiv:2511.11253}, 2025.

\bibitem[Carion et~al.(2026)Carion, Gustafson, Hu, Debnath, Hu, Suris
  Coll-Vinent, Ryali, Alwala, et~al.]{carion2026sam3}
Nicolas Carion, Laura Gustafson, Yuan-Ting Hu, Shoubhik Debnath, Ronghang Hu,
  Didac Suris Coll-Vinent, Chaitanya Ryali, Kalyan~Vasudev Alwala, et~al.
\newblock {SAM} 3: Segment anything with concepts.
\newblock In \emph{International Conference on Learning Representations
  (ICLR)}, 2026.

\bibitem[Clark and Jaini(2023)]{clark2023zeroshot}
Kevin Clark and Priyank Jaini.
\newblock Text-to-image diffusion models are zero-shot classifiers.
\newblock In \emph{Advances in Neural Information Processing Systems
  (NeurIPS)}, 2023.

\bibitem[Dahary et~al.(2025)Dahary, Cohen, Patashnik, Aberman, and
  Cohen-Or]{dahary2025decisive}
Omer Dahary, Yehonathan Cohen, Or Patashnik, Kfir Aberman, and Daniel Cohen-Or.
\newblock Be decisive: Noise-induced layouts for multi-subject generation.
\newblock In \emph{Proceedings of the Special Interest Group on Computer
  Graphics and Interactive Techniques Conference Conference Papers (SIGGRAPH
  Conference Papers '25)}, 2025.

\bibitem[Esser et~al.(2024)Esser, Kulal, Blattmann, Entezari, M{\"u}ller,
  Saini, Levi, Lorenz, et~al.]{esser2024sd3}
Patrick Esser, Sumith Kulal, Andreas Blattmann, Rahim Entezari, Jonas
  M{\"u}ller, Harry Saini, Yam Levi, Dominik Lorenz, et~al.
\newblock Scaling rectified flow transformers for high-resolution image
  synthesis.
\newblock In \emph{Proceedings of the 41st International Conference on Machine
  Learning (ICML)}, pages 12606--12633, 2024.

\bibitem[Everaert et~al.(2024)Everaert, Fitsios, Bocchio, Arpa, S{\"u}sstrunk,
  and Achanta]{everaert2024signalleak}
Martin~Nicolas Everaert, Athanasios Fitsios, Marco Bocchio, Sami Arpa, Sabine
  S{\"u}sstrunk, and Radhakrishna Achanta.
\newblock Exploiting the signal-leak bias in diffusion models.
\newblock In \emph{Proceedings of the IEEE/CVF Winter Conference on
  Applications of Computer Vision (WACV)}, pages 4025--4034, 2024.

\bibitem[Ghosh et~al.(2023)Ghosh, Hajishirzi, and Schmidt]{ghosh2023geneval}
Dhruba Ghosh, Hannaneh Hajishirzi, and Ludwig Schmidt.
\newblock {GenEval}: An object-focused framework for evaluating text-to-image
  alignment.
\newblock In \emph{Advances in Neural Information Processing Systems (NeurIPS),
  Datasets and Benchmarks Track}, pages 52132--52152, 2023.

\bibitem[Goodman and Kruskal(1954)]{goodman1954measures}
Leo~A. Goodman and William~H. Kruskal.
\newblock Measures of association for cross classifications.
\newblock \emph{Journal of the American Statistical Association}, 49\penalty0
  (268):\penalty0 732--764, 1954.

\bibitem[Gorgun et~al.(2025)Gorgun, Sammani, Deligiannis, Schiele, and
  Fischer]{gorgun2025temporal}
Ada Gorgun, Fawaz Sammani, Nikos Deligiannis, Bernt Schiele, and Jonas Fischer.
\newblock Temporal concept dynamics in diffusion models via prompt-conditioned
  interventions.
\newblock \emph{arXiv preprint arXiv:2512.08486}, 2025.

\bibitem[Gu et~al.(2026)Gu, Li, Wu, Wong, Wu, Li, Yu, and
  Kankanhalli]{gu2026noise}
Xiaoling Gu, Xuelong Li, Shengqi Wu, Yongkang Wong, Zizhao Wu, Huan Li, Zhou
  Yu, and Mohan Kankanhalli.
\newblock Mitigating noise-induced layout priors for object counting in
  diffusion models.
\newblock In \emph{Proceedings of the 43rd International Conference on Machine
  Learning (ICML)}, pages 37047--37072, 2026.

\bibitem[Guo et~al.(2025)Guo, Huo, Liang, Shi, Song, Zhang, and
  Zhuang]{t2icountbench2025}
Xuyang Guo, Jiayan Huo, Yingyu Liang, Zhenmei Shi, Zhao Song, Jiahao Zhang, and
  Zhen Zhuang.
\newblock Text-to-image diffusion models cannot count, and prompt refinement
  cannot help.
\newblock \emph{arXiv preprint arXiv:2503.06884}, 2025.

\bibitem[Gupta et~al.(2019)Gupta, Doll{\'a}r, and Girshick]{gupta2019lvis}
Agrim Gupta, Piotr Doll{\'a}r, and Ross Girshick.
\newblock {LVIS}: A dataset for large vocabulary instance segmentation.
\newblock In \emph{Proceedings of the IEEE/CVF Conference on Computer Vision
  and Pattern Recognition (CVPR)}, pages 5356--5364, 2019.

\bibitem[Hertz et~al.(2023)Hertz, Mokady, Tenenbaum, Aberman, Pritch, and
  Cohen-Or]{hertz2023prompt}
Amir Hertz, Ron Mokady, Jay Tenenbaum, Kfir Aberman, Yael Pritch, and Daniel
  Cohen-Or.
\newblock Prompt-to-prompt image editing with cross-attention control.
\newblock In \emph{International Conference on Learning Representations
  (ICLR)}, 2023.

\bibitem[Ho and Salimans(2022)]{ho2022cfg}
Jonathan Ho and Tim Salimans.
\newblock Classifier-free diffusion guidance.
\newblock \emph{arXiv preprint arXiv:2207.12598}, 2022.

\bibitem[Huang et~al.(2025)Huang, Duan, Sun, Xie, Li, and
  Liu]{huang2025t2icompbenchpp}
Kaiyi Huang, Chengqi Duan, Kaiyue Sun, Enze Xie, Zhenguo Li, and Xihui Liu.
\newblock {T2I-CompBench++}: An enhanced and comprehensive benchmark for
  compositional text-to-image generation.
\newblock \emph{IEEE Transactions on Pattern Analysis and Machine
  Intelligence}, 47\penalty0 (5):\penalty0 3563--3579, 2025.

\bibitem[Jeong et~al.(2026)Jeong, Uselis, Laina, Oh, and Rohrbach]{mosaic2026}
Yujin Jeong, Arnas Uselis, Iro Laina, Seong~Joon Oh, and Anna Rohrbach.
\newblock When do diffusion models learn to generate multiple objects?
\newblock In \emph{Proceedings of the 43rd International Conference on Machine
  Learning (ICML)}, pages 51250--51282, 2026.

\bibitem[Kaji{\'c} et~al.(2024)Kaji{\'c}, Wiles, Albuquerque, Bauer, Wang,
  Pont-Tuset, and Nematzadeh]{kajic2024geckonum}
Ivana Kaji{\'c}, Olivia Wiles, Isabela Albuquerque, Matthias Bauer, Su Wang,
  Jordi Pont-Tuset, and Aida Nematzadeh.
\newblock Evaluating numerical reasoning in text-to-image models.
\newblock In \emph{Advances in Neural Information Processing Systems (NeurIPS),
  Datasets and Benchmarks Track}, pages 42211--42224, 2024.

\bibitem[Kamath et~al.(2025)Kamath, Chang, Krishna, Zettlemoyer, Hu, and
  Ghazvininejad]{kamath2025geneval2}
Amita Kamath, Kai-Wei Chang, Ranjay Krishna, Luke Zettlemoyer, Yushi Hu, and
  Marjan Ghazvininejad.
\newblock {GenEval 2}: Addressing benchmark drift in text-to-image evaluation.
\newblock \emph{arXiv preprint arXiv:2512.16853}, 2025.

\bibitem[Kang et~al.(2025)Kang, Galim, Koo, and Cho]{kang2025countingguidance}
Wonjun Kang, Kevin Galim, Hyung~Il Koo, and Nam~Ik Cho.
\newblock Counting guidance for high fidelity text-to-image synthesis.
\newblock In \emph{Proceedings of the Winter Conference on Applications of
  Computer Vision (WACV)}, pages 899--908, 2025.

\bibitem[Khan et~al.(2025)Khan, Chen, and Schmid]{composeanything2025}
Zeeshan Khan, Shizhe Chen, and Cordelia Schmid.
\newblock {ComposeAnything}: Composite object priors for text-to-image
  generation.
\newblock \emph{arXiv preprint arXiv:2505.24086}, 2025.

\bibitem[Kirstain et~al.(2023)Kirstain, Polyak, Singer, Matiana, Penna, and
  Levy]{kirstain2023pickscore}
Yuval Kirstain, Adam Polyak, Uriel Singer, Shahbuland Matiana, Joe Penna, and
  Omer Levy.
\newblock {Pick-a-Pic}: An open dataset of user preferences for text-to-image
  generation.
\newblock In \emph{Advances in Neural Information Processing Systems
  (NeurIPS)}, pages 36652--36663, 2023.

\bibitem[Lee et~al.(2025)Lee, Lee, and Lee]{countcluster2025}
Joohyeon Lee, Jin-Seop Lee, and Jee-Hyong Lee.
\newblock {CountCluster}: Training-free object quantity guidance with
  cross-attention map clustering for text-to-image generation.
\newblock \emph{arXiv preprint arXiv:2508.10710}, 2025.

\bibitem[Li et~al.(2023{\natexlab{a}})Li, Prabhudesai, Duggal, Brown, and
  Pathak]{li2023diffclassifier}
Alexander~C. Li, Mihir Prabhudesai, Shivam Duggal, Ellis Brown, and Deepak
  Pathak.
\newblock Your diffusion model is secretly a zero-shot classifier.
\newblock In \emph{Proceedings of the IEEE/CVF International Conference on
  Computer Vision (ICCV)}, 2023{\natexlab{a}}.

\bibitem[Li and Chen(2024)]{li2024critical}
Marvin Li and Sitan Chen.
\newblock Critical windows: Non-asymptotic theory for feature emergence in
  diffusion models.
\newblock In \emph{Proceedings of the 41st International Conference on Machine
  Learning (ICML)}, pages 27474--27498, 2024.

\bibitem[Li et~al.(2025)Li, Le, Xu, and Salzmann]{li2025seeds}
Shuangqi Li, Hieu Le, Jingyi Xu, and Mathieu Salzmann.
\newblock Enhancing compositional text-to-image generation with reliable random
  seeds.
\newblock In \emph{International Conference on Learning Representations
  (ICLR)}, 2025.

\bibitem[Li et~al.(2023{\natexlab{b}})Li, Liu, Wu, Mu, Yang, Gao, Li, and
  Lee]{li2023gligen}
Yuheng Li, Haotian Liu, Qingyang Wu, Fangzhou Mu, Jianwei Yang, Jianfeng Gao,
  Chunyuan Li, and Yong~Jae Lee.
\newblock {GLIGEN}: Open-set grounded text-to-image generation.
\newblock In \emph{Proceedings of the IEEE/CVF Conference on Computer Vision
  and Pattern Recognition (CVPR)}, pages 22511--22521, 2023{\natexlab{b}}.

\bibitem[Lian et~al.(2024)Lian, Li, Yala, and Darrell]{lian2024lmd}
Long Lian, Boyi Li, Adam Yala, and Trevor Darrell.
\newblock {LLM}-grounded diffusion: Enhancing prompt understanding of
  text-to-image diffusion models with large language models.
\newblock \emph{Transactions on Machine Learning Research}, 2024.

\bibitem[Lipman et~al.(2023)Lipman, Chen, Ben-Hamu, Nickel, and
  Le]{lipman2023flow}
Yaron Lipman, Ricky T.~Q. Chen, Heli Ben-Hamu, Maximilian Nickel, and Matthew
  Le.
\newblock Flow matching for generative modeling.
\newblock In \emph{International Conference on Learning Representations
  (ICLR)}, 2023.

\bibitem[Liu et~al.(2024{\natexlab{a}})Liu, Zhang, Xie, Faccio, Xu, Xiang,
  Shou, Perez-Rua, and Schmidhuber]{liu2025tgate}
Haozhe Liu, Wentian Zhang, Jinheng Xie, Francesco Faccio, Mengmeng Xu, Tao
  Xiang, Mike~Zheng Shou, Juan-Manuel Perez-Rua, and J{\"u}rgen Schmidhuber.
\newblock Faster diffusion via temporal attention decomposition.
\newblock \emph{arXiv preprint arXiv:2404.02747}, 2024{\natexlab{a}}.
\newblock First version titled ``Cross-Attention Makes Inference Cumbersome in
  Text-to-Image Diffusion Models''.

\bibitem[Liu et~al.(2025)Liu, Liu, Liang, Li, Liu, Wang, Wan, Zhang, and
  Ouyang]{liu2025flowgrpo}
Jie Liu, Gongye Liu, Jiajun Liang, Yangguang Li, Jiaheng Liu, Xintao Wang,
  Pengfei Wan, Di Zhang, and Wanli Ouyang.
\newblock {Flow-GRPO}: Training flow matching models via online {RL}.
\newblock In \emph{Advances in Neural Information Processing Systems
  (NeurIPS)}, pages 40783--40818, 2025.

\bibitem[Liu et~al.(2024{\natexlab{b}})Liu, Zeng, Ren, Li, Zhang, Yang, Jiang,
  Li, et~al.]{liu2024gdino}
Shilong Liu, Zhaoyang Zeng, Tianhe Ren, Feng Li, Hao Zhang, Jie Yang, Qing
  Jiang, Chunyuan Li, et~al.
\newblock Grounding {DINO}: Marrying {DINO} with grounded pre-training for
  open-set object detection.
\newblock In \emph{Proceedings of the European Conference on Computer Vision
  (ECCV)}, 2024{\natexlab{b}}.

\bibitem[Liu et~al.(2023)Liu, Gong, and Liu]{liu2023rectified}
Xingchao Liu, Chengyue Gong, and Qiang Liu.
\newblock Flow straight and fast: Learning to generate and transfer data with
  rectified flow.
\newblock In \emph{International Conference on Learning Representations
  (ICLR)}, 2023.

\bibitem[Mao et~al.(2024)Mao, Wang, and Aizawa]{mao2024lottery}
Jiafeng Mao, Xueting Wang, and Kiyoharu Aizawa.
\newblock The lottery ticket hypothesis in denoising: Towards semantic-driven
  initialization.
\newblock In \emph{Proceedings of the European Conference on Computer Vision
  (ECCV)}, pages 93--109, 2024.

\bibitem[Mazur et~al.(2026)Mazur, Konovalova, and Alanov]{mazur2026diffract}
Artyom Mazur, Nina Konovalova, and Aibek Alanov.
\newblock {DifFRACT}: Diffusion feature reconstruction and attribution for
  circuit tracing.
\newblock \emph{arXiv preprint arXiv:2606.15796}, 2026.

\bibitem[Meng et~al.(2022)Meng, He, Song, Song, Wu, Zhu, and
  Ermon]{meng2022sdedit}
Chenlin Meng, Yutong He, Yang Song, Jiaming Song, Jiajun Wu, Jun-Yan Zhu, and
  Stefano Ermon.
\newblock {SDEdit}: Guided image synthesis and editing with stochastic
  differential equations.
\newblock In \emph{International Conference on Learning Representations
  (ICLR)}, 2022.

\bibitem[Miller(1995)]{miller1995wordnet}
George~A. Miller.
\newblock {WordNet}: A lexical database for {English}.
\newblock \emph{Communications of the ACM}, 38\penalty0 (11):\penalty0 39--41,
  1995.

\bibitem[Minderer et~al.(2023)Minderer, Gritsenko, and
  Houlsby]{minderer2023owlv2}
Matthias Minderer, Alexey Gritsenko, and Neil Houlsby.
\newblock Scaling open-vocabulary object detection.
\newblock In \emph{Advances in Neural Information Processing Systems
  (NeurIPS)}, pages 72983--73007, 2023.

\bibitem[Mondal et~al.(2026)Mondal, Nag, Banerjee, Llad{\'o}s, Zhu, and
  Dutta]{countloop2025}
Anindya Mondal, Sauradip Nag, Ayan Banerjee, Josep Llad{\'o}s, Xiatian Zhu, and
  Anjan Dutta.
\newblock {CountLoop}: Training-free high-instance image generation via
  iterative agent guidance.
\newblock \emph{Transactions on Machine Learning Research}, 2026.
\newblock arXiv:2508.16644v5.

\bibitem[Ngo et~al.(2026)Ngo, Ngo, Le, and Le]{ngo2026dense}
Bach-Hoang Ngo, Si-Tri Ngo, Hieu Le, and Trung-Nghia Le.
\newblock Learning to generate multiple objects from dense and occluded
  layouts.
\newblock \emph{arXiv preprint arXiv:2607.03488}, 2026.

\bibitem[Paiss et~al.(2023)Paiss, Ephrat, Tov, Zada, Mosseri, Irani, and
  Dekel]{paiss2023clipcount}
Roni Paiss, Ariel Ephrat, Omer Tov, Shiran Zada, Inbar Mosseri, Michal Irani,
  and Tali Dekel.
\newblock Teaching {CLIP} to count to ten.
\newblock In \emph{Proceedings of the IEEE/CVF International Conference on
  Computer Vision (ICCV)}, pages 3170--3180, 2023.

\bibitem[Patashnik et~al.(2023)Patashnik, Garibi, Azuri, Averbuch-Elor, and
  Cohen-Or]{patashnik2023mixing}
Or Patashnik, Daniel Garibi, Idan Azuri, Hadar Averbuch-Elor, and Daniel
  Cohen-Or.
\newblock Localizing object-level shape variations with text-to-image diffusion
  models.
\newblock In \emph{Proceedings of the IEEE/CVF International Conference on
  Computer Vision (ICCV)}, pages 23051--23061, 2023.

\bibitem[Ping et~al.(2026)Ping, Zhou, Qi, Luo, Bo, and Pang]{dppo2025}
Bowen Ping, Xiangxin Zhou, Penghui Qi, Minnan Luo, Liefeng Bo, and Tianyu Pang.
\newblock {Flow-DPPO}: Divergence proximal policy optimization for flow
  matching models.
\newblock \emph{arXiv preprint arXiv:2606.11025}, 2026.

\bibitem[Podell et~al.(2024)Podell, English, Lacey, Blattmann, Dockhorn,
  M{\"u}ller, Penna, and Rombach]{podell2024sdxl}
Dustin Podell, Zion English, Kyle Lacey, Andreas Blattmann, Tim Dockhorn, Jonas
  M{\"u}ller, Joe Penna, and Robin Rombach.
\newblock {SDXL}: Improving latent diffusion models for high-resolution image
  synthesis.
\newblock In \emph{International Conference on Learning Representations
  (ICLR)}, 2024.

\bibitem[Sclocchi et~al.(2025)Sclocchi, Favero, and Wyart]{sclocchi2025phase}
Antonio Sclocchi, Alessandro Favero, and Matthieu Wyart.
\newblock A phase transition in diffusion models reveals the hierarchical
  nature of data.
\newblock \emph{Proceedings of the National Academy of Sciences}, 122\penalty0
  (1):\penalty0 e2408799121, 2025.

\bibitem[Sepehri et~al.(2026)Sepehri, Mehradfar, Tinaz, Avestimehr, and
  Soltanolkotabi]{sepehri2026athena}
Mohammad~Shahab Sepehri, Asal Mehradfar, Berk Tinaz, Salman Avestimehr, and
  Mahdi Soltanolkotabi.
\newblock {ATHENA}: Adaptive test-time steering for improving count fidelity in
  diffusion models.
\newblock \emph{arXiv preprint arXiv:2603.19676}, 2026.

\bibitem[{Stability AI}(2024)]{sd35}
{Stability AI}.
\newblock {Stable Diffusion 3.5 Medium}.
\newblock \url{https://huggingface.co/stabilityai/stable-diffusion-3.5-medium},
  2024.

\bibitem[Sun et~al.(2026)Sun, Du, Liu, Zheng, and Snoek]{sun2026quota}
Wenfang Sun, Yingjun Du, Gaowen Liu, Yefeng Zheng, and Cees G.~M. Snoek.
\newblock {QUOTA}: Quantifying objects with text-to-image models for any
  domain.
\newblock In \emph{Proceedings of the IEEE/CVF Winter Conference on
  Applications of Computer Vision (WACV)}, 2026.

\bibitem[Testolin et~al.(2025)Testolin, Hou, and Zorzi]{testolin2025visual}
Alberto Testolin, Kuinan Hou, and Marco Zorzi.
\newblock Visual enumeration remains challenging for multimodal generative
  {AI}.
\newblock \emph{PLOS ONE}, 20\penalty0 (9):\penalty0 e0331566, 2025.

\bibitem[Wadhwa et~al.(2026)Wadhwa, Vatsa, Singh, Shah, and
  Galriya]{numbench2026}
Sandeep Wadhwa, Mayank Vatsa, Richa Singh, Parrva~Chirag Shah, and Prakhar
  Galriya.
\newblock {NumBench}: Diagnosing counting failures in text-to-image models.
\newblock \emph{arXiv preprint arXiv:2608.28206}, 2026.

\bibitem[Wallace et~al.(2019)Wallace, Wang, Li, Singh, and
  Gardner]{wallace2019numeracy}
Eric Wallace, Yizhong Wang, Sujian Li, Sameer Singh, and Matt Gardner.
\newblock Do {NLP} models know numbers? {P}robing numeracy in embeddings.
\newblock In \emph{Conference on Empirical Methods in Natural Language
  Processing (EMNLP)}, 2019.

\bibitem[Wang et~al.(2025)Wang, Wan, Belkin, and Wang]{seedsofstructure2025}
Qingsong Wang, Zhengchao Wan, Misha Belkin, and Yusu Wang.
\newblock Seeds of structure: Patch {PCA} reveals universal compositional cues
  in diffusion models.
\newblock In \emph{Advances in Neural Information Processing Systems
  (NeurIPS)}, pages 116960--116993, 2025.

\bibitem[Wu and De~la Torre(2024)]{wu2024contrastive}
Chen Wu and Fernando De~la Torre.
\newblock Contrastive prompts improve disentanglement in text-to-image
  diffusion models.
\newblock \emph{arXiv preprint arXiv:2402.13490}, 2024.

\bibitem[Wu et~al.(2025)]{wu2025qwenimage}
Chenfei Wu et~al.
\newblock Qwen-image technical report.
\newblock \emph{arXiv preprint arXiv:2508.02324}, 2025.

\bibitem[Wu et~al.(2024)Wu, Si, Jiang, Huang, and Liu]{wu2024freeinit}
Tianxing Wu, Chenyang Si, Yuming Jiang, Ziqi Huang, and Ziwei Liu.
\newblock {FreeInit}: Bridging initialization gap in video diffusion models.
\newblock In \emph{Proceedings of the European Conference on Computer Vision
  (ECCV)}, 2024.

\bibitem[Xiang et~al.(2025)Xiang, Sun, Li, Song, Li, Chen, Tang, Hu, and
  Zhang]{xiang2025instanceassemble}
Qiang Xiang, Shuang Sun, Binglei Li, Dejia Song, Huaxia Li, Nemo Chen, Xu Tang,
  Yao Hu, and Junping Zhang.
\newblock {InstanceAssemble}: Layout-aware image generation via instance
  assembling attention.
\newblock In \emph{Advances in Neural Information Processing Systems
  (NeurIPS)}, 2025.

\bibitem[Xu et~al.(2025)Xu, Zhang, and Shi]{xu2025goodseed}
Katherine Xu, Lingzhi Zhang, and Jianbo Shi.
\newblock Good seed makes a good crop: Discovering secret seeds in
  text-to-image diffusion models.
\newblock In \emph{Proceedings of the IEEE/CVF Winter Conference on
  Applications of Computer Vision (WACV)}, 2025.

\bibitem[Yang et~al.(2025)]{qwen3}
An Yang et~al.
\newblock Qwen3 technical report.
\newblock \emph{arXiv preprint arXiv:2505.09388}, 2025.

\bibitem[Yi et~al.(2024)Yi, Li, Xin, and Li]{yi2024mechanism}
Mingyang Yi, Aoxue Li, Yi Xin, and Zhenguo Li.
\newblock Towards understanding the working mechanism of text-to-image
  diffusion model.
\newblock In \emph{Advances in Neural Information Processing Systems
  (NeurIPS)}, 2024.

\bibitem[Zafar et~al.(2026)Zafar, Cohen, Wolf, and Schwartz]{zafar2026ioco}
Oz Zafar, Yuval Cohen, Lior Wolf, and Idan Schwartz.
\newblock Detection-driven object count optimization for text-to-image
  diffusion models.
\newblock In \emph{Proceedings of the IEEE/CVF Winter Conference on
  Applications of Computer Vision (WACV)}, pages 1885--1894, 2026.

\bibitem[Zeng et~al.(2025)Zeng, Zhang, Wang, Xu, Chen, Li, and
  Tu]{yolocount2025}
Guanning Zeng, Xiang Zhang, Zirui Wang, Haiyang Xu, Zeyuan Chen, Bingnan Li,
  and Zhuowen Tu.
\newblock {YOLO-Count}: Differentiable object counting for text-to-image
  generation.
\newblock In \emph{Proceedings of the IEEE/CVF International Conference on
  Computer Vision (ICCV)}, pages 16765--16775, 2025.

\bibitem[Zhao et~al.(2025)Zhao, Wang, Dong, Zhang, and Yuan]{zhao2025noise}
Yaqi Zhao, Xiaochen Wang, Li Dong, Wentao Zhang, and Yuhui Yuan.
\newblock Demystifying numerosity in diffusion models -- limitations and
  remedies.
\newblock \emph{arXiv preprint arXiv:2510.11117}, 2025.

\bibitem[Zheng et~al.(2026)Zheng, Chen, Ye, Wang, Zhang, Jiang, Su, Ermon, Zhu,
  and Liu]{zheng2025diffusionnft}
Kaiwen Zheng, Huayu Chen, Haotian Ye, Haoxiang Wang, Qinsheng Zhang, Kai Jiang,
  Hang Su, Stefano Ermon, Jun Zhu, and Ming-Yu Liu.
\newblock {DiffusionNFT}: Online diffusion reinforcement with forward process.
\newblock In \emph{International Conference on Learning Representations
  (ICLR)}, 2026.

\end{thebibliography}
}

\clearpage
\maketitlesupplementary
\setcounter{section}{0}
\renewcommand{\thesection}{\Alph{section}}
\renewcommand{\theHsection}{supp.\Alph{section}}

\section{Design and measures}
\label{sec:supp_design}
\subsection{The tests}
\label{sec:supp_tests}
\para{Starts.} Instead of pure noise, a rectified-flow sampler can start at noise level $\sigma$ from
\begin{equation}
z_\sigma = (1-\sigma)\,E(y) + \sigma\,\epsilon_{js},
\label{eq:start}
\end{equation}
with $E$ the model's encoder, $y$ an image and $\epsilon_{js}$ the type and seed's own noise; the remaining steps run unchanged. Planted starts of \sdm use $\sigma{=}0.909$ (step 9 of 40) unless stated.

\para{Same noise.} A type and a seed fix the starting noise, and sampling is deterministic: runs that repeat cells of earlier runs (the sweep's agreeing cells, the content test's own cut-outs, the probes' trajectories) reproduce those images exactly. \Cref{tab:supp_design} lists the tests.

\para{Models and settings.} \sdm: 40 steps, guidance 4.5, $768^2$ unless stated; \sdxl base 1.0: 40 Euler steps, guidance 5, $1024^2$; \flux: 28 steps, distilled guidance 3.5, $768^2$; \fluxk base 4B: 50 steps, guidance 4.0, $1024^2$ (\cref{sec:supp_fluxk}); \qwen: the pipeline's defaults, 50 steps, guidance 4.0, $1024^2$, with the transformer in 4-bit NF4 (\cref{sec:supp_qwen}). Prompts give the number in words (``a photo of twelve cribs'').

\begin{table*}[h]
\centering
\footnotesize
\setlength{\tabcolsep}{3.5pt}
\resizebox{\textwidth}{!}{%
\begin{tabular}{@{}llllll@{}}
\toprule
test & model & types & seeds & what changes, from the same noise & start $\sigma$ \\
\midrule
word (\cref{sec:supp_ask}) & \sdm $512^2$ / $1024^2$ & S1 & 3 / 1 & the request, 2--30 & 1 \\
 & \sdm $768^2$ / $1024^2$ & U2 & 2 & the request, 2--30 & 1 \\
 & \sdxl, \flux & S2 & 1 & the request, 2--30 & 1 \\
 & \fluxk $1024^2$ & U2 & 1 & the request, 2--30 & 1 \\
 & \qwen $1024^2$ & 16 of U2 & 1 & the request, 2--30 & 1 \\
asking differently (\cref{sec:supp_wording}) & \sdm & U2 & 1--4 & digits, layout sentence, guidance, best of four & 1 \\
room (\cref{sec:supp_room}) & \sdm & U2 & 1 & a size cue in the prompt (``seen from far away'' or ``tiny''), the request 2--30 & 1 \\
small-count rewards (\cref{sec:supp_reward}) & four \sdm checkpoints & S1 & 1--3 & the request, 2--30 & 1 \\
four changes (\cref{sec:supp_fourarm}) & \sdm / \flux & U2 / 32 of U2 & 1--2 & a new number, a new seed, the number in digits, or ``photo'' to ``photograph'' & 1 \\
four changes, two more models (\cref{sec:supp_fourarm2}) & \fluxk / \qwen & 32 / 16 of U2 & 1--2 & as above, at 4, 8, 12 and 20, 25, 30 / at 20, 25, 30 & 1 \\
fine-scale redraw (\cref{sec:supp_brs}) & \sdm / \fluxk / \qwen & U2 / 16 of U2 / 16 of U2 & 1--2 & the starting noise above $f$ cycles per image taken from seed 2 ($f$ = 2, 4, 8; \sdm also 16 on 16 types; \qwen 4 only) & 1 \\
text streams (\cref{sec:supp_stream}) & \sdm / \flux & U2 / 32 of U2 & 2 & the number given to each text stream, 2 vs.\ 4 or 12 vs.\ 25 & 1 \\
dropped encoder (\cref{sec:supp_stream}) & \sdm & U2 & 1 & one text encoder left empty; the request, 2--30 & 1 \\
borrowed direction (\cref{sec:supp_borrow}) & \sdm & U2 & 1 & the small pair's velocity difference, added to a request for 12 or 25 while $\sigma\geq0.909$ & 1 \\
earlier starts (\cref{sec:supp_sweep}) & \sdm & U2 & 1 & 2 vs.\ 4, 12 vs.\ 25 against a plant of the other & .991--.909 \\
own images (\cref{sec:supp_renoise}) & \sdm & U2 & 1 & 2 vs.\ 4, 12 vs.\ 25, from the model's own image & .991--.813 \\
between (\cref{sec:supp_straddle}) & \sdm & U2 & 1--2 & 2 vs.\ 4 from 3 objects, 12 vs.\ 25 from 16--20 & .991--.813 \\
amplified word (\cref{sec:supp_steer}) & \sdm & U2 & 1 & the difference between two prompts, amplified while $\sigma\geq0.909$ & 1 \\
timing (\cref{sec:supp_timing}) & \sdm & U2 & 1 & the other request of the pair during one window of five steps (0--34); amplified arm: the asked word's contrast inside one window (10--24); the base number respelled in digits (0--9) & 1 \\
timing, two more models (\cref{sec:supp_xtime}) & \sdxl / \fluxk & U2 & 1 & the other request of the pair during one window of five of 40 / six of 50 steps; the base number respelled in digits; the other request from the first step to step 10 or 20 (12 or 24) & 1 \\
word, \sdxl on U2 (\cref{sec:supp_ask}) & \sdxl & U2 & 1 & the request, 2--30 (the timing test's baselines) & 1 \\
word, \flux on U2 (\cref{sec:supp_ask}) & \flux & 16 of U2 & 1 & the request, 2--30 & 1 \\
training (\cref{sec:supp_ceiltrain}) & \sdm + LoRA & U2 & 1 & the weights: a LoRA trained on counted crowds or on shuffled counts; the request, 2--30 & 1 \\
conflict (\cref{sec:supp_conflict}) & \sdm / \sdxl / \flux & S1, S2 / S2 / S2 & 2 / 1 / 1 & request $M$ vs.\ a plant of $N{\in}\{12,20,30\}$ & .909 / .881 / .933 \\
object (\cref{sec:supp_conflict}) & \sdm & S1 & 2 & another object named vs.\ a plant of 20 & .909 \\
self-planting (\cref{sec:supp_method}) & \sdm & U1--U3 & 2 & the start: a grid of the model's own cut-out & .909 \\
layout models (\cref{sec:supp_gligen}) & GLIGEN, InstanceAssemble & U2 & 2 & the model, given the grid as boxes & 1 \\
content (\cref{sec:supp_content}) & \sdm & U2 & 1 & the grid's content: own cut-out, another's, ellipses & .909, .943 \\
planted start (\cref{sec:supp_fluxk_plant}) & \fluxk & U2 & 1 & the start: the content test's own-cut-out grid & .941 \\
noise re-layout (\cref{sec:supp_gu}) & \sdxl (Gu~\etal) & U2 & 1 & the method, 4--30, against its first pass & 1 \\
\bottomrule
\end{tabular}}
\caption{\textbf{The tests in detail.} Every comparison within a model shares object type and starting noise, and only the listed factor changes. \sdm runs at $768^2$ unless stated (reward checkpoints: $512^2$, one at $1024^2$); plants and layouts hold 12--30 objects unless stated. Object sets: \cref{sec:supp_sets}. Every evaluation image is counted by \samt, \gdino and \owl. People counted 16 types of the word test for \sdm at $512^2$ (S1) and $768^2$ (U2), \sdxl and \flux (S2 and U2), \fluxk and \qwen (U2), and of self-planting, the layout models, \fluxk's planted start, the noise re-layout, the amplified word, the timing test, both trained words, the comparison of a new number with a new seed in \sdm and \flux (U2), the four changes in \fluxk and \qwen at the edge of their reach (16 types of U2; seventh count), and the fine-scale redraws above four and eight cycles in \sdm and \fluxk (16 types of U2; eighth count).}
\label{tab:supp_design}
\end{table*}

\subsection{Object sets}
\label{sec:supp_sets}
The object types form five sets of LVIS category names, with no type in two sets (\cref{sec:supp_nouns}). Three are unscreened, with 64 types each: U1, U2 and U3. Two are screened so that \samt sees the object, with 48 types each: S1 and S2. U2 is the set of the main text.

\subsection{Counters}
\label{sec:supp_counters}
\samt, \gdino and \owl count every evaluation image; the consensus is their two-of-three majority and leaves images without a majority uncounted. Near-blank images (at most 2.1\% of a run) are counted like any other image. Against people, \samt undercounts generated crowds (\cref{tab:supp_hc_agree}), so models are compared by people's counts.

\para{Clean composites.} On S2's clean grids, \samt is exact in 0.90, 0.83 and 0.83 of composites at 12, 20 and 30 objects, \gdino in 0.85, 0.75 and 0.40, and \owl in 0.69, 0.60 and 0.40; on S1's, in 0.83, 0.83 and 0.81, 0.73, 0.58 and 0.44, and 0.65, 0.46 and 0.27. On the clean grids at 12--30, the ranges are 0.79--0.88, 0.47--0.77 and 0.35--0.58 on U1, 0.76--0.79, 0.45--0.78 and 0.36--0.59 on U2, and 0.66--0.77, 0.39--0.71 and 0.34--0.59 on U3. Most of \samt's misses are empty readings: it counts zero on 16\%, 20\% and 23\% of the unscreened sets' composites (no box at all on 11\%, 16\% and 16\%), concentrated in a few object types. \owl, which counts only 36--59\% of U2's clean composites exactly, gives the grid a smaller advantage over the scatter at 12, 20 and 30 (0.10) and the word a larger exact lift at native resolution (0.13 of its small-count value).

\subsection{Measures}
\label{sec:supp_measures}
\emph{Order}: over all pairs of requests $c<c'$ in the range from the same noise, the share in which $n(c')>n(c)$, ties excluded (\cref{eq:order}); an order below one half is a reversal. \emph{Slope}: for each object type and seed, the least-squares slope of $\log(1{+}n)$ on $\log c$ over the requests in the range, averaged over cells with at least three requests. \emph{Exact lift} over a range $\mathcal{A}$:
\begin{equation*}
L(\mathcal{A}) = \mathop{\mathbb{E}}_{j,s,\,c\in\mathcal{A}} \Big[\mathbf{1}[n_{js}(c){=}c] - \tfrac{1}{|\mathcal{A}|-1}\!\!\sum_{r\in\mathcal{A}\setminus\{c\}}\!\!\mathbf{1}[n_{js}(r){=}c]\Big],
\end{equation*}
how much more often the model draws exactly $c$ when asked for $c$ than when asked for another count of the range, from the same noise; it is bounded by the exact rate. Intervals: 95\% bootstrap over object types (20{,}000 resamples, the same resampled indices for all quantities of a test) unless stated.

\subsection{Reference models}
\label{sec:supp_refmodels}
\para{The loose estimator.} The reference model draws counts as $\mathrm{round}(c\,e^{sz})$ with $z\sim\mathcal{N}(0,1)$, independently for each request, type and seed, and $s\in\{0.3,0.45\}$: a model that follows the request as a loose estimate, which the main text calls a loose estimator. Simulated for 48 object types and three seeds and summarized with the same estimators, it gives the last row of \cref{tab:supp_ask}. For the comparison of a new number with a new seed (\cref{sec:supp_reseed}; 400 simulations of 64 types, $s=0.30$ or $0.45$) it has ratios of 2.78 and 1.96 and orders of 0.98 and 0.93 at 2, 4, 8; 2.19 and 1.62 (0.93, 0.86) at 2, 3, 4, 6, 8; 1.94 and 1.47 (0.91, 0.82) at 12, 20, 30; and 1.45 and 1.22 (0.79, 0.71) over the nine requests from 12 to 30. A model whose count does not depend on the request has a ratio of 1 and an order of 0.5.

\para{Within-type scatter.} Because a type's shared offset cancels in same-noise pairs, the order of an estimate depends only on its scatter within a type; the loose estimator's band therefore describes within-type scatter, and an estimator with errors correlated across requests reaches the same order with a larger total scatter, and lower exact and within-10\% rates.

\para{The loose estimator as the counters read it.} People counted all 32 composites of 12--30 known objects exactly, so on clean images they would read a loose estimator as it is (their disagreement on dense generated crowds lowers its order a little, \cref{sec:supp_noise}); on 16 types, as in the people-counted sets, its order by people ranges over 0.77--0.89 (scatter 0.30) and 0.68--0.83 (0.45) in 95\% of 4{,}000 simulated sets. Read through \samt's errors on the people-counted pictures (each simulated type takes one real type's \samt errors, $\log(1{+}n_{\textsc{sam}})-\log(1{+}h)$, at the five requests, so that a type \samt misses is missed at every request), its order falls to 0.74 \ci{0.63}{0.83} and 0.69 \ci{0.59}{0.79}, and its slope spreads from 0.2 to 1.6. \samt's orders thus understate an estimate's by about 0.06--0.10, which is why models are ranked by people's counts.

\section{People's counts}
\label{sec:supp_human}
\subsection{Protocol}
\label{sec:supp_hcprotocol}
\para{Task.} CloudResearch Connect, one picture per task (\$0.10; under a minute), desktop only, at most 40 pictures per person and round. A task shows the picture ($1024^2$, re-encoded, under a random file name), the object's plural, name and LVIS definition, and short instructions with a worked example. People click once on each object of the named kind (numbered dots; a click on a dot removes it) or press ``There are none'', may zoom up to three times and may tick ``Unsure''. After the first dot they answer ``Do the things you marked look like $X$s?'' (yes, clearly; partly, or hard to tell; no, they look like something else). They never see the request, the model, the method or the stratum. A pilot of 40 pictures from outside the sample confirmed the pay and led to the look-alike question and to harder checks, because people counted some GLIGEN box fills as the named object although they did not look like it. A picture is \emph{recognisable} when its human count equals the request and the majority look-alike answer is ``yes, clearly''.

\para{Rounds and checks.} R1: every picture once, in random order, mixed with checks (clean composites of known count from types outside the sample, indistinguishable from the rest). R2: the same tasks in another order, with every R1 participant excluded. R3: every picture whose R1 and R2 counts differ, with further checks, and every R1 and R2 participant excluded. The human count is the R1 count where R1 and R2 agree, and otherwise the median of the three; the look-alike judgement is the majority of the answers behind the human count (no majority: ``partly''). All of a person's answers in a round are set aside if more than half of at least two checks are off by more than two, or if their median time is under 3\,s; set-aside pictures are counted again in a top-up of the same round, open only to new people. The eight studies collected no personal data; they were run as paid annotation of generated images (\$0.10 per picture, about \$12--38 per hour) and were not submitted for IRB review.

\subsection{The eight studies}
\label{sec:supp_hcstudies}
\para{First count.} A1: \sdm $512^2$ plain images, 16 of S1's 48 types, requests 2, 4, 6, 8, 12, 16, 20, 25 and 30, seed 1 (144 images); A2: \sdxl, 16 of S2's types, the same requests (144); A3: \flux, the same 16 types, the same requests without 25 (128); A4: \sdm $768^2$ plain images of 16 of the 58 types of U2 that have a cut-out, requests as A1 (144); B: self-planted images of the same 16 types at the set's nine counts (144); C: GLIGEN (grounded at every step) at 12, 16, 20, 25 and 30 (80); D: InstanceAssemble at the nine counts (144); E: 32 self-planted images of U1 and U3 that \samt counted below half the asked count; F: 32 clean composites of known count from other types. 992 pictures; A1--A4 hold every request of each type and seed, so the summaries of \cref{sec:setup} can be computed from human counts, and A4, B, C and D share their types and counts. R1 held the 992 pictures and 80 checks (clean composites of known count 12--25), R3 20 further checks. One R1 participant was set aside (three of five checks off by more than two); their 27 pictures were counted again. Nobody took part in two rounds.

\para{Second count.} N, \fluxk's plain images of the same 16 U2 types at all nine requests (144); P, \fluxk from the planted grids at 12, 16, 20, 25 and 30 (80); G and O, the noise re-layout's output and its own first pass at the same counts (80 each); T, \sdm with the word amplified at $L{=}10$ (80). With them came 30 checks drawn from the first study's. That makes 464 pictures besides the checks, with 65 people in R1, 49 in R2 and 29 in R3 (180 pictures). Nobody took part in two rounds or was set aside, and 90--97\% of single answers on the checks were exact in each round. T shares types, counts and noise with the first study's A4 and B, and P shares types, counts and grids with B, so both are compared with them cell by cell.

\para{Third count.} The timing test's swap images of the 16 people-counted types at windows 0--4, 10--14 and 20--24, both pairs (96 pictures), with 20 checks drawn from the first study's. 41 people in R1, 39 in R2 and 17 in R3 (the 38 pictures whose first two counts differ, with the 20 checks); nobody took part in two rounds or was set aside, no picture failed to load, and 90--95\% of single answers on the checks were exact in each round. Its baselines are the first study's human counts of the same plain images (stratum A4). Participants of earlier studies were not excluded, as the analysed pictures were new (the 20 checks came from the first study).

\para{Fourth count.} \qwen's plain images of the same 16 U2 types at all nine requests (stratum Q, 144 pictures), with 20 checks drawn from the first study's; its baselines are the first study's A4 (\sdm) and the second's N (\fluxk), of the same types and requests. 37 people took part in R1, 34 in R2 and 7 in R3 (the 26 pictures whose first two counts differ, with the 20 checks); one R2 participant was set aside by the check criterion, and their 16 pictures were counted again in a top-up of R2 open only to new people (with one check at each known count). Nobody took part in two rounds, no picture failed to load, and 83--100\% of single answers on the checks were exact in each round (83\% in R2, counting the participant set aside). The first two counts agree on 82\% of the pictures (89\% at 2--8, 66\% at 25--30), against 57\% in the first three studies.

\para{Fifth count.} X, \sdxl's plain images of the same 16 U2 types at all nine requests (the timing test's baselines; 144 pictures); LC and LS, \sdm with the count and the shuffled LoRA of \cref{sec:supp_ceiltrain} (144 each); FD, \flux's plain images of the same types and requests, drawn with the sampler of its word test (144). Two batches (X, LC, LS; FD), each with 20 checks drawn from the first study's. 91, 72 and 31 people took part in the three rounds of the first batch and 43, 35 and 9 in the second; nobody took part in two rounds of a batch or was set aside, no picture failed to load, and 95--100\% of single answers on the checks were exact in each round. The first two counts agree on 65\% of the pictures (24\% of \sdxl's at 12--30). Its baselines are the first study's A4 (\sdm), the second's N (\fluxk) and the fourth's Q (\qwen), of the same types and requests.

\para{Sixth count.} On the 16 people-counted U2 types: \sdm's plain images of seeds 1 and 2 at 2, 4, 8, 12, 20 and 30, and the number in digits (seed 1) at 12, 20 and 30 (240 pictures; the 96 of seed 1 are those of stratum A4 of the first count, counted again), with 20 checks per round. 31, 30 and 12 people took part in R1, R2 and R3 (the 84 pictures whose first two counts differ); nobody took part in two rounds or was set aside, no picture failed to load, and 95--100\% of the checks were exact. The first two counts agree on 65\% of the pictures (seeds 1 and 2: 83\% and 77\% at 2--8, 52\% and 60\% at 12--30; digits 52\%), and the human count of a seed-1 picture equals the first study's count of the same picture in 71\% \ci{59}{82}. A second batch, under the same protocol, counted \flux's plain images of the same 16 types at seeds 1 and 2, at requests 2, 4, 12 and 25 (the plain cells of the stream test; \cref{sec:supp_fluxreseed}).

\para{Seventh count.} The four changes past the reach in \fluxk (batch K) and \qwen (batch Q) on the 16 people-counted U2 types at 20, 25 and 30: the plain image of seed 1 (the word test's, counted again; strata N and Q of the second and fourth counts), the plain image of seed 2, the number in digits and the edit ``a photograph of'' (192 pictures per model, 288 of them new), each batch with 20 checks drawn from the first study's. Batch K: 74, 45 and 24 people took part in R1, R2 and R3 (the 96 pictures whose first two counts differ); 90--95\% of the checks were exact in each round, and the first two counts agree on 50\% of the pictures. Batch Q: 54, 42 and 10 people took part in R1, R2 and R3 (the 39 pictures whose first two counts differ, and two whose R2 answer was still outstanding); 90--100\% of the checks were exact in each round, and the first two counts agree on 80\% of the pictures. Nobody took part in two rounds of a batch or was set aside, and no picture failed to load (\cref{sec:supp_fourarm2}).

\para{Eighth count.} The fine-scale redraws above four and eight cycles on the 16 people-counted U2 types: \sdm's at 4, 8, 12, 20 and 30 (batch S, 160 pictures) and \fluxk's at 4, 8, 12, 20, 25 and 30 (batch K, 192), each batch with 20 checks drawn from the first study's. Batch S: 84, 72 and 38 people took part in R1, R2 and R3 (the 72 pictures whose first two counts differ). One picture failed to load in R2 and was counted again in a top-up of R2, where one person answered only checks, all three off by more than two, and was set aside; in R3 one person was set aside by the same rule, and their four pictures were counted again in a top-up of R3. The first two counts agree on 55\% of the pictures. Batch K: 106, 86 and 34 people took part in R1, R2 and R3 (the 63 pictures whose first two counts differ, and one whose R2 answer was still outstanding); nobody was set aside and no picture failed to load, and the first two counts agree on 67\% of the pictures. In the main rounds of both batches 85--100\% of the checks were exact, and nobody took part in two rounds of a batch, top-ups included (\cref{sec:supp_brs}).

\subsection{Agreement}
\label{sec:supp_hcagree}
Single answers disagree most on dense crowds and on occasional misreadings of the object; the third count settles every picture where the first two differ (\cref{tab:supp_hc_rel,tab:supp_hc_agree}).

\begin{table}[h]
\centering
\footnotesize
\setlength{\tabcolsep}{2.6pt}
\begin{tabular}{@{}llccccc@{}}
\toprule
study & requests & pictures & exact & within 10\% & Spearman & ICC \\
\midrule
first & all & 960 & .55 & .61 & .59 & .52 \ci{.45}{.59} \\
 & 2--8 & 256 & .72 & .73 & .74 & .60 \ci{.43}{.75} \\
 & 12--30 & 704 & .49 & .56 & .54 & .47 \ci{.39}{.55} \\
second & all & 464 & .61 & .68 & .77 & .73 \ci{.65}{.81} \\
 & 2--8 & 64 & .86 & .86 & .72 & .67 \ci{.32}{.92} \\
 & 12--30 & 400 & .57 & .65 & .76 & .72 \ci{.64}{.80} \\
third & all & 96 & .60 & .66 & .63 & .60 \ci{.35}{.80} \\
 & 2--8 & 48 & .73 & .73 & .63 & .44 \ci{.08}{.82} \\
 & 12--30 & 48 & .48 & .58 & .53 & .54 \ci{.23}{.82} \\
all three & all & 1520 & .57 & .63 & .65 & .60 \ci{.55}{.65} \\
 & 2--8 & 368 & .74 & .75 & .74 & .62 \ci{.48}{.74} \\
 & 12--30 & 1152 & .52 & .59 & .62 & .56 \ci{.50}{.62} \\
\bottomrule
\end{tabular}
\caption{\textbf{Agreement of the two first counts} (single answers of R1 and R2, different people; generated pictures only, without the first study's 32 clean composites). Exact: equal counts; within 10\%: differ by at most 10\% of the larger; ICC(A,1) on $\log(1{+}n)$ with a bootstrap interval over pictures. Third study: requests 2 and 4 (``2--8'') and 12 and 25. Single answers disagree most on dense crowds and on occasional misreadings of the object (a 0 against a 20), which lower the ICC; the analysed count is settled by a third count wherever the first two differ.}
\label{tab:supp_hc_rel}
\end{table}

\begin{table}[h]
\centering
\footnotesize
\setlength{\tabcolsep}{2.6pt}
\resizebox{\linewidth}{!}{%
\begin{tabular}{@{}lccccc@{}}
\toprule
& & \multicolumn{2}{c}{R1 = R2} & \multicolumn{2}{c}{\samt = people} \\
\cmidrule(lr){3-4}\cmidrule(lr){5-6}
stratum & pictures & all & 12--30 & exact & bias \\
\midrule
\sdm $512^2$ plain & 144 & .67 & .55 & .67 & $-$.8 \\
\sdxl plain & 144 & .32 & .13 & .31 & $-$1.7 \\
\flux plain & 128 & .66 & .53 & .55 & $-$2.0 \\
\sdm $768^2$ plain$^\ast$ & 144 & .55 & .44 & .54 & $-$.9 \\
self-planting$^\ast$ & 144 & .63 & .63 & .51 & $-$2.7 \\
GLIGEN$^\ast$ & 80 & .46 & .46 & .48 & $-$8.1 \\
InstanceAssemble$^\ast$ & 144 & .55 & .55 & .57 & $-$4.8 \\
apparent collapses & 32 & .47 & .47 & .19 & $-$10.5 \\
clean composites & 32 & .94 & .94 & -- & -- \\
\midrule
\multicolumn{6}{@{}l}{\emph{second count}} \\
\fluxk plain$^\ast$ & 144 & .69 & .55 & .56 & $-$3.2 \\
\fluxk planted$^\ast$ & 80 & .84 & .84 & .75 & $-$3.2 \\
\sdxl first pass (Gu~\etal)$^\ast$ & 80 & .48 & .48 & .51 & $-$6.3 \\
noise re-layout$^\ast$ & 80 & .44 & .44 & .46 & $-$1.2 \\
\sdm, word amplified$^\ast$ & 80 & .56 & .56 & .51 & $-$2.9 \\
\midrule
\multicolumn{6}{@{}l}{\emph{fifth count}} \\
\sdxl plain$^\ast$ & 144 & .33 & .24 & .35 & $-$.8 \\
\flux plain$^\ast$ & 144 & .75 & .66 & .65 & $-$1.0 \\
\sdm, count LoRA$^\ast$ & 144 & .78 & .74 & .73 & $-$1.3 \\
\sdm, shuffled LoRA$^\ast$ & 144 & .72 & .65 & .64 & $-$4.9 \\
\bottomrule
\end{tabular}}
\caption{\textbf{Agreement.} $^\ast$U2. R1 = R2: the two first counts agree exactly (all requests; 12--30 only). \samt = people: \samt's count equals the human count; bias: mean \samt minus human count. People: R1 77, R2 78, R3 55 (top-ups included), and in the second count 65, 49 and 29; single answers on clean composites .969 exact, final human count 1.00 (32 pictures). Of 32 self-planted images that \samt counted below half of the request, people count 15 below half. At the five counts whose grids have complete rows (12, 16, 20, 25, 30), the two first counts agree on 59\% of the self-planted pictures (47 of 80) and on 44\% of the plain ones of the same types (35 of 80).}
\label{tab:supp_hc_agree}
\end{table}

\subsection{People's counting noise and a noise-matched reference}
\label{sec:supp_noise}
\para{A model of the noise.} A picture has a true count $T$, what careful counting finds, and is easy or hard to count: hard with probability $\mathrm{logistic}(g+0.76\,x+u)$, where $x=\log(1{+}T)-\log 11$, $g$ is the stratum's difficulty and $u\sim\mathcal{N}(0,1.28^2)$ the object type's, shared across strata. Each single count $r$, by a different person, is independently exact, zero (the object misread), off by one, or further off, with $\log(1{+}r)-\log(1{+}T)$ a Student-$t$ with 0.60 degrees of freedom (a sharp centre and very heavy tails). At $T{=}12$, an easy picture's single count is exact with probability 0.85 and zero with 0.06; a hard picture's is exact with 0.21, off by one with 0.33 and further off with 0.43. For a type of median difficulty, a picture of twelve objects is hard with probability 0.86 in \sdxl, 0.60 in \sdm, 0.40 in \flux, 0.11 in \fluxk and 0.02 in \qwen. The final count is the median of three single counts, as in the protocol (where the first two agree, they fix the median). The 34 parameters and a nonparametric prior on $T$ (0--180) are fitted by maximum likelihood to the 5{,}360 single counts of the 2{,}240 generated pictures of the first five counts (19 strata); simpler versions fit worse (AIC $+$575 with one class, $+$316 with one difficulty for all strata, $+$81 without the type effect).

\para{Checks.} Where the first two counts differ (880 pictures), the model predicts where the third falls: on one of them in 0.497 of pictures (observed 0.507), between them in 0.167 (0.160), below both in 0.157 (0.152), above both in 0.180 (0.181). Recounted by new people in the sixth count, \sdm's 96 pictures of A4 keep their final count in 0.71 \ci{0.59}{0.82} of cases (predicted 0.69), and the two final counts differ by a mean $|\Delta\log(1{+}n)|$ of 0.10 \ci{0.05}{0.17}, 0.12 \ci{0.05}{0.21} on the 48 crowds of 12--30 (0.65 of them unchanged) and 0.12 \ci{0.05}{0.20} on the 32 pictures of four and eight objects; \flux's 64 recounted pictures keep theirs in 0.875 \ci{0.80}{0.95} of cases (predicted 0.870). The model reproduces how often the first two counts agree to within 0.04 in eleven strata, A1--A4, N and X among them; it is noisier than the data for \qwen and \flux (agreement 0.73 and 0.66, against 0.82 and 0.75 observed) and has too few large disagreements on \sdxl's crowds at 12--30 (first two counts apart by a factor of 1.5 or more, zeros excluded: 0.24 of pictures, against 0.32 observed), which the single-count bound below covers. On the 32 clean composites of 12--30 objects, 64 of 66 single answers and all 32 final counts are exact: the noise belongs to generated crowds, not to counting itself. On pictures of 12--30 objects the final count is exact in 42--94\% and within 10\% in 78--96\% of pictures, depending on the model's crowds (a single count is exact in 30--82\%), with a mean $|\Delta\log(1{+}n)|$ of 0.04--0.11: its errors are mostly a few objects.

\para{The loose estimator, counted by people.} We simulate the loose estimator of \cref{sec:supp_refmodels} on the people-counted design (16 types, one seed, the nine requests; 4{,}000 sets), take each of its counts as a picture's true count and replace it by a final count drawn from this model, with the counting difficulty of the compared model's own pictures (\emph{matched}; a type's difficulty is shared by its requests and seeds), and summarize each set with the models' estimators; a reference's brackets give the 2.5th and 97.5th percentiles of its sets. So counted, with the difficulty of the word test's pictures pooled, its order at 12--30 falls only from 0.835 to 0.810 at $s{=}0.30$ and from 0.755 to 0.739 at $s{=}0.45$ (\ci{0.73}{0.87} and \ci{0.65}{0.82}); with each model's own difficulty it is the same within 0.01, and so under every fitted variant of the noise model (0.806--0.816 and 0.736--0.744). Its slope falls from 0.945 to 0.91, and its exact rate at 12--30 from 7.2\% to 6.8\% and from 4.7\% to 4.6\%. Its exact lift is 4.0 and 1.8 points at 12--30, and 2.7 \ci{$-$3.1}{9.9} and 1.2 \ci{$-$3.6}{7.3} points at 20--30. On 16 types and two seeds, a new number moves its count 1.70 \ci{1.32}{2.19} and 1.39 \ci{1.11}{1.77} times as far as a new seed among 12, 20 and 30 (1.96 and 1.48 without the noise; \sdm's difficulty), with an order of the number change of 0.88 and 0.80, and 1.93 and 1.51 times between 12 and 25 (\flux's difficulty). Counted once by one person instead, a deliberately generous bound far noisier than the final counts, it orders at 0.750 and 0.696 with a slope of 0.83, and its ratio among 12, 20 and 30 is 1.32 and 1.21. Read through \samt's errors (a model of \samt's misses and errors fitted against people's final counts, or each simulated type borrowing one real type's errors as in \cref{sec:supp_refmodels}), its order at 12--30 is 0.64--0.73 and its number/seed ratio among 12, 20 and 30 1.05--1.20: \samt's ratio for \sdm over the nine requests still lies below it (0.89 \ci{0.80}{0.99}, against 1.05 \ci{0.98}{1.12}), the ratios of the four changes in \sdm and \flux do not (1.03--1.59, \cref{sec:supp_fourarm}), and their \samt orders of the number change past ten (0.41--0.53) do. At 20--30 (\fluxk and \qwen, \cref{sec:supp_fourarm2}) the estimator read by \samt orders at 0.59--0.60 with a ratio of 1.03--1.04; \fluxk's \samt order lies below it, \qwen's (0.65) inside it.

\begin{table*}[t]
\centering
\footnotesize
\setlength{\tabcolsep}{4pt}
\begin{tabular}{@{}llcccc@{}}
\toprule
& & & \multicolumn{3}{c}{loose estimator, counted like the model} \\
\cmidrule(l){4-6}
model & measure & people & $s{=}0.45$, matched & $s{=}0.30$, matched & $s{=}0.45$, one single count \\
\midrule
\flux & order 12--30 & .36 \ci{.23}{.49} & .74 \ci{.65}{.82}$^\ast$ & .81 \ci{.74}{.88}$^\ast$ & .70 \ci{.60}{.78}$^\ast$ \\
 & slope 12--30 & $-$.36 \ci{$-$.75}{$-$.02} & .91 \ci{.52}{1.27}$^\ast$ & .91 \ci{.60}{1.19}$^\ast$ & .83 \ci{.22}{1.38}$^\ast$ \\
 & lift 12--30 & $-$1.2 \ci{$-$2.5}{$-$0.3} & 1.8 \ci{$-$2.8}{7.2} & 4.0 \ci{$-$1.2}{10.3} & 1.6 \ci{$-$2.8}{6.6} \\
 & lift 20--30 & 0.0 \ci{0.0}{0.0} & 1.2 \ci{$-$3.6}{7.3} & 2.7 \ci{$-$3.1}{10.4} & 1.0 \ci{$-$3.1}{7.3} \\
\midrule
\sdm & order 12--30 & .54 \ci{.41}{.65} & .74 \ci{.65}{.82}$^\ast$ & .81 \ci{.73}{.87}$^\ast$ & .70 \ci{.60}{.79}$^\ast$ \\
 & slope 12--30 & .01 \ci{$-$.51}{.47} & .92 \ci{.57}{1.27}$^\ast$ & .91 \ci{.60}{1.19}$^\ast$ & .84 \ci{.26}{1.41}$^\ast$ \\
 & lift 12--30 & 1.9 \ci{$-$0.9}{5.3} & 1.9 \ci{$-$2.8}{7.5} & 3.9 \ci{$-$1.6}{10.3} & 1.6 \ci{$-$2.5}{6.9} \\
 & lift 20--30 & $-$1.0 \ci{$-$2.6}{0.0} & 1.2 \ci{$-$3.6}{7.3} & 2.7 \ci{$-$3.1}{9.9} & 1.0 \ci{$-$3.6}{7.3} \\
\midrule
\sdxl & order 12--30 & .63 \ci{.52}{.74} & .74 \ci{.66}{.82}$^\ast$ & .81 \ci{.74}{.87}$^\ast$ & .69 \ci{.60}{.78} \\
 & slope 12--30 & .42 \ci{.09}{.76} & .92 \ci{.57}{1.26}$^\ast$ & .92 \ci{.63}{1.20}$^\ast$ & .85 \ci{.27}{1.39} \\
 & lift 12--30 & 0.3 \ci{$-$2.5}{3.4} & 1.7 \ci{$-$2.8}{7.2} & 3.8 \ci{$-$1.2}{10.3} & 1.4 \ci{$-$2.8}{6.6} \\
 & lift 20--30 & 2.6 \ci{$-$1.0}{7.3} & 1.1 \ci{$-$3.6}{7.3} & 2.6 \ci{$-$3.1}{9.9} & 0.9 \ci{$-$3.1}{6.8} \\
\midrule
\qwen & order 12--30 & .70 \ci{.57}{.82} & .74 \ci{.66}{.82} & .81 \ci{.74}{.88}$^\ast$ & .70 \ci{.60}{.78} \\
 & slope 12--30 & .23 \ci{$-$.01}{.43} & .88 \ci{.47}{1.27}$^\ast$ & .89 \ci{.52}{1.20}$^\ast$ & .78 \ci{.13}{1.39} \\
 & lift 12--30 & 12.2 \ci{3.8}{20.9} & 1.9 \ci{$-$2.5}{7.5} & 4.2 \ci{$-$1.2}{10.6} & 1.7 \ci{$-$2.5}{6.9} \\
 & lift 20--30 & 4.2 \ci{$-$2.6}{14.1} & 1.2 \ci{$-$3.6}{7.3} & 2.8 \ci{$-$3.1}{10.4} & 1.1 \ci{$-$3.6}{7.3} \\
\midrule
\fluxk & order 12--30 & .62 \ci{.53}{.72} & .74 \ci{.65}{.82}$^\ast$ & .81 \ci{.74}{.88}$^\ast$ & .70 \ci{.61}{.78} \\
 & slope 12--30 & .12 \ci{$-$.34}{.54} & .89 \ci{.48}{1.25}$^\ast$ & .89 \ci{.54}{1.20}$^\ast$ & .80 \ci{.18}{1.40}$^\ast$ \\
 & lift 12--30 & 9.1 \ci{3.4}{15.0} & 1.8 \ci{$-$2.5}{7.5} & 4.1 \ci{$-$1.2}{10.3} & 1.6 \ci{$-$2.5}{6.9} \\
 & lift 20--30 & 4.7 \ci{$-$0.5}{10.9} & 1.2 \ci{$-$3.6}{7.8} & 2.7 \ci{$-$3.1}{10.4} & 1.0 \ci{$-$3.6}{7.3} \\
\midrule
\multicolumn{6}{@{}l}{\emph{a new number against a new seed (two seeds, sixth count)}} \\
\sdm & ratio 2, 4, 8 & 2.77 \ci{1.87}{4.09} & 1.71 \ci{1.33}{2.20} & 2.14 \ci{1.62}{2.81} & 1.41 \ci{1.12}{1.81} \\
 & ratio 12, 20, 30 & 0.90 \ci{0.60}{1.29} & 1.39 \ci{1.11}{1.77}$^\ast$ & 1.70 \ci{1.32}{2.19}$^\ast$ & 1.21 \ci{0.99}{1.52}$^\ast$ \\
 & order 12, 20, 30 & .52 \ci{.35}{.69} & .80 \ci{.72}{.88}$^\ast$ & .88 \ci{.81}{.94}$^\ast$ & .75 \ci{.65}{.84}$^\ast$ \\
\flux & ratio 2, 4 & 2.35 \ci{1.51}{4.84} & 1.43 \ci{1.04}{2.00} & 1.73 \ci{1.22}{2.47} & 1.26 \ci{0.93}{1.73} \\
 & ratio 12, 25 & 1.64 \ci{1.13}{2.52} & 1.51 \ci{1.09}{2.08} & 1.93 \ci{1.37}{2.72} & 1.29 \ci{0.94}{1.74} \\
 & order 12, 25 & .66 \ci{.43}{.86} & .86 \ci{.73}{.97}$^\ast$ & .94 \ci{.84}{1.00}$^\ast$ & .80 \ci{.65}{.94} \\
\bottomrule
\end{tabular}
\caption{\textbf{Each model against a loose estimator counted like it} (people's counts of the 16 people-counted U2 types; above, one seed and the nine requests; below, two seeds, \cref{sec:supp_reseedpeople,sec:supp_fluxreseed}). People: the model's value and its 95\% interval over types. Loose estimator: mean and 2.5th--97.5th percentiles of 4{,}000 simulated sets of the same design, its counts read as people's final counts with the counting difficulty of the model's own pictures (matched), or as one person's single count (a generous bound); $^\ast$: the model's value lies below the 2.5th percentile. Lift: exact lift in percentage points; at 20--30, requests 20, 25 and 30, each against the other requests of 12--30. Ratio: number change over seed change; order: of the number change.}
\label{tab:supp_noise}
\end{table*}

\para{Each model against it.} \Cref{tab:supp_noise} sets each model's people's values beside the reference matched to its own pictures. \flux and \sdm lie below it on order and slope at 12--30 at both scatters and under the single-count bound, and \sdm also on the number/seed ratio past its reach (0.90) and on the order of the number change (0.52). \sdxl and \fluxk lie below on order and slope at both scatters, but only \fluxk's slope stays below the single-count bound; \qwen lies below on slope, and on order only at $s{=}0.30$. Between 12 and 25, \flux's ratio (1.64) lies inside the reference's range, and only the order of that change (0.66) lies below it, though not below the single-count bound. No model's exact lift lies below the reference. At 20--30 the reference's own lift is 1.1--2.8 points, with a 95\% range of about $-$3.6 to $+$10 points on 16 types, so lifts of $-$1.0 to $+$4.7 points cannot separate a model from a loose estimator; at 12--30, \fluxk and \qwen lie above the $s{=}0.45$ reference (9.1 and 12.2 points, mostly at twelve), and \qwen above the $s{=}0.30$ one too.

\para{Adjacent requests and an operational reach.} Taking the requests in adjacent pairs (one pair per type; a pair passes when the 2.5\% quantile of its order's type bootstrap, computed exactly, is above 0.5), the largest request whose pair with the next request passes is 6 for \flux, 4 for \sdm, 6 for \sdxl, 16 for \qwen and 8 for \fluxk, the same in only 28--68\% of type resamples, while a loose estimator counted by people stops by this rule at a median of 8 at $s{=}0.45$ and of 16 at $s{=}0.30$: with 16 types, this operational reach separates no model from a loose estimator at $s{=}0.45$ (at $s{=}0.30$ only \sdm's reach of 4 is unusual: 2\% of the reference's sets stop as early). The adjacent orders separate more: from 25 to 30, four models order their crowds below the matched $s{=}0.45$ reference's 2.5th percentile (\flux 0.12, \sdm 0.27, \fluxk 0.33 and \qwen 0.36, against 0.38), and \flux also from 8 to 12 (0.42 against 0.50), five of 40 pair tests where one is expected by chance.

\subsection{Single counts, ties and zeros}
\label{sec:supp_hcrobust}
\para{Orders from single counts.} People's orders at 12--30 do not hinge on the third count: computed from the first count alone and from the second alone, they are 0.72 and 0.70 for \sdxl (final 0.77), 0.68 and 0.67 for \qwen (0.70), 0.66 and 0.63 for \fluxk (0.62), 0.54 and 0.49, 0.48 and 0.55 for \sdm at $512^2$ and $768^2$ (0.49, 0.54), and 0.41 and 0.53 for \flux (0.52). Single counts of crowds are noisy: the two first counts' log difference has an SD of 0.94--1.13 at 12--30 in \sdxl, \fluxk and \qwen and 0.66--0.77 in the others, so a single count's own scatter in $\log(1{+}n)$ is about 0.47--0.80; the final counts, medians of three where the first two differ, are steadier, which is why the main text uses them.

\para{Ties and zeros.} Among same-type request pairs at 12--30, people's final counts tie in 17\% (\sdm $512^2$), 21\% (\sdm $768^2$), 4\% (\sdxl), 30\% (\flux; 96 pairs, no request of 25) and 9\% (\fluxk) of pairs, which the order leaves out; at 2--8 in 16\%, 21\%, 4\%, 5\% and 14\%. People counted no object in 1--9\% of these pictures at 12--30 (\sdm $768^2$: 9\%). \qwen's final counts tie in 23\% of pairs at 12--30 and 4\% at 2--8. Counting a tie as half a pair that follows the request instead of setting it aside, people's orders at 12--30 become 0.49 and 0.53 (\sdm $512^2$, $768^2$), 0.52 (\flux), 0.76 (\sdxl), 0.61 (\fluxk) and 0.66 (\qwen), within 0.05 of those in \cref{tab:supp_hc_ask,tab:supp_qwen}. (\sdxl and \flux here are the first count's, on S2.)

\section{The reach}
\label{sec:supp_ask}
\subsection{The number word, by counter and by people}
\label{sec:supp_wordcount}

\begin{figure*}[h]
\centering
\includegraphics[width=0.9\linewidth]{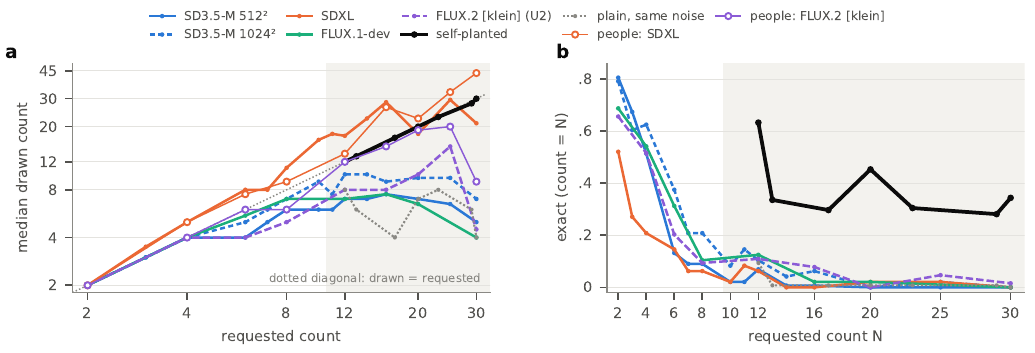}
\caption{\textbf{Drawn and exact counts by requested count, by \samt} (open circles: people). (a) Median drawn count, log scales (dotted: drawn = requested); (b) exact rate. Shaded: requests past ten. Plain prompts in three models on the screened sets (48 types) and in \fluxk on U2 (64); self-planting (\cref{sec:supp_method}) on U1, with the plain prompt from the same noise (\sdm at $768^2$). \samt counts dense crowds lower than people (open circles: \sdxl and \fluxk, 16 types each). \qwen's \samt counts are in \cref{tab:supp_qwen}.}
\label{fig:ask_sam3}
\end{figure*}

\begin{table*}[h]
\centering
\small
\resizebox{\linewidth}{!}{%
\begin{tabular}{@{}llcccc@{}}
\toprule
& & \multicolumn{2}{c}{slope} & \multicolumn{2}{c}{order} \\
\cmidrule(lr){3-4}\cmidrule(lr){5-6}
Model & counter & 2--8 & 12--30 & 2--8 & 12--30 \\
\midrule
\sdm $512^2$ (S1, three seeds) & \samt & .53 \ci{.46}{.59} & $-$.19 \ci{$-$.40}{.03} & .88 & .49 \ci{.42}{.55} \\
 & \gdino & .43 \ci{.36}{.49} & $-$.68 \ci{$-$.84}{$-$.53} & .84 & .31 \ci{.27}{.35} \\
 & \owl & .44 \ci{.37}{.50} & $-$.17 \ci{$-$.38}{.02} & .82 & .48 \ci{.42}{.54} \\
\midrule
\sdm $1024^2$ (S1) & \samt & .61 \ci{.52}{.69} & $-$.38 \ci{$-$.75}{$-$.02} & .90 & .53 \ci{.44}{.61} \\
 & \gdino & .59 \ci{.52}{.66} & $-$.67 \ci{$-$.96}{$-$.38} & .90 & .39 \ci{.31}{.48} \\
 & \owl & .57 \ci{.46}{.69} & .02 \ci{$-$.28}{.32} & .85 & .57 \ci{.49}{.66} \\
\midrule
\sdxl (S2) & \samt & .80 \ci{.66}{.94} & $-$.35 \ci{$-$.83}{.12} & .87 & .56 \ci{.48}{.63} \\
 & \gdino & .44 \ci{.28}{.61} & $-$.32 \ci{$-$.70}{.05} & .73 & .42 \ci{.33}{.50} \\
 & \owl & .68 \ci{.52}{.83} & .08 \ci{$-$.33}{.47} & .78 & .58 \ci{.49}{.66} \\
\midrule
\flux (S2) & \samt & .59 \ci{.40}{.76} & $-$.17 \ci{$-$.46}{.14} & .86 & .43 \ci{.34}{.53} \\
 & \gdino & .54 \ci{.41}{.66} & $-$.44 \ci{$-$.64}{$-$.24} & .84 & .31 \ci{.23}{.39} \\
 & \owl & .63 \ci{.50}{.76} & $-$.20 \ci{$-$.48}{.07} & .86 & .45 \ci{.37}{.54} \\
\midrule
\sdm $768^2$ (U2, two seeds) & \samt & .45 \ci{.35}{.55} & $-$.04 \ci{$-$.21}{.13} & .86 & .50 \ci{.45}{.55} \\
\sdm $1024^2$ (U2, 2, 4, 8 / 12, 20, 30) & \samt & .49 \ci{.39}{.57} & $-$.31 \ci{$-$.49}{$-$.13} & .87 & .38 \ci{.31}{.46} \\
\sdm + Flow-GRPO $512^2$ (S1, three seeds) & \samt & .65 & $-$.43 & .95 & .38 \ci{.32}{.43} \\
\sdm + DiffusionNFT $512^2$ (S1, three seeds) & \samt & .64 & $-$.57 & .92 & .32 \ci{.25}{.39} \\
\midrule
\emph{reference model, log-normal scatter .3 / .45} & & & .95 & & .83 / .76 \\
\bottomrule
\end{tabular}}
\caption{\textbf{The number word under three counters} (same noise, \cref{sec:setup}; one seed unless stated; \flux's large requests are 12, 16, 20 and 30). Every order interval at 12--30 lies below the reference model's (\cref{sec:supp_refmodels}). Orders below one half are reversals: as counted, the larger request more often gives fewer objects. They occur mostly under \gdino, which undercounts dense images most (\cref{sec:supp_counters}), and for the two post-trained checkpoints (\cref{sec:supp_reward}).}
\label{tab:supp_ask}
\end{table*}

Under every counter the word orders the crowd from two to eight and much less from twelve to thirty (\cref{tab:supp_ask}).

\para{Direction among the largest requests.} On request pairs that are both at least twenty, the larger request gives more objects in 44\% of the pairs whose counts differ for \sdm at $512^2$, 25\% for Flow-GRPO and 31\% for DiffusionNFT, reversals in all three; over all requests from 12 to 30 the order lies 0.39, 0.58 and 0.61 below its value from two to eight. Both hold under every counter.

\begin{table*}[h]
\centering
\scriptsize
\setlength{\tabcolsep}{2.4pt}
\begin{tabular}{@{}lcccccccccccccccccc@{}}
\toprule
requested $N$ & 2 & 3 & 4 & 6 & 7 & 8 & 10 & 11 & 12 & 13 & 14 & 16 & 17 & 20 & 23 & 25 & 29 & 30 \\
\midrule
\sdm $512^2$ & .81 & .67 & .51 & .13 & .09 & .09 & .02 & .02 & .07 &  & .01 & .01 &  & .00 &  & .00 &  & .00 \\
\sdm $1024^2$ & .79 & .60 & .62 & .38 & .21 & .21 & .08 & .15 & .10 &  & .04 & .06 &  & .00 &  & .02 &  & .00 \\
\sdxl & .52 & .27 & .21 & .15 & .06 & .06 & .02 & .08 & .06 &  & .00 & .00 &  & .02 &  & .02 &  & .00 \\
\flux & .69 &  & .54 & .31 &  & .10 &  &  & .13 &  &  & .02 &  & .02 &  &  &  & .00 \\
\fluxk, U2 & .66 &  & .52 & .20 &  & .09 &  &  & .11 &  &  & .08 &  & .00 &  & .05 &  & .02 \\
\sdm $768^2$ plain, U1 &  &  &  &  &  &  &  &  & .09 & .01 &  &  & .01 & .01 & .01 &  & .01 & .00 \\
self-planting, U1 &  &  &  &  &  &  &  &  & .63 & .34 &  &  & .30 & .45 & .30 &  & .28 & .34 \\
\bottomrule
\end{tabular}
\caption{\textbf{Exact rate by requested count} (\samt), the data of \cref{fig:ask_sam3}. Empty cells were not requested.}
\label{tab:supp_exact}
\end{table*}

\begin{table}[h]
\centering
\footnotesize
\setlength{\tabcolsep}{2.2pt}
\resizebox{\linewidth}{!}{%
\begin{tabular}{@{}llccc@{}}
\toprule
& & order & slope & exact \\
\midrule
\sdm $512^2$ & people 2--8 & .95 \ci{.89}{1.00} & .57 \ci{.47}{.69} & .33 \ci{.25}{.39} \\
 & people 12--30 & .49 \ci{.34}{.64} & $-$.12 \ci{$-$.55}{.28} & .04 \ci{.00}{.10} \\
 & \samt 2--8 & .88 \ci{.75}{.97} & .46 \ci{.23}{.63} & .33 \ci{.25}{.39} \\
 & \samt 12--30 & .64 \ci{.47}{.81} & .10 \ci{$-$.28}{.47} & .00 \ci{.00}{.00} \\
\midrule
\sdm $768^2$ & people 2--8 & .80 \ci{.66}{.93} & .37 \ci{.16}{.53} & .36 \ci{.28}{.45} \\
 & people 12--30 & .54 \ci{.41}{.65} & .01 \ci{$-$.51}{.47} & .04 \ci{.00}{.08} \\
 & \samt 2--8 & .82 \ci{.65}{.96} & .35 \ci{.03}{.62} & .31 \ci{.22}{.42} \\
 & \samt 12--30 & .56 \ci{.41}{.70} & .00 \ci{$-$.55}{.52} & .04 \ci{.00}{.08} \\
\midrule
\flux & people 2--8 & .93 \ci{.85}{1.00} & .67 \ci{.55}{.76} & .53 \ci{.41}{.66} \\
 & people 12--30 & .52 \ci{.33}{.71} & $-$.09 \ci{$-$.48}{.28} & .08 \ci{.02}{.16} \\
 & \samt 2--8 & .81 \ci{.64}{.95} & .30 \ci{$-$.02}{.59} & .42 \ci{.31}{.55} \\
 & \samt 12--30 & .44 \ci{.26}{.63} & .02 \ci{$-$.46}{.63} & .03 \ci{.00}{.09} \\
\midrule
\sdxl & people 2--8 & .91 \ci{.85}{.97} & .81 \ci{.62}{1.01} & .30 \ci{.19}{.41} \\
 & people 12--30 & .77 \ci{.68}{.85} & 1.10 \ci{.63}{1.62} & .04 \ci{.00}{.08} \\
 & \samt 2--8 & .89 \ci{.82}{.96} & .79 \ci{.53}{1.01} & .27 \ci{.16}{.38} \\
 & \samt 12--30 & .62 \ci{.49}{.74} & .34 \ci{$-$.54}{1.25} & .03 \ci{.00}{.06} \\
\midrule
\fluxk & people 2--8 & .89 \ci{.80}{.96} & .58 \ci{.40}{.72} & .59 \ci{.50}{.70} \\
 & people 12--30 & .62 \ci{.53}{.72} & .12 \ci{$-$.34}{.54} & .13 \ci{.06}{.19} \\
 & \samt 2--8 & .71 \ci{.53}{.87} & .16 \ci{$-$.11}{.42} & .39 \ci{.27}{.53} \\
 & \samt 12--30 & .56 \ci{.42}{.69} & .10 \ci{$-$.54}{.90} & .05 \ci{.01}{.10} \\
\bottomrule
\end{tabular}}
\caption{\textbf{The word, by human count} (16 types per model, one seed; order, slope and exact rate as in \cref{sec:supp_measures}, 95\% noun-cluster intervals). \sdm $512^2$: S1; \sdm $768^2$: U2; \flux and \sdxl: S2 (first count); \fluxk: U2, second count. Without pictures marked unsure the picture is the same (\sdxl at 12--30: order 0.77, slope 1.08; \fluxk 0.66, 0.17).}
\label{tab:supp_hc_ask}
\end{table}

\para{People's orders.} By people (\cref{tab:supp_hc_ask}), every model orders the crowd from two to eight. On U2 (fifth count), people's order at 12--30 is 0.63 for \sdxl and 0.36 for \flux, a reversal; their caps are in \cref{tab:supp_ceiling}. On the screened types of S2, \flux's order at 12--30 is 0.52, not a reversal.

\para{Medians.} Median human count (and \samt) at requests 12, 16, 20, 25, 30 (with an even number of pictures a median can be a half): \sdm $512^2$ 8.5 (8), 8.5 (6.5), 8 (8), 9.5 (9), 7.5 (5.5); \sdm $768^2$ 6.5 (5), 6.5 (6), 6 (4), 8.5 (2.5), 4.5 (3.5); \flux (S2) at 12, 16, 20, 30: 9 (5), 9 (7), 8 (6), 7 (4.5); \sdxl (S2) 13.5 (15.5), 26.5 (27.5), 22.5 (21), 33 (31), 43.5 (26); \fluxk 12 (7), 15 (7), 19 (8.5), 20 (10), 9 (3.5). People's medians of all five models on U2 are in \cref{tab:supp_ceiling}.

\para{Models compared on the same object types.} Orders of different models are compared within a set, on the same 16 people-counted types (paired type bootstrap, 4{,}000 resamples; requests 12, 16, 20, 25 and 30, or 12, 16, 20 and 30 on S2, where \flux has no 25):

\begin{center}\small
\resizebox{\linewidth}{!}{%
\begin{tabular}{@{}llcc@{}}
\toprule
set & difference of order at 12--30 & people & \samt \\
\midrule
U2 & \qwen $-$ \sdm & $+$.16 \ci{$+$.01}{$+$.31} & $+$.22 \ci{$+$.06}{$+$.38} \\
U2 & \fluxk $-$ \sdm & $+$.08 \ci{$-$.07}{$+$.24} & .00 \ci{$-$.23}{$+$.23} \\
U2 & \qwen $-$ \fluxk & $+$.08 \ci{$-$.09}{$+$.24} & $+$.22 \ci{$+$.02}{$+$.42} \\
S2 & \sdxl $-$ \flux & $+$.26 \ci{.00}{$+$.51}$^a$ & $+$.16 \ci{$-$.05}{$+$.36} \\
\bottomrule
\end{tabular}}
\end{center}
At 2--8, \sdxl and \flux do not differ (people: $-$.02 \ci{$-$.12}{$+$.07}); \qwen orders above \sdm ($+$.19 \ci{$+$.07}{$+$.33}). $^a$The lower bound is 0.003. With 16 types and one seed, \fluxk cannot be ranked against \sdm or \qwen by order; the fifth count's comparisons of \sdxl and \flux with the U2 models are in \cref{sec:supp_ceiling}.

\para{\sdxl on U2.} By \samt the order is 0.76 \ci{0.68}{0.84} at 2--8 and 0.61 \ci{0.53}{0.69} at 12--30 (ties 26\%), the slope at 12--30 0.28 \ci{$-$0.05}{0.61}, exact rate at 12--30 0.9\%, and the median count 9.5--10 at every request from 12 to 30 except 25 (16); \owl: order 0.63 \ci{0.56}{0.69}; \gdino: 0.52 \ci{0.45}{0.60}.

\subsection{Where the crowd stops growing}
\label{sec:supp_ceiling}
\para{Model.} With $y=\log(1{+}n)$ and a type intercept $a_j$, the cap model is $y = a_j + \log(1{+}\min(c,K)) + \varepsilon$: the crowd follows the request up to the request $K$ and stays flat above it. $K$ is fitted by least squares on 500 log-spaced values from 2 to 60; $K\geq30$ means no ceiling within the requests. Intervals: 95\% bootstrap over object types (2{,}000 resamples), with $K$ refitted in every resample; paired ratios use the same resampled types for both models. As alternatives we fit a power law $y = a_j + \log(1{+}c^{\beta})$ and a broken stick with a profiled knee. Fitted again with other resamples, \sdm, \fluxk and \qwen keep their point estimates, and their intervals differ by at most 0.7 (\cref{tab:supp_ceiling} gives one fit for all U2 models).

\begin{table}[h]
\centering
\footnotesize
\setlength{\tabcolsep}{2.2pt}
\resizebox{\linewidth}{!}{%
\begin{tabular}{@{}lcccc@{}}
\toprule
& $K$, people & $K$, \samt & AIC$_{\text{pow}}{-}$AIC$_{\text{cap}}$ & people's medians at 12, 16, 20, 25, 30 \\
\midrule
\multicolumn{5}{@{}l}{\emph{U2, 16 types (fifth count)}} \\
\flux & 4.2 \ci{2.8}{6.8} & 2.8 \ci{2.0}{4.4} & $+$10.5 & 4, 8, 3.5, 2, 1.5 \\
\sdm & 6.5 \ci{3.0}{11.2} & 4.3 \ci{2.5}{9.5} & $-$3.7 & 6.5, 6.5, 6, 8.5, 4.5 \\
\sdxl & 10.9 \ci{7.4}{17.8} & 5.1 \ci{2.5}{10.8} & $-$3.8 & 14, 10, 14, 18, 18 \\
\qwen & 13.5 \ci{10.0}{15.1} & 10.3 \ci{2.3}{17.2} & $+$20.7 & 12, 12, 15.5, 16, 15 \\
\fluxk & 15.2 \ci{10.3}{20.7} & 5.4 \ci{2.7}{19.4} & $+$6.4 & 12, 15, 19, 20, 9 \\
\sdm, count LoRA & 9.8 \ci{7.6}{12.3} & 8.5 \ci{5.0}{13.4} & $+$32.7 & 12, 12, 12, 12, 9.5 \\
\sdm, shuffled LoRA & 12.0 \ci{9.5}{14.0} & 6.9 \ci{5.2}{11.4} & $+$38.5 & 36, 35, 33, 34.5, 17.5 \\
\multicolumn{5}{@{}l}{\emph{screened sets, 16 types (first count)}} \\
\sdm $512^2$ (S1) & 10.5 \ci{7.3}{13.7} & 6.9 \ci{3.6}{13.2} & $+$11.1 & 8.5, 8.5, 8, 9.5, 7.5 \\
\flux (S2) & 7.7 \ci{5.3}{10.6} & 4.5 \ci{2.3}{7.5} & $+$13.1 & 9, 9, 8, --, 7 \\
\sdxl (S2) & $\geq$30 \ci{21.3}{} & 13.7 \ci{6.7}{27.6} & $-$0.1 & 13.5, 26.5, 22.5, 33, 43.5 \\
\bottomrule
\end{tabular}}
\caption{\textbf{Ceilings} (cap model; 95\% intervals over types). AIC$_{\text{pow}}{-}$AIC$_{\text{cap}}$: positive when the capped crowd fits better than an uncapped power law. \sdxl on U2 fits a power law (exponent 0.66) slightly better, and \sdm on U2, whose crowd already falls below the request from four, about as well, though its slope at 12--30 is 0.01. \sdxl's $K$ on S2 is at least 30 in 57\% of resamples (power 1.01 \ci{0.88}{1.14}). \flux on S2 was not asked for 25.}
\label{tab:supp_ceiling}
\end{table}

\para{Paired ratios of $K$.} Fifth count (same resampled types): \sdxl over \flux 2.58 \ci{1.88}{3.94}, \sdxl over \sdm 1.68 \ci{1.22}{2.96}, over \qwen 0.81 \ci{0.56}{1.32} and over \fluxk 0.72 \ci{0.45}{1.20}; \flux over \sdm 0.65 \ci{0.37}{1.24}, over \qwen 0.31 \ci{0.21}{0.49} and over \fluxk 0.28 \ci{0.17}{0.42}; the count over the shuffled LoRA 0.82 \ci{0.68}{0.99} and over the untrained model 1.51 \ci{0.90}{2.80}. Screened over unscreened types, unpaired (each set's types resampled independently): \sdxl 2.75 \ci{1.47}{4.02}, \flux 1.82 \ci{1.09}{3.14}. First counts: \qwen over \sdm 2.07 \ci{1.25}{4.33}, \fluxk over \sdm 2.34 \ci{1.24}{5.69}, \qwen over \fluxk 0.88 \ci{0.65}{1.24}, \sdxl over \flux on S2 at least 3.89 \ci{2.54}{5.17}. Paired orders at 12--30 on U2 (fifth count): \sdxl minus \flux $+$0.28 \ci{0.11}{0.43}, minus \sdm $+$0.09 \ci{$-$0.04}{0.23}, minus \fluxk $+$0.01 \ci{$-$0.14}{0.17}, minus \qwen $-$0.07 \ci{$-$0.21}{0.08}; \flux minus \sdm $-$0.18 \ci{$-$0.28}{$-$0.08}, minus \fluxk $-$0.26 \ci{$-$0.41}{$-$0.12} and minus \qwen $-$0.34 \ci{$-$0.51}{$-$0.18}.

\para{Scatter net of counting noise.} For each picture of the first four counts the two first-round counts are independent, so the covariance of their residuals from the cap fit estimates the variance of the picture's own deviation from the fitted crowd. In log units: \sdm 0.51 \ci{0.34}{0.66} ($512^2$) and 0.46 \ci{0.20}{0.61} ($768^2$), \flux 0.38 \ci{0.26}{0.50}, \sdxl 0.40 \ci{0.21}{0.55}, \fluxk 0.50 \ci{0.32}{0.62}, \qwen 0.00 \ci{0.00}{0.25} (its crowds sit on the fit). One person's counting noise is 0.38--0.66 over all requests and 0.47--0.80 at 12--30. Without pictures marked unsure, $K$ changes by at most 1.4 (\sdxl's on S2 falls from at least 30 to 28.3); without the request of 30, \fluxk's rises to 21.1 \ci{13.0}{25.1}, \sdxl's on S2 is 25.1, the top of the reduced range, and the others move by at most 0.4.

On the same 16 U2 types, \flux's crowd stops growing first and \fluxk's last; on the screened types, \sdxl's keeps growing up to thirty.

\subsection{Ways of asking and guidance}
\label{sec:supp_wording}
\sdm, U2. Digits (``12 $X$s''), the rows-and-columns sentence and native resolution use the requests 12, 20 and 30; guidance $w\in\{1.5,3,4.5,7,10\}$ uses 12, 16, 20, 25 and 30, with 2, 4 and 6 for the small-count lift; the best of four plain samples, chosen by \samt among four seeds, uses all nine counts and costs about as much as one self-planted image of a new object type, including its cut-out.

Exact at 12--30 by \samt: plain prompt 0.014 (nine requests, two seeds), digits 0.026, the rows-and-columns sentence 0.062 and native resolution 0.034 (12, 20 and 30; two seeds for native resolution), best of four 0.050 (nine requests), against 0.406 for self-planting on the same cells. Exact at 12--30 by guidance scale: 0.006, 0.019, 0.025, 0.028 and 0.034; exact lift at 12--30: 0.000, 0.004, 0.005, 0.009 and 0.023 \ci{0.005}{0.043}; at 2--6: 0.229, 0.341, 0.391, 0.409 and 0.424. At every scale the slope at 12--30 is $-0.31$ to $-0.18$ and the order 0.39--0.46, while the slope over 2, 4 and 6 is 0.40--0.53. On S1, digits give an exact lift of 0.001 \ci{$-$0.015}{0.019} at 10--30. No wording and no guidance scale makes exact counts past ten common, while self-planting on the same cells does.

\subsection{Reward-tuned checkpoints}
\label{sec:supp_reward}
Flow-GRPO~\cite{liu2025flowgrpo} (GenEval reward, counts of two to four) and DiffusionNFT~\cite{zheng2025diffusionnft} (a multi-reward checkpoint) are run like the base model on S1. Relative to the base model on the same cells, Flow-GRPO raises the exact lift by 0.323 \ci{0.255}{0.388} at 2--4 objects and by 0.017 \ci{$-$0.002}{0.035} at 10--16 ($+$0.012 at 10--30). Two further checkpoints, Flow-GRPO without its KL term (rewarded at two to four) and a GenEval2-rewarded LoRA~\cite{dppo2025,kamath2025geneval2} (rewarded at two to seven, at $1024^2$ against a $1024^2$ base), raise the exact lift by $+0.16$ and $+0.11$ where they are rewarded and move it by $-0.005$ \ci{$-$0.018}{0.007} and $+0.009$ \ci{$-$0.029}{0.045} at 10--30; under \gdino the second moves it by $+0.044$ \ci{0.016}{0.070}. The second checkpoint's own lift at 10--30 is $+0.034$ \ci{0.007}{0.061} under \samt, the largest we measured above ten: it draws the exact count in 11.2\% of images at 10--30 against 5.7\% for its base, most of them at requests of ten and eleven (31\% and 38\%), and in 3.5\% at 12--30. Two Flow-GRPO checkpoints tuned for other rewards (PickScore and text rendering) move it by $+0.22$ and $+0.05$ at two to four and by $+0.012$ \ci{$-$0.012}{0.040} and $-0.003$ \ci{$-$0.016}{0.009} at 10--30. Rewards on small counts raise the exact lift where they reward it and barely above ten, and the largest lift above ten comes mostly from requests of ten and eleven.

\subsection{A newer model: \fluxk}
\label{sec:supp_fluxk}
\para{Design.} \fluxk base 4B~\cite{flux2klein2026}, the undistilled four-billion-parameter model whose text encoder is the language model Qwen3-4B~\cite{qwen3}, in diffusers 0.41 at $1024^2$ with 50 steps and guidance 4.0, on U2's 64 types with U2's prompts, requests 2, 4, 6, 8, 12, 16, 20, 25 and 30, seed 1 (576 images). One starting latent per type is shared by the nine requests. \sdm's U2 images of the same cells and seed are the comparison under the same counters. One of the 576 images is flat.

\begin{table}[h]
\centering
\footnotesize
\setlength{\tabcolsep}{2.6pt}
\resizebox{\linewidth}{!}{%
\begin{tabular}{@{}llcccccc@{}}
\toprule
& & \multicolumn{2}{c}{order} & slope & exact & lift & none \\
\cmidrule(lr){3-4}
model & counter & 2--8 & 12--30 & 12--30 & 12--30 & 12--30 & 12--30 \\
\midrule
\fluxk & \samt & .81 & .67 \ci{.60}{.74} & .29 \ci{.05}{.54} & .050 & .030 & .39 \\
 & \gdino & .89 & .45 \ci{.39}{.51} & $-$.55 & .028 & .009 & \\
 & \owl & .79 & .51 \ci{.44}{.58} & $-$.13 & .025 & .009 & \\
 & consensus & .87 & .59 \ci{.44}{.74} & $-$.37 & .025 & .012 & \\
 & people (16 types) & .89 & .62 \ci{.53}{.72} & .12 & .125 & & \\
\midrule
\sdm $768^2$ & \samt & .89 & .46 \ci{.38}{.53} & $-$.20 & .025 & .005 & .21 \\
 & \gdino & .78 & .39 \ci{.31}{.47} & $-$.35 & .009 & .001 & \\
 & \owl & .85 & .45 \ci{.38}{.53} & $-$.24 & .034 & .016 & \\
 & consensus & .89 & .33 \ci{.22}{.46} & $-$.43 & .016 & .006 & \\
\bottomrule
\end{tabular}}
\caption{\textbf{\fluxk and \sdm on the same cells} (U2, 64 types, seed 1; none: \samt finds none of the object). \fluxk minus \sdm, order at 12--30: $+$.22 \ci{.11}{.32} under \samt, $+$.06 under \gdino and \owl (intervals include 0), $+$.26 \ci{.05}{.45} under the consensus (74 and 120 differing pairs). Reference model with this design's types and seed: 0.84 (scatter 0.3) and 0.76 (0.45).}
\label{tab:supp_fluxk}
\end{table}

The counters fail on \fluxk's dense images. \samt finds none of the object in 39\% of them at 12--30 (21\% for \sdm), and the detectors' medians fall to 1--6 objects at 20--30, where people's medians on 16 of the types are 19, 20 and 9. \samt's median count is 2, 4, 4 and 5 for requests 2--8 and 8, 8, 10, 15 and 4.5 for 12--30; its exact rate .66, .52, .20, .09 and .11, .08, 0, .05, .02. People's counts of 16 types (\cref{sec:supp_human}) place the order at 12--30 between one half and a loose estimator: 0.62 \ci{0.53}{0.72}, 0.66 \ci{0.55}{0.76} without pictures marked unsure, with a slope of 0.12 \ci{$-$0.34}{0.54}. Their median count is 12, 15, 19, 20 and 9 for requests 12, 16, 20, 25 and 30. Over 12--25 alone the order is 0.73 \ci{0.62}{0.84} and the slope 0.84 \ci{0.29}{1.40}, where the reference model with these requests gives 0.74--0.82 and 0.94: up to twenty-five \fluxk follows the request about as loosely as the looser reference, and at thirty it collapses. People find the asked number in 13\% of these images (10 of 80), against 4\% for \sdm on the same types and requests (difference 0.09 \ci{0.03}{0.15}).

\subsection{A current large model: \qwen}
\label{sec:supp_qwen}
\para{Design.} \qwen~\cite{wu2025qwenimage}, a 20B transformer whose text encoder is the vision-language model Qwen2.5-VL-7B~\cite{bai2025qwen25vl}, from the official weights (revision \texttt{75e0b4b}) in diffusers 0.41 with the pipeline's defaults: $1024^2$, 50 steps, true classifier-free guidance 4.0 with the negative prompt `` ''. To fit a 24\,GB L4 GPU, the transformer is quantized to 4-bit NF4 with bitsandbytes (compute in bf16); the text encoder, VAE and scheduler are unchanged, and the sampler passes the text encoder's padding mask as the pipeline returns it. Prompts are U2's (``a photo of twelve cribs''), requests 2, 4, 6, 8, 12, 16, 20, 25 and 30, seed 1; one starting latent per type, drawn on the CPU from the type's seed, is shared by the nine requests. \qwen was run on the 16 people-counted types of U2 (144 images), which people counted in the fourth study (\cref{sec:supp_human}). An image takes about 173\,s.

\begin{table}[h]
\centering
\footnotesize
\setlength{\tabcolsep}{2.4pt}
\resizebox{\linewidth}{!}{%
\begin{tabular}{@{}llccccc@{}}
\toprule
& & \multicolumn{2}{c}{order} & slope & exact & within 10\% \\
\cmidrule(lr){3-4}
model & counter & 2--8 & 12--30 & 12--30 & 12--30 & 12--30 \\
\midrule
\qwen & \samt & .70 & .78 \ci{.67}{.88} & .49 & .138 & .20 \\
 & \gdino & .92 & .47 \ci{.29}{.64} & $-$.44 & .138 & \\
 & \owl & .71 & .56 \ci{.39}{.73} & $-$.01 & .088 & \\
 & consensus & .78 & .72 \ci{.51}{.90} & .12 & .125 & \\
 & people & .99 & .70 \ci{.57}{.82} & .23 & .213 & .28 \\
\midrule
\sdm $768^2$ & \samt & .82 & .56 \ci{.41}{.70} & .00 & .038 & \\
 & people (A4) & .80 & .54 \ci{.41}{.65} & .01 & .038 & .10 \\
\fluxk & \samt & .71 & .56 \ci{.42}{.69} & .10 & .050 & \\
 & people (N) & .89 & .62 \ci{.53}{.72} & .12 & .125 & .23 \\
\bottomrule
\end{tabular}}
\caption{\textbf{\qwen, \sdm and \fluxk on the same cells} (the 16 people-counted types of U2, seed 1). \qwen minus \sdm by people: order at 12--30 $+$.16 \ci{.02}{.31}, exact $+$.18 \ci{.11}{.24}, within 10\% $+$.18 \ci{.08}{.28}, recognisable $+$.15 \ci{.09}{.21}; minus \fluxk: $+$.08 \ci{$-$.09}{.24}, $+$.09 \ci{$-$.01}{.19}, $+$.05 \ci{$-$.05}{.15}, $+$.11 \ci{.01}{.21}. Under \samt, \qwen minus \sdm: order $+$.22 \ci{.06}{.39}, exact $+$.10 \ci{.04}{.16}.}
\label{tab:supp_qwen}
\end{table}

\para{Results.} By people, \qwen is exact in 100\%, 94\%, 94\% and 50\% of images at 2, 4, 6 and 8, and in 63\%, 25\%, 13\%, 6\% and 0\% at 12, 16, 20, 25 and 30 (within 10\%: 75\%, 31\%, 25\%, 6\% and 0\%); its median count is 12, 12, 15.5, 16 and 15. People's order at 12--30 is 0.70 \ci{0.57}{0.82} (0.70 without pictures marked unsure) and their exact rate 0.21: \qwen's word orders large crowds partly and rarely gives their exact count. People tie on 23\% of same-type pairs at 12--30, and judge the marked objects to look like the named one in 75\% of the 142 pictures with marked objects (``partly'' in the rest). Counters miss the larger grids: against people, \samt is exact on 55\% of \qwen's pictures and undercounts by 2.5 objects on average (by 3.0 at 25--30); it finds no object in 24\% of \qwen's images, most often at requests of 4 to 12.

\para{Grids and object scale.} \samt's boxes form a complete grid (at least two rows and two columns, every row holding the same number of objects, the count equal to rows times columns; rows and columns found from gaps of half the median box size) in 34\% of \qwen's images at 12--30, against 9\%, 14\% and 0\% for \sdm, \fluxk and \sdxl on the same types; people find the requested count in 37\% of \qwen's complete grids and in 13\% of its other images (\sdm: 29\% of its 7 complete grids against 1\%; \fluxk: 18\% of 11 against 12\%), and 10 of \qwen's 17 exact images at 12--30 are complete grids. \qwen's objects shrink with the request at 2--8 (scale slope $-$1.03 \ci{$-$1.41}{$-$0.77}) and not at 12--30 ($-$0.12 \ci{$-$0.33}{0.09}).

\subsection{The room test}
\label{sec:supp_room}
\para{Design.} \sdm at $768^2$, 40 steps, guidance 4.5, U2's 64 types, seed 1, the plain sampler; every image copies the task of U2's plain image of its type and request (same noise) and changes only the prompt: ``a photo of twelve cribs, seen from far away'' (\emph{far}) or ``a photo of twelve tiny cribs'' (\emph{tiny}), at 2, 4, 6, 8, 12, 16, 20, 25 and 30 (1{,}152 images), counted by \samt, \gdino and \owl. If the frame's room limited the count, a cue that makes the objects smaller should let the crowd follow the request; how much a cue shrinks the objects is measured by the area ratio of \cref{tab:supp_room}.

\para{Results.} \emph{far} shrinks the objects to 0.47 times the plain area \ci{0.35}{0.61}; they cover 29--36\% of the median image at 12--30, against 50--64\% without the cue. Its order at 12--30 is 0.54 \ci{0.46}{0.61}, against 0.46 \ci{0.38}{0.53} for the plain prompt from the same noise (difference 0.08 \ci{$-$0.02}{0.19}); the median count rises to 9--10 but stays flat across requests. \emph{tiny} shrinks the objects less (0.72 \ci{0.60}{0.87}), yet raises the order to 0.648 \ci{0.57}{0.72} (difference 0.19 \ci{0.11}{0.28}); under \owl it raises it too (0.62, difference 0.16 \ci{0.07}{0.25}), but not under \gdino (0.46, difference 0.07 \ci{$-$0.04}{0.18}). Neither cue changes the exact rate at 12--30 (2.2\% and 2.5\%, against 2.5\%) or the order at 2--8 (0.88 and 0.85, against 0.89). Halving the objects' area does not make the crowd follow the request, so room does not limit the count; ``tiny'' moves the order partway under \samt and \owl but not under \gdino.

\begin{table}[h]
\centering
\footnotesize
\setlength{\tabcolsep}{2.2pt}
\resizebox{\linewidth}{!}{%
\begin{tabular}{@{}lccccccc@{}}
\toprule
& area & box \% & \multicolumn{2}{c}{order} & slope & exact & median \\
\cmidrule(lr){4-5}
prompt & ratio & 12 / 30 & 2--8 & 12--30 & 12--30 & 12--30 (\%) & 12 / 20 / 30 \\
\midrule
\multicolumn{8}{@{}l}{\emph{\samt}} \\
plain & 1 & 6.8 / 10.8 & .89 & .46 \ci{.38}{.53} & $-$.20 & 2.5 & 8 / 6 / 4.5 \\
``seen from far away'' & .47 & 4.2 / 3.3 & .88 & .54 \ci{.46}{.61} & $-$.04 & 2.2 & 10 / 9 / 9 \\
``tiny'' & .72 & 4.2 / 4.8 & .85 & .65 \ci{.57}{.72} & .27 & 2.5 & 7 / 7 / 7.5 \\
\multicolumn{8}{@{}l}{\emph{\gdino}} \\
plain &  &  & .78 & .39 \ci{.31}{.47} & $-$.35 & .9 & 4 / 1 / 1 \\
``seen from far away'' &  &  & .81 & .39 \ci{.33}{.46} & $-$.55 & 1.6 & 6 / 2 / 1.5 \\
``tiny'' &  &  & .82 & .46 \ci{.38}{.55} & $-$.43 & 1.9 & 8 / 4.5 / 2 \\
\multicolumn{8}{@{}l}{\emph{\owl}} \\
plain &  &  & .85 & .45 \ci{.38}{.53} & $-$.24 & 3.4 & 8 / 7 / 5 \\
``seen from far away'' &  &  & .86 & .55 \ci{.48}{.61} & $-$.02 & 1.9 & 9 / 9 / 8 \\
``tiny'' &  &  & .83 & .62 \ci{.55}{.68} & .27 & 2.2 & 8 / 10 / 9 \\
\multicolumn{8}{@{}l}{\emph{consensus}} \\
plain &  &  & .89 & .33 \ci{.22}{.46} & $-$.43 & 1.6 & 6.5 / 1 / 1.5 \\
``seen from far away'' &  &  & .94 & .54 \ci{.40}{.68} & $-$.13 & .9 & 8 / 1 / 1 \\
``tiny'' &  &  & .89 & .57 \ci{.39}{.72} & $-$.48 & 1.3 & 8 / 2 / 2 \\
\bottomrule
\end{tabular}}
\caption{\textbf{The room test} (\sdm, U2, seed 1, same noise as the plain prompt). Area ratio: geometric mean, over the type--request pairs at 12--30 with an object in both images, of the cue image's median \samt box area over the plain image's; box \%: median box area at 12 and 30 (images with an object).}
\label{tab:supp_room}
\end{table}

\subsection{Object size}
\label{sec:supp_objsize}
\begin{figure}[h]
\centering
\includegraphics[width=0.86\linewidth]{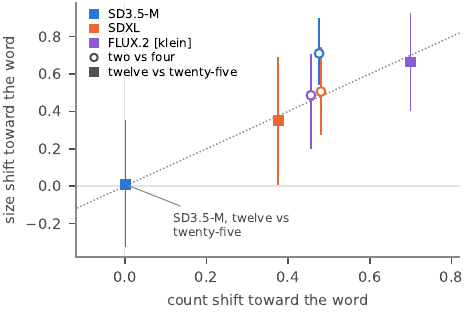}
\caption{\textbf{A word that moves the count also shrinks the objects} (the word asked at every step against the base request's image, \samt, 64 types). Shift in object size toward the word (log median box area, smaller for the larger number) against shift in count, per model and pair, with 95\% intervals; dotted: equal shifts. Past \sdm's reach, ``twelve'' against ``twenty-five'' does neither.}
\label{fig:objsize}
\end{figure}
With the word swapped in for the whole trajectory (the timing test's full-swap cells, \samt; \cref{fig:objsize}), wherever the word moves the count toward itself it also moves the objects' size: for ``two'' against ``four'' in all three models of the timing test, and for ``twelve'' against ``twenty-five'' in \sdxl and \fluxk, a larger number gives more and smaller objects, with a shift of the log median box area 0.9--1.5 times the shift of $\log(1{+}n)$ (\cref{fig:objsize}). Past \sdm's reach the same swap shifts neither on average (size $+$0.01 \ci{$-$0.32}{0.35}, count 0.00), although it changes the crowd nearly as much as a new seed: the model makes room for more objects by shrinking them, as long as the word points it there. How a model fits a larger crowd into the same frame shows in its object sizes (\cref{tab:supp_objsize}; \samt boxes of the plain images with at least one object found). \sdm's objects shrink with the request up to about ten and then stop: past ten its median box covers 6--11\% of the image whatever the request, and the boxes together cover a similar share of the image (54--62\% in the median image at 12--30; two seeds). \fluxk's objects keep shrinking past ten, to about 2\% of the image at 25 and 30. \sdxl's objects keep shrinking past ten on U2 (to 2.0\% of the image at thirty); on its screened set they are already small at twelve and shrink less by \samt, which misses the smallest objects of its densest crowds (\cref{tab:supp_hc_agree}) and finds none of the object in 23--27\% of \sdxl's U2 images at 12--30 (left out of the box medians). \qwen's objects (16 types of U2) shrink with the request up to about twelve and then stop, as its crowd does.
\begin{table}[!htbp]
\centering
\footnotesize
\setlength{\tabcolsep}{2.2pt}
\resizebox{\linewidth}{!}{%
\begin{tabular}{@{}lcccccccccc@{}}
\toprule
& \multicolumn{8}{c}{median box area (\% of image), by request} & \multicolumn{2}{c}{slope of log area} \\
\cmidrule(lr){2-9}\cmidrule(lr){10-11}
model & 2 & 4 & 8 & 12 & 16 & 20 & 25 & 30 & 2--8 & 12--30 \\
\midrule
SD3.5-M, U2 (two seeds) & 27.2 & 16.2 & 10.7 & 7.2 & 8.0 & 7.7 & 6.0 & 9.2 & $-$0.70 \ci{$-$0.82}{$-$0.58} & $+$0.18 \ci{$-$0.05}{$+$0.41} \\
FLUX.2 [klein], U2 & 26.5 & 14.7 & 7.4 & 5.0 & 4.5 & 2.7 & 2.2 & 2.0 & $-$0.75 \ci{$-$0.95}{$-$0.54} & $-$0.73 \ci{$-$1.15}{$-$0.34} \\
SDXL, U2 & 15.9 & 10.7 & 4.6 & 3.8 & 3.1 & 3.1 & 2.5 & 2.0 & $-$0.75 \ci{$-$0.98}{$-$0.52} & $-$0.55 \ci{$-$0.85}{$-$0.26} \\
SDXL, screened set & 16.0 & 10.9 & 5.0 & 2.9 & 2.7 & 2.2 & 2.0 & 2.0 & $-$0.81 \ci{$-$1.00}{$-$0.61} & $-$0.20 \ci{$-$0.65}{$+$0.32} \\
Qwen-Image, U2 (16 types) & 29.4 & 15.3 & 9.1 & 6.4 & 6.7 & 5.4 & 6.4 & 4.5 & $-$1.03 \ci{$-$1.41}{$-$0.77} & $-$0.12 \ci{$-$0.33}{$+$0.09} \\
\bottomrule
\end{tabular}}
\caption{\textbf{Object size against the request} (\samt, plain images). Slope: per type, the least-squares slope of the log median box area on the log request (images with at least one object; types with at least three such requests), averaged over types with a 95\% bootstrap interval (\qwen: 9 types at 2--8 and 13 at 12--30); $-1$ means objects shrink just enough for the requested number to fit, 0 that their size ignores the request.}
\label{tab:supp_objsize}
\end{table}

\subsection{Orders by window, effective scatter and an operational reach}
\label{sec:supp_pairs}
\para{Design.} A descriptive re-analysis of existing counts (no new images): the annotators' final counts of the word tests, 16 types per set and seed 1, and \samt's counts of all types of each set. The order is that of \cref{eq:order}, pooled over types as a ratio of sums with ties dropped; intervals use the fixed type-bootstrap indices of the other tests (20{,}000 resamples, all seeds of a type resampled together). Windows: 2--8; 12--30; 12--25, exactly the pairs of 12--30 without thirty; the four pairs with thirty; adjacent requests; and sliding windows of three consecutive requests. \emph{Effective scatter}: the loose estimator of \cref{sec:supp_refmodels}, $n=\mathrm{round}(c\,e^{sz})$, is simulated for $s=0.05,0.10,\dots,3.00$ and read through the annotators' noise model (\cref{sec:supp_noise}) with the counting difficulty of the model's own pictures; its expected order on the window (100{,}000 simulated types per $s$, made non-increasing in $s$) is inverted at the model's order, and the order's interval is mapped through the same curve. Matched to the models' pictures, the estimator orders 12--25 at 0.79--0.80 ($s{=}0.30$) and 0.72 ($s{=}0.45$), and 2--8 at about 0.92 and 0.85. \emph{Reach} (sliding): the middle request of the first window of three consecutive requests whose effective scatter exceeds $s^*$, or 30 if none; we use $s^*{=}1$, a typical factor of 2.7, about twice the loosest reference scatter (\cref{tab:supp_pairs_reach}). Alternatives: $s^*{=}0.75$, and the first adjacent pair whose order lies below the 2.5th percentile of the $s{=}0.45$ estimator.

\begin{table*}[!htbp]
\centering
\footnotesize
\setlength{\tabcolsep}{3pt}
\begin{tabular}{@{}lcccccc@{}}
\toprule
model, set & 2--8 & 12--30 & 12--25 & pairs with 30 & with 30 $-$ 12--25 & 20/25 \\
\midrule
\flux & .90 \ci{.79}{.98} & .36 \ci{.23}{.49} & .47 \ci{.30}{.64} & .17 \ci{.03}{.35} & $-$.29 \ci{$-$.51}{$-$.08} & .38 (3/5/8) \\
\sdm & .80 \ci{.66}{.93} & .54 \ci{.41}{.65} & .57 \ci{.43}{.69} & .49 \ci{.29}{.67} & $-$.08 \ci{$-$.29}{$+$.14} & .67 (8/4/4) \\
\sdxl & .89 \ci{.76}{.98} & .63 \ci{.52}{.74} & .65 \ci{.53}{.78} & .60 \ci{.42}{.77} & $-$.05 \ci{$-$.27}{$+$.14} & .57 (8/6/2) \\
\qwen & .99 \ci{.97}{1.00} & .70 \ci{.57}{.82} & .78 \ci{.65}{.90} & .59 \ci{.40}{.76} & $-$.19 \ci{$-$.37}{$-$.02} & .45 (5/6/5) \\
\fluxk & .89 \ci{.80}{.96} & .62 \ci{.53}{.72} & .73 \ci{.62}{.84} & .44 \ci{.23}{.65} & $-$.30 \ci{$-$.54}{$-$.04} & .50 (8/8/0) \\
\midrule
\sdm $512^2$, S1 & .95 \ci{.89}{1.00} & .49 \ci{.34}{.64} & .57 \ci{.41}{.72} & .36 \ci{.16}{.56} & $-$.21 \ci{$-$.38}{$-$.05} & .82 (9/2/5) \\
\sdxl, S2 & .91 \ci{.85}{.97} & .77 \ci{.68}{.85} & .78 \ci{.68}{.87} & .75 \ci{.61}{.88} & $-$.03 \ci{$-$.18}{$+$.12} & .81 (13/3/0) \\
\flux, S2$^a$ & .93 \ci{.85}{1.00} & .52 \ci{.33}{.71} & .65 \ci{.44}{.82} & .42 \ci{.18}{.66} & $-$.23 \ci{$-$.46}{$-$.01} & -- \\
\bottomrule
\end{tabular}
\caption{\textbf{Same-noise order by window} (annotators' final counts, 16 types per set, seed 1; 95\% type-bootstrap intervals with the fixed resampled indices used throughout). Pairs with 30: 12/30, 16/30, 20/30 and 25/30; ``12--25'' is exactly the set of pairs of 12--30 without thirty, and the difference uses the same resamples. 20/25: order (up/down/tie of the 16 types). The first five rows share the 16 U2 types. $^a$Not asked for 25: its 12--25 is 12, 16, 20.}
\label{tab:supp_pairs_orders}
\end{table*}

\begin{table*}[!htbp]
\centering
\footnotesize
\setlength{\tabcolsep}{3pt}
\begin{tabular}{@{}lcccccccc@{}}
\toprule
model, set & 2/4 & 4/6 & 6/8 & 8/12 & 12/16 & 16/20 & 20/25 & 25/30 \\
\midrule
\flux & 1.00 (14/0/2) & .69 (9/4/3) & .92 (11/1/4) & .42 (5/7/4) & .67 (8/4/4) & .45 (5/6/5) & .38 (3/5/8) & .12 (1/7/8) \\
\sdm & .87 (13/2/1) & .88 (7/1/8) & .56 (5/4/7) & .71 (10/4/2) & .58 (7/5/4) & .50 (7/7/2) & .67 (8/4/4) & .27 (3/8/5) \\
\sdxl & .77 (10/3/3) & .92 (12/1/3) & 1.00 (8/0/8) & .71 (10/4/2) & .58 (7/5/4) & .71 (10/4/2) & .57 (8/6/2) & .50 (6/6/4) \\
\qwen & .94 (15/1/0) & 1.00 (15/0/1) & 1.00 (14/0/2) & .87 (13/2/1) & .86 (6/1/9) & .77 (10/3/3) & .45 (5/6/5) & .36 (5/9/2) \\
\fluxk & .93 (13/1/2) & .93 (14/1/1) & .60 (6/4/6) & 1.00 (12/0/4) & .75 (9/3/4) & .75 (12/4/0) & .50 (8/8/0) & .33 (5/10/1) \\
\midrule
\sdm $512^2$, S1 & 1.00 (12/0/4) & .75 (9/3/4) & 1.00 (12/0/4) & 1.00 (13/0/3) & .46 (6/7/3) & .54 (7/6/3) & .82 (9/2/5) & .17 (2/10/4) \\
\sdxl, S2 & .93 (14/1/1) & .94 (15/1/0) & .71 (10/4/2) & .80 (12/3/1) & .69 (11/5/0) & .64 (9/5/2) & .81 (13/3/0) & .44 (7/9/0) \\
\flux, S2$^a$ & .94 (15/1/0) & .86 (12/2/2) & .93 (14/1/1) & .64 (7/4/5) & .60 (6/4/6) & .67 (8/4/4) & -- & -- \\
\bottomrule
\end{tabular}
\caption{\textbf{Adjacent requests} (annotators, 16 types, seed 1): order (up/down/tie). $^a$\flux on S2 was not asked for 25 (its 20/30 pair: .55, 6/5/5).}
\label{tab:supp_pairs_adj}
\end{table*}

\begin{table*}[!htbp]
\centering
\footnotesize
\setlength{\tabcolsep}{3pt}
\begin{tabular}{@{}lccccccc@{}}
\toprule
& \multicolumn{3}{c}{effective scatter (annotators, matched)} & \multicolumn{3}{c}{reach} & \\
\cmidrule(lr){2-4}\cmidrule(lr){5-7}
model, set & 2--8 & 12--25 & 12--30 & sliding, $s^*{=}1$ & sliding, $s^*{=}0.75$ & adjacent pairs & cap $K$ \\
\midrule
\flux & 0.30 \ci{$<$0.05}{0.63} & $>$3 \ci{0.74}{$>$3} & $>$3 \ci{$>$3}{$>$3} & 8 \ci{8}{20} & 8 \ci{4}{20} & 8 \ci{4}{25} & 4.2 \ci{2.8}{6.8} \\
\sdm & 0.55 \ci{0.20}{1.31} & 1.55 \ci{0.53}{$>$3} & $>$3 \ci{0.78}{$>$3} & 16 \ci{6}{25} & 16 \ci{4}{16} & 25 \ci{6}{30} & 6.5 \ci{3.0}{11.2} \\
\sdxl & 0.25 \ci{$<$0.05}{0.64} & 0.69 \ci{0.32}{$>$3} & 0.90 \ci{0.45}{$>$3} & 12 \ci{8}{30} & 12 \ci{4}{25} & 30 \ci{2}{30} & 10.9 \ci{7.4}{17.8} \\
\qwen & $<$0.05 \ci{$<$0.05}{0.18} & 0.33 \ci{0.17}{0.68} & 0.55 \ci{0.29}{1.71} & 25 \ci{20}{30} & 25 \ci{20}{25} & 25 \ci{20}{30} & 13.5 \ci{10.0}{15.1} \\
\fluxk & 0.36 \ci{0.17}{0.60} & 0.42 \ci{0.24}{0.89} & 0.99 \ci{0.52}{$>$3} & 25 \ci{16}{25} & 25 \ci{6}{25} & 25 \ci{6}{30} & 15.2 \ci{10.3}{20.7} \\
\midrule
\sdm $512^2$, S1 & 0.18 \ci{$<$0.05}{0.35} & 1.65 \ci{0.46}{$>$3} & $>$3 \ci{0.87}{$>$3} & 16 \ci{12}{30} & 16 \ci{12}{25} & 25 \ci{12}{25} & 10.5 \ci{7.3}{14.1} \\
\sdxl, S2 & 0.21 \ci{$<$0.05}{0.39} & 0.32 \ci{0.18}{0.58} & 0.38 \ci{0.23}{0.66} & 30 \ci{12}{30} & 30 \ci{8}{30} & 30 \ci{16}{30} & $\geq$30 \ci{20.9}{$\geq$30} \\
\flux, S2$^a$ & 0.23 \ci{$<$0.05}{0.45} & 0.64 \ci{0.23}{$>$3} & $>$3 \ci{0.58}{$>$3} & 20 \ci{12}{30} & 12 \ci{12}{30} & 30 \ci{8}{30} & 7.7 \ci{5.3}{10.6} \\
\bottomrule
\end{tabular}
\caption{\textbf{Effective scatter and reach} (annotators, 16 types, seed 1). Effective scatter: the $s$ at which a loose estimator, counted with the annotators' noise at the model's own counting difficulty, has the model's expected order; interval: the order's interval mapped through the same curve; $>$3: the order lies below the curve at $s{=}3$ (an order at or below 0.5 has no finite $s$). Reach, sliding: the middle request of the first window of three consecutive requests whose effective scatter exceeds $s^*$ (30 if none); adjacent pairs: the lower request of the first adjacent pair whose order lies below the 2.5th percentile of the $s{=}0.45$ estimator; intervals: 2.5th--97.5th percentiles over type resamples. Cap $K$: the cap model of \cref{sec:supp_ceiling}.}
\label{tab:supp_pairs_reach}
\end{table*}

\para{The pairs with thirty.} \fluxk and \qwen order 12--25 as a loose estimator would (effective scatters 0.42 and 0.33) and lose order on the pairs with thirty (differences $-$0.30 \ci{$-$0.54}{$-$0.04} and $-$0.19 \ci{$-$0.37}{$-$0.02}; \cref{tab:supp_pairs_orders}). \samt on all 64 of \fluxk's types agrees (12--25: 0.75 \ci{0.67}{0.83}; pairs with 30: 0.55 \ci{0.44}{0.65}); on \qwen's 16 types it shows no thirty effect ($+$0.01 \ci{$-$0.20}{$+$0.24}). \flux's reversal at 12--30 (0.36) comes from its largest requests: at 12--25 its order is 0.47, on the pairs with thirty 0.17. \sdm at $768^2$ and \sdxl show no thirty effect, \sdm at $512^2$ (S1) does. \sdm's word test on all 64 U2 types, which also asked 14, 19, 22 and 27, gives by \samt: pairs $(c,25)$ with $c$ from 12 to 22, 0.54 \ci{0.45}{0.63}; $(c,27)$ 0.54; $(c,30)$ 0.48; 25/27 0.47 and 27/30 0.43; its median for ``nineteen'' (4) is as low as for ``thirty''. Annotators' median at thirty lies below that at twenty-five in five of the seven sets that asked both (\sdm on U2 8.5 to 4.5 and on S1 9.5 to 7.5, \flux 2 to 1.5, \fluxk 20 to 9, \qwen 16 to 15; \sdxl 18 to 18 on U2 and 33 to 43.5 on S2). Single number words past the reach carry their own crowd sizes, so a verdict over 12--30 can hinge on one word.

\para{Between 20 and 30 in \fluxk and \qwen.} The seventh count's per-pair orders cannot be computed from its stored pooled values; its seed-2 order follows exactly from them (\qwen 28 of 38 differing pairs, 0.74; \fluxk 15 of 41, 0.37). On the same seed-1 pictures, the earlier annotators' count gives 20/25, 20/30 and 25/30 orders of 0.50 (8/8/0), 0.25 and 0.33 for \fluxk and 0.45, 0.45 and 0.36 for \qwen; \samt on both seeds gives 0.52 \ci{0.32}{0.72}, 0.20 \ci{0.00}{0.44} and 0.27 \ci{0.09}{0.47} for \fluxk and 0.74 \ci{0.53}{0.91}, 0.78 \ci{0.62}{0.92} and 0.57 \ci{0.33}{0.80} for \qwen. \fluxk's missing direction between 20 and 30 is thus no direction between twenty and twenty-five and a reversal for thirty; taken alone, its 20/25 order (0.50 over 16 pairs) is above the 2.5th percentile of the matched $s{=}0.45$ estimator for that pair (0.38). \qwen's two seeds disagree (0.46 and 0.74), which 16 types per seed cannot resolve.

\para{Models compared.} Paired on the 16 shared U2 types at 12--25, \fluxk and \qwen order their crowds more than \sdm ($+$0.16 \ci{$+$0.02}{$+$0.32} and $+$0.21 \ci{$+$0.03}{$+$0.40}) and than \flux ($+$0.26 \ci{$+$0.04}{$+$0.48} and $+$0.31 \ci{$+$0.09}{$+$0.52}); \sdxl cannot be separated from either group, nor \sdm from \flux ($+$0.10 \ci{$-$0.03}{$+$0.22}). By the sliding reach ($s^*{=}1$) the models rank \flux 8, \sdxl 12, \sdm 16, \fluxk and \qwen 25, but only \flux and \qwen are separated (difference $-$4 request steps \ci{$-$4}{$-$1}); the adjacent-pair reach separates none. A loose estimator itself is assigned a sliding reach below 30 in 34--41\% of 16-type sets at $s{=}0.45$ and in 3--7\% at $s{=}0.30$, so with 16 types the reach is located coarsely. The cap $K$ separates \flux and \sdm from \sdxl, \fluxk and \qwen (all six paired ratios across the two groups exclude 1, none within them).

\para{Does a number's change repeat across seeds?} In \sdm on U2 (64 types, two seeds, nine requests; \samt), the change of $\log(1{+}n)$ from one request to another on seed 1 correlates with the same change on seed 2 at $+$0.16 \ci{0.03}{0.27} past ten (Spearman $+$0.23 \ci{0.11}{0.34}; 33 of the 36 pairs of requests above zero), and $+$0.21 \ci{0.03}{0.39} without the types \samt ever reads as empty. In a variance split of $\log(1{+}n)$ at 12--30, the object type takes 66\%, the noise 8\%, the request 0.2\% \ci{0}{1.3}, type by request 4.4\% \ci{0.9}{8.2} and the residual 22\%: 16\% \ci{4}{27} of a same-noise change repeats on the other seed (39\% \ci{16}{57} at 12, 20 and 30; 30\% within 2--8). The repeated part differs in sign between types and has no common direction. Errors of \samt that depend on the type and the request together could mimic part of it.

\para{How much direction is ruled out?} Over the nine requests from 12 to 30 and two seeds (\samt, 64 types), \sdm's order is 0.502 with a 90\% interval of \ci{0.461}{0.543}, so an order above 0.54 is ruled out at the 5\% level; a loose estimator with a scatter up to 2.47 as drawn (2.14 read by \samt) would order these crowds above it. The number/seed ratio is 0.888 with a 90\% interval of \ci{0.814}{0.975}. Paired and unpaired orders (requests compared across the two seeds) agree within 0.04; past ten the unpaired interval is even slightly narrower, since same-noise pairs tie more often.

\section{A new number against a new seed}
\label{sec:supp_reseed}
\subsection{\sdm, by \samt}
\label{sec:supp_reseedsam}
The test reuses the word test's plain U2 images of \sdm (64 types, seeds 1 and 2) at $768^2$ (requests 2, 3, 4, 6, 8 and the nine requests 12, 14, 16, 19, 20, 22, 25, 27, 30) and at $1024^2$ (2, 4, 8 and 12, 20, 30). For a range of requests, the \emph{number change} is the mean $|\Delta\log(1{+}n)|$ over all pairs of the range's requests from the same noise (both seeds), the \emph{seed change} the mean between seeds 1 and 2 at the same request, the order the share of differing pairs in which the larger request gives more objects, and the signed shift the mean $\Delta\log(1{+}n)$ from the smaller to the larger request. Intervals: type-cluster bootstrap; the ratio's replicates divide the two means of the same resample. The loose estimator's ratios and orders on the same requests are in \cref{sec:supp_refmodels}.

\para{Number, seed and object type.} Over the nine requests at $768^2$ the number change is 0.46 against a seed change of 0.52; over 12, 20 and 30 at $1024^2$, 0.49 against 0.57. Across 12--30 the request explains 0.004 of the variance of $\log(1{+}n)$ at $768^2$ and 0.012 at $1024^2$ (between-request variance over total), the object type 0.71 and 0.72.

\begin{table}[h]
\centering
\footnotesize
\setlength{\tabcolsep}{2.0pt}
\resizebox{\linewidth}{!}{%
\begin{tabular}{@{}lllccc@{}}
\toprule
& requests & counter & number / seed & order & signed shift \\
\midrule
$768^2$ & 2, 4, 8 & \samt & 2.01 \ci{1.59}{2.63} & .89 \ci{.83}{.94} & $+$.42 \\
 & & \gdino, \owl, cons. & 2.04, 1.85, 2.94 & .82, .86, .89 & $+$.34, $+$.44, $+$.38 \\
 & 2, 3, 4, 6, 8 & \samt & 1.73 \ci{1.42}{2.17} & .86 \ci{.81}{.91} & $+$.31 \\
 & 12, 20, 30 & \samt & 0.96 \ci{0.76}{1.19} & .46 \ci{.37}{.55} & $-$.10 \ci{$-$.22}{.01} \\
 & & \gdino, \owl, cons. & 1.00, 0.92, 0.85 & .36, .42, .29 & $-$.26, $-$.18, $-$.21 \\
 & nine, 12--30 & \samt & 0.89 \ci{0.80}{0.99} & .50 \ci{.45}{.55} & $-$.01 \ci{$-$.07}{.05} \\
 & & \gdino, \owl, cons. & 0.85, 0.86, 0.89 & .42, .48, .40 & $-$.15, $-$.06, $-$.13 \\
$1024^2$ & 2, 4, 8 & \samt & 1.76 \ci{1.46}{2.19} & .87 \ci{.81}{.92} & $+$.45 \\
 & & \gdino, \owl, cons. & 1.97, 1.49, 2.45 & .81, .87, .90 & $+$.34, $+$.44, $+$.42 \\
 & 12, 20, 30 & \samt & 0.85 \ci{0.69}{1.05} & .38 \ci{.31}{.46} & $-$.19 \ci{$-$.30}{$-$.08} \\
 & & \gdino, \owl, cons. & 0.99, 0.88, 1.05 & .30, .40, .20 & $-$.27, $-$.17, $-$.36 \\
\bottomrule
\end{tabular}}
\caption{\textbf{Number change against seed change} (\sdm, U2, 64 types, two seeds). 384 number and 192 seed comparisons per row of three requests, 1{,}280 and 320 for five, 4{,}608 and 576 for nine; fewer under the consensus (cons.), which leaves images without a majority uncounted. The spelling control at 12, 20 and 30 is in \cref{tab:supp_spelling}.}
\label{tab:supp_reseed}
\end{table}

Within the reach a new number moves the crowd about twice as far as a new seed, toward the number; past it, about as far as a new seed and not toward the number: at $1024^2$ and under the other counters the orders past ten are reversals (\cref{tab:supp_reseed}).

\subsection{People's check}
\label{sec:supp_reseedpeople}
People counted the plain images of the 16 people-counted U2 types at both seeds, at 2, 4, 8, 12, 20 and 30, and the number in digits (seed 1) at 12, 20 and 30 (sixth count, \cref{sec:supp_hcstudies}). The \emph{noise-free ratio} uses the two independent first counts: with $d_1$ and $d_2$ the same difference of $\log(1{+}n)$ computed from R1's and from R2's counts, $E[d_1d_2]$ is the squared difference of the pictures' true counts, and the ratio is $\sqrt{\overline{d_1d_2}_{\text{number}}/\overline{d_1d_2}_{\text{seed}}}$. People agree with \samt (\cref{tab:supp_reseed_people}): within the reach the ratio's interval lies above 1, toward the number; past it, the intervals of both the ratio and the noise-free ratio lie below the loose estimator's ratio at these requests (1.47 at the larger scatter, \cref{sec:supp_refmodels}), and the order is near one half.

\begin{table}[h]
\centering
\footnotesize
\setlength{\tabcolsep}{2.0pt}
\resizebox{\linewidth}{!}{%
\begin{tabular}{@{}llccccc@{}}
\toprule
& requests & number & seed & number / seed & noise-free & order \\
\midrule
people & 2, 4, 8 & .52 & .19 & 2.77 \ci{1.87}{4.09} & 2.04 \ci{0.96}{4.10} & .92 \ci{.80}{1.00} \\
 & 12, 20, 30 & .48 & .53 & 0.90 \ci{0.60}{1.29} & 0.93 \ci{0.41}{1.28} & .52 \ci{.35}{.69} \\
\samt & 2, 4, 8 & .57 & .35 & 1.63 \ci{1.16}{2.59} & & .82 \ci{.67}{.95} \\
 & 12, 20, 30 & .50 & .69 & 0.72 \ci{0.42}{1.06} & & .51 \ci{.34}{.69} \\
\bottomrule
\end{tabular}}
\caption{\textbf{People's check of the number against the seed} (\sdm, 16 types, seeds 1 and 2). Mean $|\Delta\log(1{+}n)|$ when the number changes and when the seed changes; small over large ratio: people 3.07 \ci{1.80}{5.10}, \samt 2.25 \ci{1.29}{4.71}. Digits against words at 12, 20 and 30 (seed 1): people .52, \samt .50.}
\label{tab:supp_reseed_people}
\end{table}

\subsection{Value, spelling and seed}
\label{sec:supp_spelling}
From the same noise (seed 1) at requests 12, 20 and 30, the mean $|\Delta\log(1{+}n)|$ when the number word changes (the three pairs of ``twelve'', ``twenty'', ``thirty''), when only its spelling changes (``twelve'' against ``12'', and likewise for twenty and thirty), and when the seed changes (seeds 1 and 2, same request); value order: among value changes that alter the count, the share in which the larger request gives more objects. Past the reach, a new number, a new spelling and a new seed move \sdm's crowd about equally far under every counter, and every value order is below one half, a reversal (\cref{tab:supp_spelling}).

\begin{table}[h]
\centering
\footnotesize
\setlength{\tabcolsep}{2.6pt}
\resizebox{\linewidth}{!}{%
\begin{tabular}{@{}lcccc@{}}
\toprule
\sdm, U2 & value & spelling & seed & value order \\
\midrule
\samt & .51 \ci{.39}{.63} & .48 \ci{.38}{.62} & .49 \ci{.40}{.60} & .41 \ci{.32}{.51} \\
\gdino & .53 & .44 & .54 & .40 \\
\owl & .60 & .57 & .59 & .43 \\
consensus & .33 & .37 & .39 & .25 \\
\bottomrule
\end{tabular}}
\caption{\textbf{Value, spelling and seed} (mean change of $\log(1{+}n)$; 192 comparisons per entry, 74--81 under the consensus, which leaves images without a majority uncounted; value order over 133, 131, 149 and 32 differing pairs).}
\label{tab:supp_spelling}
\end{table}

\subsection{Layout from the same noise}
\label{sec:supp_layout}
The layout comparison uses the plain images of \cref{sec:supp_reseedsam} at $768^2$ (64 types; requests 2, 4, 8, 12, 20 and 30; seeds 1 and 2) and the number in digits of \cref{sec:supp_spelling} (12, 20 and 30; seed 1), 960 images in all. Each image is reduced to two $24{\times}24$ maps: a grayscale thumbnail, each cell the mean gray level of a $32{\times}32$ block of pixels, and a map of the objects' boxes, each cell the share of it covered by \samt's boxes (score $\geq0.6$). Two images are compared by the Pearson correlation of their thumbnails, or of their box maps, over the 576 cells. The pairs are: a new number, every pair of requests within 2, 4, 8 and within 12, 20, 30 from the same noise (both seeds; 384 pairs per range); a new seed, seeds 1 and 2 at the same request (192); digits, the number in words against the same number in digits from the same noise (seed 1; 192); and, as a floor, each type against the next type of the list at the same request and seed (seed 1; 192). Box-map pairs in which either map is constant are left out (at 12, 20, 30, 270 of the 384 number pairs and 133 of the 192 seed pairs remain). Each entry is the mean correlation over its pairs, with a 95\% interval from the type bootstrap (20{,}000 resamples of the 64 types, the same indices for every entry); the paired difference takes, for each type, the mean over its number pairs minus the mean over its seed pairs.

From the same noise, a new number keeps the picture's layout and much of the objects' placement (\cref{tab:supp_layout}): past the reach the thumbnails correlate at 0.71 \ci{0.68}{0.75} for a new number and 0.76 for digits, against 0.10 \ci{0.04}{0.17} for a new seed and 0.02 between types, and the box maps at 0.58 for a new number against 0.32 for a new seed. Yet past the reach a new number changes the count about as much as a new seed does (\cref{tab:supp_reseed}).

\begin{table}[h]
\centering
\footnotesize
\setlength{\tabcolsep}{2.4pt}
\resizebox{\linewidth}{!}{%
\begin{tabular}{@{}lcccc@{}}
\toprule
& \multicolumn{2}{c}{thumbnails} & \multicolumn{2}{c}{box maps} \\
\cmidrule(lr){2-3}\cmidrule(lr){4-5}
pair & 12, 20, 30 & 2, 4, 8 & 12, 20, 30 & 2, 4, 8 \\
\midrule
a new number & .71 \ci{.68}{.75} & .62 \ci{.59}{.66} & .58 \ci{.52}{.64} & .55 \ci{.49}{.60} \\
digits & .76 \ci{.73}{.79} & & .64 \ci{.58}{.70} & \\
a new seed & .10 \ci{.04}{.17} & .19 \ci{.12}{.26} & .32 \ci{.26}{.39} & .48 \ci{.43}{.53} \\
another type & .02 \ci{$-$.02}{.07} & .06 \ci{.00}{.12} & .20 \ci{.15}{.25} & .30 \ci{.25}{.35} \\
\midrule
number $-$ seed & $+$.61 \ci{.55}{.68} & $+$.43 \ci{.36}{.51} & $+$.24 \ci{.18}{.30} & $+$.08 \ci{.03}{.12} \\
\bottomrule
\end{tabular}}
\caption{\textbf{Layout from the same noise} (\sdm, $768^2$, U2, 64 types; Pearson correlation of two images' $24{\times}24$ grayscale thumbnails or \samt box maps, mean over pairs with 95\% intervals over types). A new number: two requests from the same noise; digits: the same noise and number, in words and in digits (seed 1); a new seed: the same request at seeds 1 and 2; another type: the next type of the list, same request and seed (a floor). Number $-$ seed: per type, the mean over its number pairs minus the mean over its seed pairs (box maps: the 50 and 54 types with both kinds of pairs).}
\label{tab:supp_layout}
\end{table}

\subsection{A new number against a new seed, digits and an edit}
\label{sec:supp_fourarm}
\para{Design.} From the noise of each type's plain image (seed 1), four changes are compared: a new number (pairs of different requests of the range, plain prompts), a new seed (seed 1 against seed 2, same request), the same number in digits (``a photo of 12 $X$s'') and an edit elsewhere in the prompt (``a photograph of twelve $X$s''); each change is the mean $|\Delta\log(1{+}n)|$ against the plain image of the same type and request. \sdm: 64 types at $768^2$ (40 steps, guidance 4.5), within the reach 2, 4, 8 and past it 12, 20, 30 (the digits at 12, 20, 30 are those of \cref{tab:supp_spelling}); \flux: the stream test's 32 types and settings, within 2, 4 and past 12, 25 (832 new images). The main quantity is the ratio of a new number's change to the edit's (\samt), within and past the reach.
\begin{table*}[h]
\centering
\footnotesize
\setlength{\tabcolsep}{3pt}
\begin{tabular}{@{}llcccccccc@{}}
\toprule
& & \multicolumn{4}{c}{mean $|\Delta\log(1{+}n)|$} & & & & \\
\cmidrule(lr){3-6}
model, requests & counter & number & seed & digits & edit & number / seed & number / edit & edit / seed & order of number \\
\midrule
\sdm, 2, 4, 8 & \samt & 0.63 & 0.29 & 0.25 & 0.18 & 2.17 \ci{1.72}{2.81} & 3.50 \ci{2.69}{4.84} & 0.62 \ci{0.44}{0.82} & .91 \ci{.85}{.96} \\
 & \gdino & 0.53 & 0.25 & 0.24 & 0.20 & 2.08 & 2.66 & 0.78 & .80 \\
 & \owl & 0.64 & 0.34 & 0.29 & 0.18 & 1.85 & 3.52 & 0.53 & .87 \\
 & consensus & 0.58 & 0.18 & 0.15 & 0.10 & 3.25 & 5.53 & 0.59 & .89 \\
\midrule
\sdm, 12, 20, 30 & \samt & 0.51 & 0.49 & 0.48 & 0.42 & 1.03 \ci{0.79}{1.30} & 1.22 \ci{0.97}{1.56} & 0.84 \ci{0.63}{1.12} & .41 \ci{.32}{.51} \\
 & \gdino & 0.52 & 0.54 & 0.44 & 0.38 & 0.96 & 1.40 \ci{1.10}{1.81} & 0.69 & .40 \ci{.29}{.51} \\
 & \owl & 0.60 & 0.59 & 0.57 & 0.44 & 1.01 & 1.36 \ci{1.07}{1.71} & 0.74 & .43 \ci{.33}{.53} \\
 & consensus & 0.33 & 0.39 & 0.37 & 0.22 & 0.84 & 1.49 \ci{0.93}{2.54} & 0.57 & .25 \ci{.07}{.43} \\
\midrule
\flux, 2, 4 & \samt & 0.53 & 0.17 & 0.18 & 0.15 & 3.03 \ci{2.19}{4.83} & 3.44 \ci{2.34}{5.67} & 0.88 \ci{0.53}{1.48} & .84 \ci{.68}{.96} \\
 & \gdino & 0.49 & 0.22 & 0.17 & 0.16 & 2.20 & 3.19 & 0.69 & .87 \\
 & \owl & 0.64 & 0.35 & 0.29 & 0.25 & 1.80 & 2.56 & 0.70 & .75 \\
 & consensus & 0.56 & 0.16 & 0.14 & 0.15 & 3.54 & 3.64 & 0.97 & .86 \\
\midrule
\flux, 12, 25 & \samt & 0.58 & 0.42 & 0.45 & 0.37 & 1.40 \ci{0.87}{2.09} & 1.59 \ci{0.81}{3.22} & 0.88 \ci{0.44}{1.66} & .45 \ci{.23}{.67} \\
 & \gdino & 0.60 & 0.54 & 0.43 & 0.40 & 1.13 & 1.51 \ci{0.85}{2.62} & 0.74 & .26 \ci{.09}{.45} \\
 & \owl & 0.69 & 0.55 & 0.72 & 0.47 & 1.26 & 1.45 \ci{0.87}{2.56} & 0.87 & .46 \ci{.27}{.65} \\
 & consensus & 0.45 & 0.26 & 0.32 & 0.32 & 1.74 & 1.41 \ci{0.51}{4.89} & 1.23 & .20 \ci{.00}{.50} \\
\bottomrule
\end{tabular}
\caption{\textbf{Four changes from the same noise} (seed 1; the seed change compares seeds 1 and 2; 95\% intervals over types). Number: a new request of the range; digits: ``a photo of 12 $X$s''; edit: ``a photograph of twelve $X$s''. Order: share of differing pairs in which the larger request gives more objects.}
\label{tab:supp_fourarm}
\end{table*}

\para{Results.} Within the reach, a new number moves the crowd 1.80--3.54 times as far as a new seed and 2.56--5.53 times as far as the edit under the four counters, toward the number (orders 0.75--0.91). Past the reach, in \sdm a new number moves it about as far as a new seed (0.84--1.03 under the four counters) and 1.22--1.49 times as far as the edit, without direction (order .41 \ci{.32}{.51} by \samt; the larger request gives slightly fewer objects on average, signed shift $-$0.16 \ci{$-$0.30}{$-$0.02}). In \flux, on 32 types, it moves it 1.40 \ci{0.87}{2.09} times as far as a new seed and 1.59 \ci{0.81}{3.22} times as far as the edit, also without direction (order .45 \ci{.23}{.67}; .26 by \gdino). The edit alone moves a crowd past the reach 0.57--0.85 times as far as a new seed in \sdm and 0.74--1.23 times in \flux. Shares of counts left unchanged by a new number, a new seed, digits and the edit (\samt): \sdm within the reach 17\%, 49\%, 56\%, 66\%, past it 31\%, 26\%, 35\%, 36\%; \flux within 22\%, 73\%, 77\%, 75\%, past 38\%, 45\%, 52\%, 61\%. Signed shifts past the reach (\samt): digits $-$0.20 \ci{$-$0.33}{$-$0.04} and the edit $+$0.10 \ci{$-$0.02}{$+$0.22} in \sdm; digits $+$0.10 \ci{$-$0.11}{$+$0.32} and the edit $+$0.20 \ci{$+$0.00}{$+$0.45} in \flux.

\subsection{The four changes in \fluxk and \qwen, counted by people}
\label{sec:supp_fourarm2}
\para{Design.} The four changes of \cref{sec:supp_fourarm}, from the noise of each type's plain image of the word test (seed 1), with each model's word-test sampler: \fluxk base 4B at $1024^2$ (50 steps, guidance 4.0) on the 32 types of \flux's four-change set (the 16 people-counted U2 types and 16 drawn from the other 48), within its reach at 4, 8 and 12 and at its edge at 20, 25 and 30 (576 new images); \qwen (transformer in NF4, $1024^2$, 50 steps, true guidance 4.0) on the 16 people-counted types at the edge of its reach, at 20, 25 and 30 (144 new images). Both models' crowds stop growing at about 13--15 objects (cap $K$, \cref{tab:supp_ceiling}), while their orders follow the request up to twenty-five (\cref{sec:reach}), so twelve lies inside their reach and 20--30 at its edge. Every new seed-1 image starts from the plain image's own latent (checked on its logged start: 384 of 384 for \fluxk; \qwen's 144 noise seeds as logged), and each host first reproduced two stored word-test images pixel for pixel. People counted the 16 people-counted types at 20--30, all four images per cell (seventh count, \cref{sec:supp_hcstudies}).
\begin{table*}[h]
\centering
\footnotesize
\setlength{\tabcolsep}{3pt}
\begin{tabular}{@{}llcccccccc@{}}
\toprule
& & \multicolumn{4}{c}{mean $|\Delta\log(1{+}n)|$} & & & & \\
\cmidrule(lr){3-6}
model, requests & counter & number & seed & digits & edit & number / seed & number / edit & edit / seed & order of number \\
\midrule
\fluxk, 4, 8, 12 & \samt & 0.57 & 0.40 & 0.38 & 0.23 & 1.41 \ci{0.88}{2.28} & 2.43 \ci{1.57}{4.52} & 0.58 \ci{0.30}{0.96} & .89 \ci{.78}{.97} \\
 & \gdino & 0.56 & 0.39 & 0.30 & 0.30 & 1.45 & 1.90 & 0.76 & .75 \\
 & \owl & 0.55 & 0.51 & 0.49 & 0.28 & 1.08 & 1.93 & 0.56 & .83 \\
 & consensus & 0.50 & 0.24 & 0.23 & 0.13 & 2.11 & 3.90 & 0.54 & .93 \\
\midrule
\fluxk, 20, 25, 30 & \samt & 0.52 & 0.70 & 0.74 & 0.50 & 0.74 \ci{0.45}{1.17} & 1.04 \ci{0.61}{1.67} & 0.72 \ci{0.51}{0.94} & .46 \ci{.36}{.55} \\
 & \gdino & 0.65 & 0.57 & 0.65 & 0.56 & 1.14 \ci{0.84}{1.51} & 1.16 \ci{0.79}{1.67} & 0.98 \ci{0.73}{1.32} & .47 \ci{.34}{.60} \\
 & \owl & 0.69 & 0.87 & 0.71 & 0.73 & 0.80 \ci{0.60}{1.04} & 0.95 \ci{0.68}{1.31} & 0.84 \ci{0.61}{1.10} & .45 \ci{.33}{.57} \\
 & consensus & 0.17 & 0.30 & 0.35 & 0.51 & 0.58 & 0.34 & 1.70 & .71 \\
 & people$^b$ & 0.66 & 0.92 & 0.83 & 0.86 & 0.72 \ci{0.37}{1.21} & 0.77 \ci{0.52}{1.08} & 0.93 \ci{0.64}{1.34} & .43 \ci{.28}{.58} \\
\midrule
\qwen, 20, 25, 30 & \samt & 0.46 & 0.49 & 0.37 & 0.32 & 0.93 \ci{0.66}{1.38} & 1.44 \ci{0.85}{3.31} & 0.65 \ci{0.34}{1.05} & .65 \ci{.48}{.81} \\
 & \gdino & 0.33 & 0.64 & 0.43 & 0.28 & 0.52 \ci{0.21}{1.07} & 1.17 \ci{0.50}{2.32} & 0.44 \ci{0.28}{0.62} & .31 \ci{.10}{.54} \\
 & \owl & 0.43 & 0.68 & 0.55 & 0.51 & 0.63 \ci{0.29}{1.10} & 0.84 \ci{0.42}{1.71} & 0.75 \ci{0.33}{1.36} & .45 \ci{.29}{.63} \\
 & consensus & 0.33 & 0.31 & 0.39 & 0.11 & 1.08 & 3.07 & 0.35 & .57 \\
 & people$^b$ & 0.31 & 0.41 & 0.44 & 0.21 & 0.76 \ci{0.30}{1.54} & 1.46 \ci{0.54}{4.32} & 0.52 \ci{0.20}{1.17} & .46 \ci{.27}{.64} \\
\bottomrule
\end{tabular}
\caption{\textbf{Four changes in \fluxk and \qwen} (as \cref{tab:supp_fourarm}; \fluxk 32 types, \qwen 16; 95\% intervals over types). $^b$People's final counts of the 16 people-counted types (the seventh count), on the same pictures as \samt's 16-type values in the text.}
\label{tab:supp_fourarm2}
\end{table*}

\para{\fluxk.} Within its reach, a new number moves the crowd 1.08--2.11 times as far as a new seed and 1.90--3.90 times as far as the edit under the four counters, toward the number (orders 0.75--0.93); it leaves 36\% of the counts unchanged, the edit 70\% (\samt). At the edge of its reach (20--30), a new number moves the crowd 0.58--1.14 times as far as a new seed and 0.34--1.16 times as far as the edit, and not toward the number (orders 0.45--0.47 by \samt, \gdino and \owl; 0.71 by the consensus, which counts few of these crowds); the signed shift of a new number is $-$0.003 \ci{$-$0.19}{0.21}. Written in digits, the number gives fewer objects at 20--30 (signed shift $-$0.27 \ci{$-$0.53}{$-$0.01}; people's medians 5.5--7.5 objects against 12.5--20 for the words). On the 16 people-counted types, people find a new number moving the crowd 0.72 \ci{0.37}{1.21} times as far as a new seed and 0.77 \ci{0.52}{1.08} times as far as the edit, with an order of 0.43 \ci{0.28}{0.58} (0.40 \ci{0.27}{0.53} over both seeds, where the number change compares the requests of either seed), and \samt on the same pictures 0.86, 1.13 and 0.38 \ci{0.26}{0.49}. The first two people's counts agree on 48--52\% of these pictures, so the ratio with the raters' disagreement removed is too imprecise to report (0.52 \ci{0.00}{3.25}).

\para{\qwen.} At the edge of its reach (20--30), a new number moves the crowd 0.52--1.08 times as far as a new seed and 0.84--3.07 times as far as the edit under the four counters, which disagree on its direction: by \samt the larger request gives more objects in 0.65 \ci{0.48}{0.81} of the differing pairs (signed shift $+$0.30 \ci{0.03}{0.61}), by \gdino in 0.31 and by \owl in 0.45. By people (16 types), a new number moves the crowd 0.76 \ci{0.30}{1.54} times as far as a new seed and 1.46 \ci{0.54}{4.32} times as far as the edit (0.67 \ci{0.28}{1.04} and 1.20 with the raters' disagreement removed), with an order of 0.46 \ci{0.27}{0.64} on seed 1 and 0.60 \ci{0.48}{0.72} over both seeds, and a signed shift of $-$0.01 \ci{$-$0.08}{0.06}; the first two people's counts agree on 75--83\% of these pictures.

\para{A loose estimator at 20, 25 and 30.} Between requests that differ by at most 1.5 times, the number/seed ratio cannot separate estimating from ignoring: the loose estimator of \cref{sec:supp_refmodels}, simulated on this design (16 types; seed-1 number pairs, a new seed, and an edit that redraws the estimate at the same request; 4{,}000 sets, \cref{sec:supp_noise}), moves its count 1.10--1.22 times as far as a new seed as drawn, 1.08--1.16 times counted by people with the difficulty of \fluxk's or \qwen's pictures (2.5th percentiles 0.74--0.76), 1.05--1.08 counted once by one person, and 1.03--1.04 read by \samt. Its order separates it: 0.66--0.74 as drawn, 0.65--0.72 counted by people (2.5th percentiles 0.50 and 0.58 at $s{=}0.45$ and 0.30), 0.63--0.68 counted once (0.47--0.48 and 0.53), and 0.59--0.60 read by \samt (0.43; on 32 types 0.47--0.48). \fluxk's people's order (0.43) lies below the 2.5th percentile of every people-counted reference, the single-count bound included, and its \samt order on 32 types (0.46) below the \samt reference's; its people's ratio (0.72) lies below the people-counted references' 2.5th percentiles but not below the single-count bound. Over both seeds, the people-counted estimator orders these crowds at 0.65 \ci{0.54}{0.76} and 0.72 \ci{0.62}{0.82} ($s{=}0.45$ and 0.30): \fluxk's 0.40 lies below both, \qwen's 0.60 inside the first (with \fluxk's counting difficulty: 0.66 \ci{0.55}{0.76} and 0.72 \ci{0.62}{0.81}). \qwen's seed-1 order (0.46) lies below the people-counted references' 2.5th percentiles and its \samt order (0.65) inside the \samt reference's range; its ratio with the raters' disagreement removed (0.67) lies below the estimator's ratio as drawn (2.5th percentiles 0.82 and 0.91).

\subsection{\flux}
\label{sec:supp_fluxreseed}
\para{By \samt.} On the plain cells of the stream test (\cref{sec:supp_stream}; the plain images of the smaller and the larger request from the same noise, and seed 1 against seed 2 in the same cell; 32 types), a new number moves \flux's crowd 2.51 \ci{1.78}{4.27} times as far as a new seed between two and four (order 0.83 \ci{0.69}{0.94}) and 1.25 \ci{0.87}{1.81} times between twelve and twenty-five (order 0.53 \ci{0.33}{0.71}, mean shift $-$0.17 \ci{$-$0.41}{0.07}); by \gdino, \owl and the consensus the ratios past ten are 0.99, 1.22 and 1.87 and the orders 0.25 \ci{0.11}{0.41}, 0.51 and 0.35, under \gdino and the consensus a reversal. On the same cells, \sdm's ratios are 1.70 \ci{1.27}{2.44} (order 0.91) and 1.07 \ci{0.85}{1.33} (order 0.53).

\para{By people.} People counted the same plain cells of the 16 people-counted types at both seeds (sixth count, \cref{sec:supp_hcstudies}): a new number moves \flux's crowd 1.64 \ci{1.13}{2.52} times as far as a new seed between twelve and twenty-five (2.35 \ci{1.51}{4.84} between two and four), but not consistently toward the number (people's order 0.66 \ci{0.43}{0.86}).

Past its reach, a new number moves \flux's crowd further than a new seed does by people's counts (by \samt, about as far or further), with no reliable direction.

\subsection{Redrawing only the fine-scale noise}
\label{sec:supp_brs}
\para{Design.} From the starting noise of a type's seed-1 image, $z_1$, and of its seed-2 image, $z_2$, we form a start that keeps $z_1$ below $f$ cycles per image and takes $z_2$ above it: with $F_1$, $F_2$ the 2D Fourier transforms of the two noises over the spatial axes, $r=\sqrt{k_x^2+k_y^2}$ the radial frequency in cycles per image, and $M=\mathbf{1}[r\le f]$,
\begin{equation}
z_f=\mathcal{F}^{-1}\big[M F_1+(1-M)F_2\big],
\end{equation}
computed in double precision and cast back to the sampler's type. The two masks are complementary, so $z_f$ is again white Gaussian noise. With $f$ above the highest frequency the start equals $z_1$ bit for bit, and with $f<0$ it equals $z_2$; on every host and model, both settings reproduced the stored seed-1 and seed-2 images pixel for pixel before any new image was drawn. Everything else is the word test's unchanged sampler and plain prompt. Cells: \sdm (U2, $768^2$) at requests 2, 4, 8, 12, 20 and 30 with $f$ = 2, 4 and 8 on all 64 types and $f$ = 16 on the 16 annotated types (1{,}248 images); \fluxk at 4, 8, 12, 20, 25 and 30 with $f$ = 2, 4 and 8 on the 16 annotated types (288); \qwen at 4, 12, 20 and 30 with $f$ = 4 on the 16 annotated types (64); and \fluxk's plain prompt from the word test's noise at ``twenty-two'', ``twenty-seven'', ``thirty-five'' and ``forty'' (64). Every image is counted by \samt, \gdino and \owl. Measures, over types: the count change of the redraw against the seed-1 image at the same request, of a new seed (seed 2 against seed 1), and of a new number (seed 1; over the neighbouring requests for a single request, and over all pairs of a range's requests for a range), each as the mean $|\Delta\log(1{+}n)|$, their ratios, and the correlation of $24{\times}24$ grayscale thumbnails with the seed-1 image.

\begin{figure*}[h]
\centering
\includegraphics[width=0.92\linewidth]{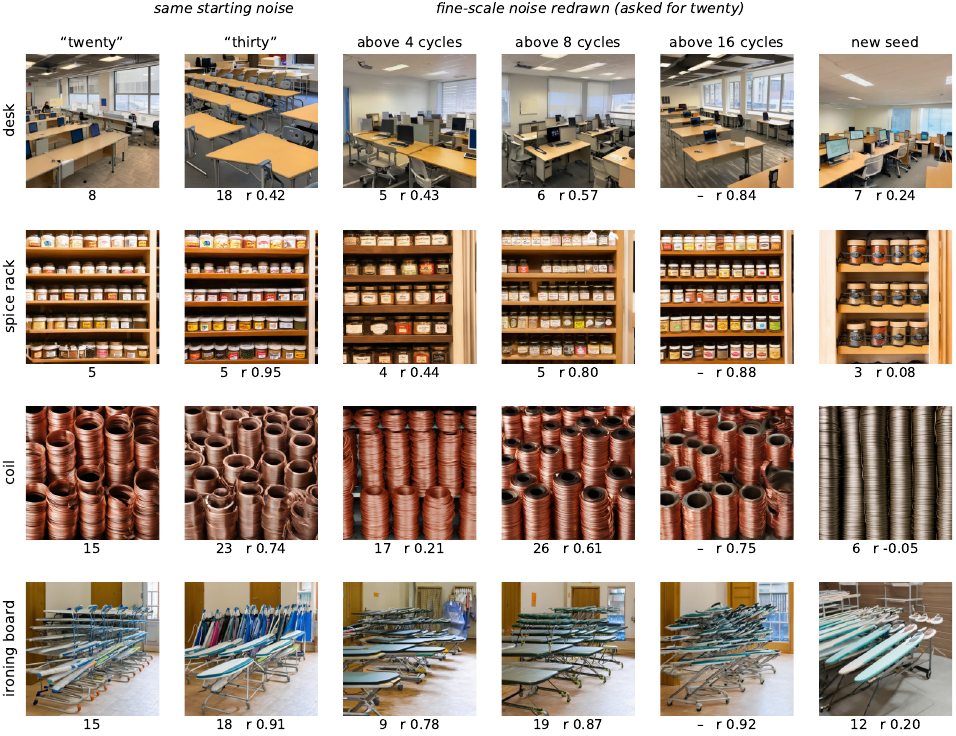}
\caption{\textbf{Fine-scale redraws next to a new number and a new seed} (\sdm, $768^2$; four of the 16 people-counted types, drawn at random). From the noise of the image asked for twenty (first column): thirty asked from the same noise; the noise above 4, 8 and 16 cycles per image redrawn from the type's second seed, still asked for twenty; and twenty from the second seed. Below each image: people's final count (people did not count the 16-cycle redraws) and the correlation of its $24{\times}24$ grayscale thumbnail with the first column's.}
\label{fig:supp_redraw}
\end{figure*}

\para{\sdm.} Past the reach (12, 20, 30) a new number moves the count by 0.51 and a new seed by 0.49 (\samt). Redrawing above $f$ = 2, 4 and 8 cycles moves it 0.97 \ci{0.79}{1.21}, 0.67 \ci{0.51}{0.89} and 0.66 \ci{0.51}{0.83} times as far as a new number, while keeping the layout at a thumbnail correlation of 0.40, 0.54 and 0.74 (new number 0.70 \ci{0.65}{0.74}, new seed 0.10). \gdino and \owl agree (0.69 and 0.74 at $f$ = 8). Within the reach (2, 4, 8) the redraws move the count only 0.31--0.40 times as far as a new number, though 0.67--0.87 times as far as a new seed, which moves small crowds little. On the 16 annotated types, a redraw above 16 cycles still moves a large crowd 0.71 \ci{0.49}{0.98} times as far as a new number and 0.60 \ci{0.35}{0.95} times as far as a new seed, with a thumbnail correlation of 0.86 (\cref{fig:supp_redraw}). Without the 26 types that \samt ever reads as empty, the ratios past the reach are 0.98, 0.79 and 0.68. The ratio to a new number rises with the request, as the number's own effect fades; the ratio to a new seed does not.

\para{\fluxk and \qwen.} \samt reads many of \fluxk's crowds as empty (10 of the 16 types, up to half of the images of a cell), which blurs its readings. By \gdino and \owl, the redraws move \fluxk's count 0.43--0.75 times as far as a new number within its reach (4, 8, 12) and 0.74--1.10 times at its edge (20, 25, 30), the pattern of \sdm. In \qwen, a redraw above 4 cycles moves the count 0.44 \ci{0.20}{0.63} times as far as a new number at 4 and 12 and 0.82 \ci{0.59}{1.38} times at 20 and 30 (\samt), keeping the layout at 0.59 against 0.70 for a new number. \fluxk's plain prompt at the new number words: by \samt and \owl, ``thirty'' draws fewer objects than the mean of ``twenty-seven'' and ``thirty-five'' (differences in $\log(1{+}n)$ $-$0.29 \ci{$-$0.86}{0.22} and $-$0.34 \ci{$-$0.76}{0.03}; by \owl, twenty-seven and thirty-five draw about as many as twenty-five), and ``forty'' draws fewer than ``twenty-five'' ($-$0.74 \ci{$-$1.41}{$-$0.15} and $-$0.40 \ci{$-$0.86}{$-$0.02}); \gdino shows no drop at thirty. Part of the drop at thirty may thus belong to the word itself, and these 16 types cannot settle how much; \samt reads half of the ``forty'' crowds as empty.

\para{People's counts.} People counted the redraws above four and eight cycles on the 16 people-counted types: \sdm's at 4, 8, 12, 20 and 30 and \fluxk's at 4, 8, 12, 20, 25 and 30 (352 pictures; eighth count, \cref{sec:supp_hcstudies}). The plain images they are compared with were counted earlier under the same protocol: \sdm's seeds 1 and 2 in the sixth count, \fluxk's seed 1 in the second count and its seed 2 (at 20, 25 and 30) in the seventh. The measures are those above, over the requests people counted (for \sdm within the reach, four and eight only). In \sdm the mean changes of $\log(1{+}n)$ are, within the reach and past it, 0.37 and 0.43 for a new number, 0.25 and 0.53 for a new seed, 0.29 and 0.32 for a redraw above four cycles, and 0.19 and 0.30 above eight cycles; two final counts of the same picture from different studies differ by 0.12 at both (\cref{sec:supp_noise}). In \sdm, past the reach, a redraw above eight cycles moves the count 0.69 \ci{0.38}{1.20} times as far as a new number and 0.57 \ci{0.35}{0.91} times as far as a new seed, and a redraw above four cycles 0.73 \ci{0.42}{1.31} and 0.60 \ci{0.41}{0.87} times; \samt on the same pictures gives 0.51 and 0.66 times a new number. At four and eight objects the redraws move the count 0.52 \ci{0.27}{0.99} ($f$ = 8) and 0.81 \ci{0.50}{1.37} ($f$ = 4) times as far as a new number, against 0.48 and 0.51 by \samt, 0.54 and 0.85 by \gdino, and 0.44 and 0.82 by \owl on the same pictures; request by request, the ratio for $f$ = 8 rises from 0.34 at four objects to 0.96--0.97 at twenty and thirty. In \fluxk, the redraws move the count 0.41 \ci{0.19}{0.70} ($f$ = 8) and 0.52 \ci{0.29}{0.89} ($f$ = 4) times as far as a new number within its reach (4, 8, 12), and 0.74 \ci{0.44}{1.14} and 0.85 \ci{0.63}{1.19} times at its edge (20, 25, 30), where they move it about as far as a new seed (0.99 and 1.12 times). People disagree on whether some of \fluxk's large crowds show the object at all: of the 48 seed-1 pictures at 20--30 that both the second and the seventh count include, 29 get the same count and 6 get zero in one count and five or more in the other. With the seventh count's seed-1 counts, the ratios at 20--30 become 1.15 and 1.47. Without the eight types that have any count of zero (a check we added after seeing the counts), they are 0.18 and 0.21 within the reach and 0.71 and 0.96 at its edge; in \sdm, which loses two types, the ratios change by at most 0.08. The first two counts agree on 55\% of \sdm's redraws and 67\% of \fluxk's. Noise-free ratios from the two independent first counts (\cref{sec:supp_reseed}) are too uncertain here to add anything: half of their intervals reach zero, and the others are wide.

\para{Relative error.} For requests of 12, 16, 20, 25 and 30, a model that always draws sixteen objects has a mean relative error $|n-c|/c$ of 0.27, and the loose estimator of \cref{sec:supp_refmodels}, $\mathrm{round}(c\,e^{sz})$, has 0.38 at $s$ = 0.45 and 0.25 at $s$ = 0.30 (simulated).

\section{The number at the denoiser, and what fails to give it a direction}
\label{sec:supp_heard}\label{sec:supp_index}
\subsection{Probes of the conditioning and of the velocity}
\label{sec:supp_mech}
\sdm, U2, seed 1. For U2's 64 types and 23 number words, a ridge regression trained on the other words orders a held-out word in 0.89 of same-object pairs at 12--30 and 0.98 at 2--8 from \sdm's pooled CLIP vector, and in 0.85--0.97 from every text stream: the number word is decodable, in order, from the conditioning. Along U2's plain trajectories, from $\sigma{=}1$ to 0.909, changing the object word moves the conditional velocity 1.18--1.57 times as far as the guidance direction of the trajectory's own prompt, ``two'' to ``four'' 0.37--0.70 times, ``twelve'' to ``twenty-five'' 0.37--0.42 times and ``twenty'' to ``thirty'' 0.21--0.26 times. These probes are descriptive.

\subsection{Value and spelling at the denoiser}
\label{sec:supp_vdir}
\para{Design.} Along \sdm's own trajectories (the plain images' noise of all 64 types, asked for four and for twenty, seed 1, guidance 4.5, 40 steps; 128 trajectories, each of which reproduces its stored image exactly), at each probe step the conditional velocity $v(c)$ is evaluated at the same state for the prompts ``a photo of $c$ $X$s'' with $c$ = two, four, twelve, twenty, twenty-five, thirty, the digits ``2'', ``12'' and ``20'', twenty of the next type, and the empty prompt. A change of word is the difference of two such velocities; its size is given relative to the guidance direction $v(\text{own prompt})-v(\text{empty})$. Quantities are averaged over the first ten steps ($\sigma\geq0.909$) per trajectory; 95\% intervals over types. The question: does the large number's value move the velocity more than its spelling, $\lVert v(25)-v(12)\rVert/\lVert v(\text{``12''})-v(12)\rVert$?

\begin{table}[!htbp]
\centering
\footnotesize
\setlength{\tabcolsep}{3pt}
\resizebox{\linewidth}{!}{%
\begin{tabular}{@{}lc@{}}
\toprule
early steps ($\sigma\geq0.909$), 128 trajectories & estimate \\
\midrule
value / spelling: twelve $\to$ twenty-five over twelve $\to$ ``12'' & 1.34 \ci{1.25}{1.44} \\
\phantom{value / spelling:} two $\to$ four over two $\to$ ``2'' & 2.79 \ci{2.54}{3.03} \\
\phantom{value / spelling:} twenty $\to$ thirty over twenty $\to$ ``20'' & 0.78 \ci{0.72}{0.84} \\
size relative to guidance: 12$\to$25, 12$\to$``12'' & 0.40, 0.33 \\
\phantom{size relative to guidance:} 2$\to$4, 2$\to$``2''; another object & 0.60, 0.25; 1.45 \\
\bottomrule
\end{tabular}}
\caption{\textbf{Value and spelling at the denoiser} (\sdm). By step, the large pair's value / spelling ratio is 1.79 and 1.76 at the first two steps and 1.17--1.38 at later steps; the small pair's 2.3--3.4. Trajectories asked for four and for twenty give the same picture (1.30 and 1.38).}
\label{tab:supp_vdir}
\end{table}

\para{Results.} The large number's value moves the velocity only 1.34 times as far as its spelling, where a small number's value moves it 2.8 times as far. The velocity norms shared with the probe of \cref{sec:supp_mech} agree with it (median relative difference 0.7\%). The denoiser receives a large number mostly as a different word, not as a larger value.

\subsection{The timing test}
\label{sec:supp_timing}
\para{Design.} \sdm at $768^2$, 40 steps, guidance 4.5, U2's 64 types, seed 1. Every image shares prompt, noise and size with U2's plain image of its base request, which is its baseline. Pairs: 2 vs.\ 4 (control) and 12 vs.\ 25. Direction alternates over the sorted 16 people-counted types and over the sorted other 48, starting with ``up'' (32 types each): up, the base request is the lower count and the window's word the higher; down, the reverse. \emph{Swap}: the prompt asks the base request at every step except one window of five steps, where it asks the window's word; windows start at steps 0, 5, 10, 15, 20, 25 and 30 ($\sigma$ 1.000--0.963, 0.953--0.909, 0.897--0.843, 0.828--0.760, 0.741--0.653, 0.628--0.509, 0.475--0.308); 896 images. \emph{Amplified} (large pair): the prompt asks the window word's side throughout and receives the amplification contrast $+L\,(c(\text{asked})-c(\text{partner}))$, $L{=}5$, only inside a window starting at step 10, 15 or 20; baseline, the plain image of the asked request; 192 images. \emph{Floor}: the amplified arm's tasks for the 16 people-counted types at $L{=}10^{-6}$ (48 images). The sampler is the amplification test's (\cref{sec:supp_steer}). Per pair and window, over types ($s{=}{+}1$ up, $-1$ down): shift, the mean of $s\,[\log(1{+}n_W)-\log(1{+}n_0)]$; toward, among types whose count changed, the share that moved toward the window's word; changed, the share of types whose count changed. Counters and bootstrap as elsewhere. Images swapped in the first window differ from both the base request's and the window word's plain image (median pixel distance 26--46 and 16--20; none identical); images swapped in the last window are near their base (median about 1).

\para{Results} (\cref{tab:supp_timing}). The small word acts early: its shift intervals lie above zero at windows 0 and 5, with most changes toward the word, and its shift is near zero from window 15 on. At window 0 the word changes the count of 0.58 (small pair) and 0.64 (large pair) of the types, but no large-pair window has a shift interval above zero, at window 0 only 0.46 of the large word's changes go its way, and the amplified windows have no effect either. People's share toward the word is within 0.15 of \samt's on the same pictures at three of the six windows (the small pair at steps 0--4 and 10--14, the large pair at 0--4), but not at the small pair's steps 20--24 (one changed type) or the large pair's 10--14 and 20--24 (\cref{tab:supp_timing_people}). People's counts show one later shift toward the large word, at steps 20--24 ($+$0.20 \ci{0.01}{0.53}; four changed types, all toward the word), which no automatic counter shows (\samt on the same pictures: $+$0.02 \ci{$-$0.02}{0.08}); it rests on pictures that people count inconsistently (below). The large word thus changes many counts in the first window, without a direction, and does not act later under any automatic counter.

\begin{table*}[h]
\centering
\footnotesize
\setlength{\tabcolsep}{3pt}
\resizebox{\textwidth}{!}{%
\begin{tabular}{@{}llcccccccc@{}}
\toprule
& & \multicolumn{3}{c}{2 vs.\ 4, \samt} & \multicolumn{3}{c}{12 vs.\ 25, \samt} & 2 vs.\ 4 & 12 vs.\ 25 \\
\cmidrule(lr){3-5}\cmidrule(lr){6-8}\cmidrule(lr){9-9}\cmidrule(lr){10-10}
steps & $\sigma$ & shift & toward ($n$) & changed & shift & toward ($n$) & changed & shift: \gdino / \owl / cons. & shift: \gdino / \owl / cons. \\
\midrule
0--4 & 1.000 & $+$.15 \ci{.03}{.27} & .76 (37) & .58 & $-$.09 \ci{$-$.30}{.12} & .46 (41) & .64 & $+$.14 / $+$.10 / $+$.11 & $-$.17 / $-$.11 / $-$.30 \\
5--9 & .953 & $+$.16 \ci{.03}{.29} & .85 (26) & .41 & $+$.03 \ci{$-$.12}{.19} & .68 (31) & .48 & $+$.09 / $+$.06 / $+$.07 & $-$.09 / $-$.01 / $-$.02 \\
10--14 & .897 & $+$.00 \ci{$-$.09}{.08} & .50 (14) & .22 & $+$.08 \ci{$-$.03}{.23} & .64 (28) & .44 & $+$.03 / $+$.07 / $-$.01 & $-$.11 / $-$.03 / $+$.01 \\
15--19 & .828 & $-$.02 \ci{$-$.09}{.05} & .42 (12) & .19 & $+$.05 \ci{$-$.05}{.18} & .59 (22) & .34 & $+$.01 / $+$.03 / $+$.00 & $-$.04 / $-$.01 / $-$.01 \\
20--24 & .741 & $-$.02 \ci{$-$.05}{.01} & .25 (4) & .06 & $+$.04 \ci{$-$.04}{.17} & .40 (20) & .31 & $+$.00 / $-$.01 / $-$.01 & $-$.04 / $+$.01 / $+$.00 \\
25--29 & .628 & $+$.02 \ci{$-$.01}{.06} & .75 (4) & .06 & $+$.02 \ci{$-$.03}{.07} & .50 (14) & .22 & $+$.00 / $+$.00 / $+$.00 & $-$.03 / $+$.00 / $+$.00 \\
30--34 & .475 & $+$.00 \ci{$-$.02}{.01} & .33 (3) & .05 & $+$.02 \ci{$-$.02}{.08} & .43 (14) & .22 & $+$.00 / $-$.02 / $+$.00 & $-$.01 / $+$.00 / $+$.00 \\
\bottomrule
\end{tabular}}
\caption{\textbf{The timing test} (\sdm, U2, 64 types, seed 1). Shift with 95\% noun-cluster intervals; toward: share of the changed types that moved toward the window's word ($n$: changed types). Under \gdino the large pair's shift is negative at every window (window 10--14: $-$0.11 \ci{$-$0.19}{$-$0.03}) and its toward at most 0.42; under \owl the large pair's toward lies between 0.40 and 0.61; the consensus changes few large-pair counts after window 0--4. Amplified arm (\samt): window 10--14: shift $-$.02 \ci{$-$.08}{.05}, toward .50 (20); window 15--19: shift $+$.00 \ci{$-$.03}{.02}, toward .46 (13); window 20--24: shift $+$.02 \ci{$-$.01}{.05}, toward .58 (12). Floor: 87.5\% of counts equal the plain image's (\gdino 95.8\%, \owl 66.7\%, consensus 95.8\%). The consensus counts only images with a two-of-three majority, so its large-pair summaries rest on 23--36 of the 64 types (small pair 43--51). By direction: the large pair's small positive shifts at windows 5--19 come from ``down'' (base 25, window ``twelve''; $+$0.13, $+$0.15, $+$0.09); ``up'' gives $-$0.07, 0.00 and $+$0.01.}
\label{tab:supp_timing}
\end{table*}

\begin{table}[h]
\centering
\footnotesize
\setlength{\tabcolsep}{2.2pt}
\resizebox{\linewidth}{!}{%
\begin{tabular}{@{}llcccccc@{}}
\toprule
& & \multicolumn{3}{c}{people} & \multicolumn{2}{c}{\samt, same pictures} & toward \\
\cmidrule(lr){3-5}\cmidrule(lr){6-7}
pair & steps & shift & toward ($n$) & changed & shift & toward ($n$) & diff. \\
\midrule
2 vs.\ 4 & 0--4 & $+$.13 \ci{$-$.09}{.31} & .77 (13) & .81 & $+$.39 \ci{.13}{.68} & .85 (13) & $-$.08 \\
 & 10--14 & $+$.02 \ci{$-$.04}{.07} & .67 (3) & .19 & $+$.11 \ci{$-$.01}{.26} & .75 (4) & $-$.08 \\
 & 20--24 & $+$.03 \ci{.00}{.08} & 1.00 (1) & .06 & $-$.05 \ci{$-$.16}{.05} & .33 (3) & $+$.67 \\
12 vs.\ 25 & 0--4 & $-$.03 \ci{$-$.43}{.33} & .46 (13) & .81 & $-$.18 \ci{$-$.60}{.18} & .44 (9) & $+$.02 \\
 & 10--14 & $+$.20 \ci{$-$.06}{.58} & .60 (10) & .63 & $+$.03 \ci{$-$.04}{.11} & .80 (5) & $-$.20 \\
 & 20--24 & $+$.20 \ci{.01}{.53} & 1.00 (4) & .25 & $+$.02 \ci{$-$.02}{.08} & .50 (4) & $+$.50 \\
\bottomrule
\end{tabular}}
\caption{\textbf{People and the timing test} (16 types; people: the window picture's human count against the first study's human count of the same plain image; \samt: the same pictures against \samt on the plain image). The two first counts agree exactly on 60\% of the 96 pictures; checks .95, .95 and .90 exact per round.}
\label{tab:supp_timing_people}
\end{table}

\para{The large pair's later windows, picture by picture.} People's later changes on the large pair come mostly from pictures they count inconsistently: the first two counts agree on 5 of the 16 pictures at window 10--14 and 6 of 16 at window 20--24 (small pair 10--13 of 16 at every window; large pair at window 0--4, 12 of 16). The four changes at window 20--24: jewels (plain image 24 by people; the window picture's three counts 26, 1 and 1, so 1; \samt 1 on both), ironing boards (27; 0, 20 and 29, so 20; \samt 0 on both), candy bars (21; 18, 8 and 17, so 17; \samt 24 and 25) and spice racks (4; 5, 20 and 5, so 5; \samt 1 on both). The jewels' change from 24 to 1 is also one of the six changes toward the word at window 10--14 (counts 27, 1 and 1).

\subsection{The timing test in two more models}
\label{sec:supp_xtime}
\para{Design.} U2's 64 types, seed 1, the pairs and the direction per type of the timing test (\cref{sec:supp_timing}). \sdxl: the sampler of the word test (40 Euler steps, guidance 5, $1024^2$) with a prompt switch that replaces the text embeddings and the pooled embedding inside the window; plain images at 2, 4, 6, 8, 12, 16, 20, 25 and 30 (576; the baselines, and \sdxl's word test on U2); windows of five steps starting at steps 0, 5, \dots, 30 (896); the base number respelled in digits (``a photo of 12 cribs'' inside ``a photo of twelve cribs'') at windows 0 and 5 (256); the window's word from step 0 up to step 10 or 20 (256). \fluxk: the word test's sampler (50 steps, guidance 4, $1024^2$), every prompt encoded once by the pipeline and switched by its step callback; baselines are the word test's images; windows of six steps (12\% of sampling, as five of 40) starting at steps 0, 6, \dots, 36 for the large pair and 0, 6, 12 for the small one (640); respelled at windows 0 and 6 (large) and 0 (small; 192); the window's word up to step 12 or 24 (large; 128). \sdm: the base number respelled in digits inside the timing test's windows 0--4 and 5--9, both pairs (256), with its sampler. In every model the unconditional branch is unchanged. 3{,}200 images, counted by \samt, \gdino and \owl. A window \emph{acts} when its shift interval lies above 0 and toward is at least 0.75 (small pair) or 0.65 (large pair); \emph{full}: the window's word asked at every step (its plain image); \emph{prefix share}: prefix shift over full shift. In every model the first window's images differ from both plain images, while the last window's lie near the base request's plain image (\cref{tab:supp_xt_sdxl,tab:supp_xt_flux}).

\para{Results} (\samt). In \sdxl the large word moves the count over the whole trajectory ($+$0.30) and its first ten steps carry all of it (share 1.03); at its first window the small word changes more types than the respelling ($+$0.28 \ci{0.14}{0.42}), and all four \sdxl respelling intervals contain 0. The small word's first window shifts the count (0.17 \ci{0.02}{0.31}) with 74\% of the changes toward the word, and its windows at steps 5--9 and 10--14 act; no single large window acts. In \fluxk the large word moves the count over the whole trajectory ($+$0.45); the small word acts at steps 0--5 and 6--11, the large word's first window acts, and its first twelve steps carry the whole effect (share 1.00); in that first window the large word changes no more types than its respelling ($+$0.03 \ci{$-$0.08}{0.14}), although only the word's changes have a direction (82\% against 58\% toward). In \sdm the large word's first window changes about as many types as its respelling ($+$0.05), while the small word's changes 0.33 more. Late windows stay small: every window from step 15 (\sdxl) or 18 (\fluxk) has its interval within $\pm$0.10 except one in each model (\sdxl steps 20--24: $[-0.004, 0.18]$; \fluxk steps 18--23: $[-0.07, 0.10]$, upper end 0.101). \sdxl's order on U2 is 0.76 at 2--8 and 0.61 at 12--30 (\cref{sec:supp_wordcount}).

\para{What it shows.} In \fluxk the earliest acting large window is the first, no later one acts and the prefix carries the whole effect: where its large number word moves the count, it acts early. In \sdxl no single window passes the direction criterion, but the first ten steps act and carry about all of the effect (share 1.03 \ci{0.53}{2.33}): an effect spread over the early windows, below single-window detection; under \owl the first window acts (0.30 \ci{0.15}{0.46}, 69\% toward). No window from step 10 acts in either model, so large numbers do not act late. In \sdm the large word's early change equals the respelling's and has no direction.

\para{Counters.} For \sdxl, \owl gives the clearest large-pair effect (whole trajectory 0.45 \ci{0.25}{0.65}, prefix share 0.86) and \gdino the weakest (0.15 \ci{$-$0.07}{0.40}; its steps 5--9 window acts, and its respelling at steps 0--4 also leans toward the word, 0.25 \ci{0.06}{0.43}). For \fluxk the detectors disagree with \samt on direction: \gdino reads fewer objects for twenty-five over the whole trajectory ($-$0.38 \ci{$-$0.61}{$-$0.16}, set within the first twelve steps, share 1.01) and \owl no change ($-$0.04 \ci{$-$0.30}{0.22}); both miss many of \fluxk's objects, which shrink to about 2\% of the image at 25--30 (\cref{tab:supp_objsize}), while people's counts in the word test grow from 12 for twelve to 20 for twenty-five. In both models and under every counter, every window from step 15 or 18 on stays within $-$0.07 to 0.18.

\para{Up and down.} At the first window, the large word moves the count toward itself in both directions by the point estimates: in \sdxl by $+$0.16 for the 32 types asked to go up and $+$0.29 for the 32 asked to go down, in \fluxk by $+$0.16 and $+$0.33; each direction's interval, on 32 types, contains 0.

\para{Noise levels.} The three schedules reach a given noise level at different steps. From each sampler's own settings, the noise level $\sigma$ at the first step of the five-step windows 0, 5, 10, \dots, 30 is 1.00, 0.95, 0.90, 0.83, 0.74, 0.63 and 0.48 in \sdm (FlowMatch, shift 3) and, as the rectified-flow equivalent $\sigma_{\text{Euler}}/(1+\sigma_{\text{Euler}})$, 0.93, 0.87, 0.79, 0.70, 0.60, 0.50 and 0.39 in \sdxl (scaled-linear betas, leading spacing; its first step is not pure noise); at the six-step windows 0, 6, 12, \dots, 36 of \fluxk (dynamic shift for $1024^2$) it is 1.00, 0.98, 0.96, 0.93, 0.89, 0.83 and 0.75. The small word's acting windows therefore end at $\sigma$ 0.91 in \sdm, 0.96 in \fluxk and 0.72 in \sdxl; the numbers are not strictly comparable across models, whose latent spaces differ.

\para{Object sizes in the windows.} Against the base request's plain image, the change in log median \samt box area, signed so that positive means the objects changed size the way the window's word asks (smaller for the larger request): \fluxk, large pair, $+$0.66 \ci{0.41}{0.92} with the word at every step, $+$0.35 \ci{0.15}{0.55} at steps 0--5, $+$0.33 \ci{0.18}{0.48} at 6--11, $+$0.09 at 12--17, at most $+$0.04 later, $+$0.69 for the twelve-step prefix; \sdxl, large pair, $+$0.35 \ci{0.01}{0.69} at every step, $+$0.21 \ci{$-$0.09}{0.52} at steps 0--4, at most $+$0.03 later, $+$0.35 for the ten-step prefix; \sdm, large pair, $+$0.01 \ci{$-$0.32}{0.35} at every step. The small pairs rescale by $+$0.49 to $+$0.71 at every step and by $+$0.28 to $+$0.36 in the first window; respelling rescales by at most $+$0.17 (intervals containing 0 except \sdxl's small pair at steps 5--9, $+$0.15 \ci{0.02}{0.32}).

\begin{table*}[h]
\centering
\footnotesize
\resizebox{\textwidth}{!}{%
\begin{tabular}{@{}lccccl@{}}
\toprule
& \multicolumn{3}{c}{\samt} & & \\
\cmidrule(lr){2-4}
window (steps) & shift & toward (types) & changed & shift: \gdino / \owl / consensus & \\
\midrule
\multicolumn{6}{@{}l}{\emph{\sdxl, two vs.\ four}} \\
2 vs.\ 4: every step & $+$.41 \ci{.27}{.55} & .85 (54) & .84 & $+$.38 / $+$.43 / $+$.35 &  \\
\quad swap 0--4 & $+$.17 \ci{.02}{.31} & .74 (39) & .61 & $+$.19 / $+$.19 / $+$.23 &  \\
\quad swap 5--9 & $+$.16 \ci{.06}{.27} & .77 (31) & .48 & $+$.09 / $+$.14 / $+$.18 & acts \\
\quad swap 10--14 & $+$.09 \ci{.00}{.17} & .76 (17) & .27 & $+$.02 / $+$.03 / $+$.06 & acts \\
\quad swap 15--19 & $+$.01 \ci{$-$.02}{.03} & .54 (13) & .20 & $+$.00 / $-$.01 / $+$.00 & $\pm$.10 \\
\quad swap 20--24 & $+$.01 \ci{$-$.01}{.03} & .57 (7) & .11 & $+$.01 / $+$.00 / $+$.02 & $\pm$.10 \\
\quad swap 25--29 & $+$.00 \ci{$-$.03}{.01} & .60 (5) & .08 & $+$.00 / $+$.00 / $+$.01 & $\pm$.10 \\
\quad swap 30--34 & $+$.00 \ci{$-$.02}{.01} & .57 (7) & .11 & $-$.01 / $-$.03 / $+$.01 & $\pm$.10 \\
\quad respelled 0--4 & $+$.03 \ci{$-$.05}{.10} & .57 (21) & .33 & $+$.04 / $+$.04 / $+$.02 &  \\
\quad respelled 5--9 & $+$.01 \ci{$-$.06}{.08} & .56 (16) & .25 & $-$.02 / $+$.01 / $-$.03 &  \\
\quad prefix 0--9 & $+$.29 \ci{.15}{.44} & .85 (47) & .73 & $+$.34 / $+$.32 / $+$.37 & acts (share .72 \ci{.45}{.94}) \\
\quad prefix 0--19 & $+$.40 \ci{.26}{.53} & .91 (53) & .83 & $+$.41 / $+$.41 / $+$.40 & acts (share .98 \ci{.90}{1.07}) \\
\midrule
\multicolumn{6}{@{}l}{\emph{\sdxl, twelve vs.\ twenty-five}} \\
12 vs.\ 25: every step & $+$.30 \ci{.04}{.56} & .60 (50) & .78 & $+$.15 / $+$.45 / $+$.59 &  \\
\quad swap 0--4 & $+$.22 \ci{.02}{.43} & .62 (45) & .70 & $+$.14 / $+$.30 / $+$.20 &  \\
\quad swap 5--9 & $+$.02 \ci{$-$.10}{.12} & .54 (39) & .61 & $+$.14 / $+$.10 / $+$.00 &  \\
\quad swap 10--14 & $-$.04 \ci{$-$.10}{.00} & .44 (32) & .50 & $+$.04 / $+$.00 / $+$.01 &  \\
\quad swap 15--19 & $+$.01 \ci{$-$.03}{.05} & .52 (29) & .45 & $-$.01 / $-$.02 / $+$.00 & $\pm$.10 \\
\quad swap 20--24 & $+$.06 \ci{.00}{.18} & .56 (25) & .39 & $+$.02 / $+$.00 / $+$.01 &  \\
\quad swap 25--29 & $+$.00 \ci{$-$.02}{.02} & .50 (18) & .28 & $+$.01 / $+$.01 / $+$.00 & $\pm$.10 \\
\quad swap 30--34 & $+$.01 \ci{$-$.01}{.04} & .63 (19) & .30 & $+$.01 / $+$.00 / $+$.00 & $\pm$.10 \\
\quad respelled 0--4 & $-$.03 \ci{$-$.32}{.26} & .53 (45) & .70 & $+$.25 / $+$.08 / $+$.07 &  \\
\quad respelled 5--9 & $+$.01 \ci{$-$.14}{.16} & .55 (33) & .52 & $+$.07 / $+$.02 / $+$.02 &  \\
\quad prefix 0--9 & $+$.30 \ci{.08}{.52} & .70 (47) & .73 & $+$.17 / $+$.38 / $+$.09 & acts (share 1.03 \ci{.53}{2.33}) \\
\quad prefix 0--19 & $+$.32 \ci{.04}{.61} & .62 (50) & .78 & $+$.16 / $+$.44 / $+$.46 &  (share 1.07 \ci{.61}{1.85}) \\
\bottomrule
\end{tabular}}
\caption{\textbf{The timing test in \sdxl} (U2, 64 types, seed 1; 95\% intervals over types). Shift: mean change in $\log(1{+}n)$ toward the window's word against the plain image of the base request; toward: share of the types whose count changed that moved toward the word (number of such types); changed: share of types whose count changed. ``acts'': shift interval above 0 and toward at least 0.75 (small pair) or 0.65 (large pair); $\pm$.10: interval within $\pm$0.10 (windows from step 15); share: prefix shift over the shift with the word at every step.}
\label{tab:supp_xt_sdxl}
\end{table*}

\begin{table*}[h]
\centering
\footnotesize
\resizebox{\textwidth}{!}{%
\begin{tabular}{@{}lccccl@{}}
\toprule
& \multicolumn{3}{c}{\samt} & & \\
\cmidrule(lr){2-4}
window (steps) & shift & toward (types) & changed & shift: \gdino / \owl / consensus & \\
\midrule
\multicolumn{6}{@{}l}{\emph{\fluxk, two vs.\ four}} \\
2 vs.\ 4: every step & $+$.26 \ci{.10}{.41} & .82 (50) & .78 & $+$.43 / $+$.27 / $+$.25 &  \\
\quad swap 0--5 & $+$.17 \ci{.03}{.31} & .83 (30) & .47 & $+$.21 / $+$.21 / $+$.17 & acts \\
\quad swap 6--11 & $+$.16 \ci{.10}{.23} & .96 (24) & .38 & $+$.16 / $+$.11 / $+$.12 & acts \\
\quad swap 12--17 & $-$.01 \ci{$-$.07}{.04} & .63 (8) & .13 & $+$.03 / $-$.01 / $+$.01 &  \\
\quad respelled 0--5 & $-$.01 \ci{$-$.12}{.10} & .55 (11) & .17 & $+$.04 / $+$.01 / $+$.04 &  \\
\midrule
\multicolumn{6}{@{}l}{\emph{\fluxk, twelve vs.\ twenty-five}} \\
12 vs.\ 25: every step & $+$.45 \ci{.20}{.69} & .83 (42) & .66 & $-$.38 / $-$.04 / $-$.16 &  \\
\quad swap 0--5 & $+$.24 \ci{.01}{.48} & .82 (38) & .59 & $-$.17 / $-$.11 / $+$.16 & acts \\
\quad swap 6--11 & $+$.11 \ci{$-$.06}{.29} & .84 (31) & .48 & $-$.17 / $-$.05 / $-$.13 &  \\
\quad swap 12--17 & $+$.06 \ci{$-$.03}{.18} & .86 (14) & .22 & $-$.02 / $-$.05 / $-$.02 &  \\
\quad swap 18--23 & $+$.02 \ci{$-$.07}{.10} & .64 (14) & .22 & $+$.01 / $+$.00 / $+$.00 &  \\
\quad swap 24--29 & $-$.02 \ci{$-$.07}{.01} & .50 (10) & .16 & $+$.00 / $+$.00 / $+$.00 & $\pm$.10 \\
\quad swap 30--35 & $-$.03 \ci{$-$.07}{.00} & .25 (8) & .13 & $+$.01 / $+$.00 / $+$.00 & $\pm$.10 \\
\quad swap 36--41 & $+$.00 \ci{$-$.05}{.04} & .56 (9) & .14 & $+$.01 / $+$.00 / $+$.00 & $\pm$.10 \\
\quad respelled 0--5 & $+$.05 \ci{$-$.17}{.29} & .58 (36) & .56 & $-$.05 / $+$.04 / $-$.08 &  \\
\quad respelled 6--11 & $+$.18 \ci{$-$.06}{.44} & .63 (27) & .42 & $+$.05 / $+$.04 / $-$.02 &  \\
\quad prefix 0--11 & $+$.45 \ci{.19}{.71} & .83 (41) & .64 & $-$.38 / $-$.04 / $-$.16 & acts (share 1.00 \ci{.83}{1.18}) \\
\quad prefix 0--23 & $+$.47 \ci{.22}{.73} & .83 (42) & .66 & $-$.38 / $-$.05 / $-$.09 & acts (share 1.06 \ci{.99}{1.18}) \\
\midrule
\multicolumn{6}{@{}l}{\emph{\sdm, the base number respelled inside the timing test's windows}} \\
SD3.5-M small respelled 0--4 & $+$.05 \ci{$-$.05}{.16} & .50 (16) & .25 & $+$.04 / $-$.08 / $+$.05 &  \\
SD3.5-M small respelled 5--9 & $-$.03 \ci{$-$.09}{.02} & .38 (13) & .20 & $+$.02 / $-$.06 / $+$.01 &  \\
SD3.5-M large respelled 0--4 & $-$.10 \ci{$-$.27}{.08} & .34 (38) & .59 & $-$.07 / $-$.02 / $-$.10 &  \\
SD3.5-M large respelled 5--9 & $+$.06 \ci{$-$.07}{.22} & .54 (26) & .41 & $-$.01 / $+$.00 / $-$.06 &  \\
\bottomrule
\end{tabular}}
\caption{\textbf{The timing test in \fluxk, and the respelling control in \sdm} (as \cref{tab:supp_xt_sdxl}; \fluxk windows of six of 50 steps, $\pm$.10 from step 18). \sdm's value swaps in the same windows are in \cref{sec:supp_timing}.}
\label{tab:supp_xt_flux}
\end{table*}

\subsection{When the model's own crowd is fixed}
\label{sec:supp_commit}
A commitment probe follows \sdm's own plain trajectories asked for 4, 12 and 25 objects (64 U2 types, seed 1; every final image equals U2's, mean absolute difference 0). At each step the guided velocity gives an estimate $\hat x_0$ of the clean latent; we correlate it, after an ideal low-pass at 2, 4 or 8 cycles per image width, with the final latent, and \samt counts the decoded $\hat x_0$ at twelve steps. The coarse layout is fixed at once: the correlation at 4 cycles reaches 0.9 by step 2 ($\sigma{=}0.98$) in the median trajectory at every request (mean 0.84--0.86 at step 0, 0.99 at step 9), so its share fixed by step 9 is 1.00 at 4, 12 and 25. The count, as \samt reads the estimated image, settles later: within $\max(1, 10\%)$ of the final count from step 9 on ($\sigma{=}0.91$) in 47\% \ci{35}{60} of trajectories asked for four, 40\% \ci{27}{53} for twelve and 28\% \ci{16}{41} for twenty-five (25 minus 4: $-$0.19 \ci{$-$0.36}{$-$0.02}), with median settling steps 12, 12 and 15 ($\sigma$ 0.87--0.83); blurred early estimates may be read late, so these bound the count's commitment from above. The final counts stay small whatever the request (medians 4, 8 and 6). By step 9 the layout is set at every request, while the count is set in fewer than half of the trajectories at any request.

\subsection{Amplifying the number word}
\label{sec:supp_steer}
\sdm, U2, seed 1, \samt unless stated. During the first ten steps ($\sigma\geq0.909$), $L\,(v(c)-v(c'))$ was added to the guided velocity, $c'$ being the prompt with the pair's other request, a contrastive guidance~\cite{wu2024contrastive} used to improve counts up to ten~\cite{sepehri2026athena} ($L{=}2$, 5, 10; 768 images), or the prompt without its number ($L{=}5$, 10; 896 images). The small pair stays ordered (0.86--0.92; 0.94 unsteered); the large pair does not become ordered (0.52 unsteered, 0.41 \ci{0.28}{0.54} at $L{=}10$). Against the number-free prompt the exact rate at 12--30 moves from 2.5\% to 4.1\% and 3.1\% ($L{=}5$ minus $L{=}0$: $+$0.016 \ci{$-$0.006}{0.041}), and people find 4\% exact at $L{=}10$, the plain prompt's rate; their medians at 12, 16, 20, 25 and 30 are 8, 6, 8.5, 10.5 and 4.5 (\samt: 6.5, 2, 6, 4.5 and 5). In the timing test's amplified arm (\cref{sec:supp_timing}) the large word was amplified inside windows from step 10 to step 24 without effect. Amplified, the large number word gains no direction.

\subsection{A direction borrowed from small numbers}
\label{sec:supp_borrow}
\para{Design.} \sdm, 64 types, seed 1, the plain images' noise ($768^2$, 40 steps, guidance 4.5). The prompt asks for twelve (the main case) or twenty-five; during the first ten steps ($\sigma\geq0.909$) the guided velocity gets $+L\,(v(\text{four})-v(\text{two}))$ (``up'') or the reverse (``down''), where $v(c)$ is the conditional velocity for ``a photo of $c$ $X$s'' at the same state, with $L=1,2,4,8$, and $L=0$ is the plain image (1{,}152 images); every cell differs from every other cell of its noise. The \emph{L-order} is the share of same-noise pairs of strengths whose counts differ in which the stronger push moves the count the borrowed way (more objects up, fewer down).

\begin{table}[!htbp]
\centering
\footnotesize
\setlength{\tabcolsep}{2.4pt}
\resizebox{\linewidth}{!}{%
\begin{tabular}{@{}llccc@{}}
\toprule
request, push & counter & L-order & shift at $L{=}8$ & medians at $L=0,1,2,4,8$ \\
\midrule
twelve, up & \samt & .36 \ci{.27}{.46} & $-$0.23 \ci{$-$0.44}{$-$0.03} & 8, 8, 8, 6, 4 \\
 & \gdino, \owl, cons. & .50, .35, .32 & $+$0.09, $-$0.21, $-$0.12 & \\
twelve, down & \samt & .73 \ci{.66}{.79} & $-$0.59 \ci{$-$0.78}{$-$0.40} & 8, 9, 8, 5, 2 \\
 & \gdino, \owl, cons. & .66, .66, .77 & $-$0.36, $-$0.59, $-$0.64 & \\
twenty-five, up & \samt & .43 \ci{.35}{.52} & $-$0.10 \ci{$-$0.34}{0.14} & 6, 7, 6, 6, 4 \\
twenty-five, down & \samt & .68 \ci{.60}{.75} & $-$0.60 \ci{$-$0.84}{$-$0.35} & 6, 7, 3, 2, 2 \\
\bottomrule
\end{tabular}}
\caption{\textbf{The small pair's difference added to a crowd prompt} (\sdm, 64 types, seed 1). Shift: mean $\log(1{+}n_L)-\log(1{+}n_0)$; 95\% intervals over types. Images in which \samt finds no object: 17\% at $L{=}0$ and 16--20\% at $L{=}8$.}
\label{tab:supp_borrow}
\end{table}

\para{Results.} The up push moves the crowd the other way: its L-order is a reversal, with its interval below 0.5, and at $L{=}8$ it shrinks the crowd. The down push shrinks it, the way it pushes (0.73). Both pushes bring the crowd toward the small pair's own counts (medians about four and two at $L{=}8$), and the exact rate at twelve falls from 9\% to 5\%. The small numbers' difference points toward small counts, not toward ``more''.

\subsection{Text streams and dropped encoders}
\label{sec:supp_stream}
\para{Design.} From the noise of each type's plain image (seeds 1 and 2), each text stream is given one number of a pair, two against four or twelve against twenty-five, in the prompt ``a photo of $c$ $X$s''. Cells are named by the number given to CLIP and to T5: LL and HH are the plain images of the smaller and the larger request; HL gives CLIP the larger and T5 the smaller, LH the reverse. In \sdm, CLIP means both CLIP token sequences and the pooled CLIP vector; two more cells change the pooled vector alone. In \flux, CLIP is its only CLIP input, a pooled vector. The contrast for a stream compares two cells that differ only in that stream's number (CLIP: HL against LL and HH against LH; T5: LH against LL and HH against HL; both: HH against LL); its order is the share of contrasts whose counts differ in which the cell with that stream's larger number has more objects. \sdm: 64 types, 1{,}536 images; \flux: 32 types (the 16 counted by people and 16 drawn at random from the other 48), 512 images, at the settings of its word test.

\begin{table}[!htbp]
\centering
\footnotesize
\setlength{\tabcolsep}{2.2pt}
\resizebox{\linewidth}{!}{%
\begin{tabular}{@{}llcccccc@{}}
\toprule
& & \multicolumn{3}{c}{two vs four} & \multicolumn{3}{c}{twelve vs twenty-five} \\
\cmidrule(lr){3-5}\cmidrule(lr){6-8}
& counter & CLIP & T5 & both & CLIP & T5 & both \\
\midrule
\sdm & \samt & .85 \ci{.79}{.92} & .85 \ci{.78}{.91} & .91 \ci{.84}{.97} & .47 \ci{.38}{.56} & .55 \ci{.47}{.64} & .53 \ci{.41}{.65} \\
 & \gdino & .88 & .80 & .88 & .30 \ci{.22}{.38} & .42 & .31 \\
 & \owl & .80 & .74 & .89 & .45 & .55 & .46 \\
 & consensus & .91 & .90 & .94 & .26 \ci{.13}{.42} & .61 & .34 \\
\flux & \samt & .63 \ci{.38}{.83} & .84 \ci{.73}{.93} & .83 \ci{.69}{.94} & .64 \ci{.48}{.78} & .36 \ci{.15}{.56} & .53 \ci{.33}{.71} \\
 & \gdino & .56 & .85 & .88 & .49 & .28 \ci{.16}{.40} & .25 \ci{.11}{.41} \\
 & \owl & .51 & .81 & .77 & .57 & .39 & .51 \\
 & consensus & .57 & .89 & .88 & .43 & .28 \ci{.07}{.52} & .35 \\
\bottomrule
\end{tabular}}
\caption{\textbf{Stream test, every counter} (order of each stream's contrast; 95\% intervals over types). \sdm's pooled vector alone: 0.56 \ci{0.44}{0.66} at two vs four and 0.54 \ci{0.45}{0.63} at twelve vs twenty-five (\samt). Interaction (both minus the sum of the two streams' effects, in $\log(1{+}n)$): \sdm $-$0.02 \ci{$-$0.11}{0.08} and $+$0.08 \ci{$-$0.10}{0.25}; \flux $+$0.04 \ci{$-$0.06}{0.15} and $-$0.08 \ci{$-$0.27}{0.08}.}
\label{tab:supp_stream}
\end{table}

\para{Results.} In both models at least one stream carries two against four to the crowd in order, and no stream carries twelve against twenty-five (\cref{tab:supp_stream}). \sdm's CLIP and T5 streams each order the small pair and neither the large one; under \gdino and the consensus its CLIP stream reverses the large pair. \flux's T5 stream orders the small pair and moves the large one the wrong way if anything, a reversal under every counter.

\para{One stream removed.} \sdm was trained with each encoder dropped at random, so it can draw with one of them empty (T5 as 77 zero tokens; CLIP as zero tokens and a zero pooled vector). With T5 dropped (CLIP only), the order at 2--8 stays at 0.88 \ci{0.83}{0.93} and at 12--30 is 0.41 \ci{0.34}{0.49} (plain images: 0.89 and 0.46; 64 types, seed 1, \samt); with CLIP dropped (T5 only), the order is 0.75 \ci{0.67}{0.83} at 2--8 and 0.36 \ci{0.28}{0.44} at 12--30, and the crowds past ten nearly vanish (medians 2, 1 and 1 objects at 12, 20 and 30, against 8, 6 and 4.5). Neither arm extends the reach: relative to the plain images, the order at 12--30 changes by $-$0.05 \ci{$-$0.13}{0.04} and $-$0.10 \ci{$-$0.20}{0.01}, and both orders past ten are reversals. The large number is not lost in one encoder: in \sdm each stream carries a small number to the crowd and none a large one, and in \flux the T5 stream carries a small number and not a large one.

\subsection{Starts against a planted count}
\label{sec:supp_planted}
\sdm, U2, seed 1, \samt unless stated.

\para{Earlier starts against a plant of the other count.}\label{sec:supp_sweep}
A grid of 2, 4, 12 or 25 cut-outs (58 types) was planted at $\sigma$ from 0.991 to 0.909 and asked for the other count of its pair (116 conflicting and 116 agreeing cells per level). The small word pulls the count toward itself at every level (pull in $\log(1{+}n)$ from $+$0.40 at 0.991 to $+$0.18 at 0.909; all intervals exclude zero) and raises the share drawn with the asked count by 0.53 to 0.20 over the agreeing cells; the large word raises it by $-$0.01 to $+$0.03, and its pull lies between $-$0.20 and $+$0.14 with intervals that include zero except at 0.982 ($-$0.40 to $-$0.02). Under \gdino, \owl and their consensus the small-pair lift lies between 0.13 and 0.53 and the large-pair lift between $-$0.01 and $+$0.03. A planted 2 or 4 leaves free canvas while a planted 12 or 25 fills it; the re-noise test below removes this difference.

\para{The model's own images, re-noised.}\label{sec:supp_renoise}
U2's plain images asked for six and twenty objects (64 types; median \samt count 5 and 6) were re-noised with their own noise to $\sigma$ 0.991--0.813 and finished asking for 2 or 4, or for 12 or 25 (2{,}560 renders). The small pair stays ordered at 0.991 (order 0.93, pull $+$0.41) and loses order as the start sharpens (0.59 at 0.813); the large pair's order is 0.45--0.56 down to $\sigma{=}0.872$ (pulls $-$0.08 to 0.00) and 0.80 at 0.813 (pull $+$0.07 \ci{0.02}{0.16}), where most starts keep their count (0.69). Under \gdino, \owl and the consensus the large-pair pull never exceeds $+$0.02.

\para{Starts between the two requests.}\label{sec:supp_straddle}
Here the planted crowd lies between the pair's two requests, so that either word must add or remove objects. For the large pair it is the model's own image of the type with 16--20 objects by \samt, closest to 18 (51 types: 49 self-planted images, \cref{sec:supp_method}, and 2 plain ones; mean \samt count 17.9); for the small pair, a plain image with exactly three objects (32 types); seed 1, or seed 2 for a type without such an image. Each start is re-noised with its own noise (\cref{eq:start}) to $\sigma$ 0.991, 0.963, 0.909, 0.872 or 0.813 and finished asking for 12 or 25, or for 2 or 4: 51 types $\times$ 2 requests $\times$ 5 levels and 32 $\times$ 2 $\times$ 5, 830 renders. At 0.991 the small pair is ordered (0.96) and the large pair is not (0.51 \ci{0.36}{0.66}; small minus large 0.45 \ci{0.28}{0.61}): asking for twelve thins the crowd in 98\% of renders, and asking for twenty-five adds objects in only 18\% (mean counts 8.3 and 10.6 from about 18). From the later starts the large pair's order reaches 0.66 \ci{0.51}{0.80} at $\sigma{=}0.963$ and 0.80 \ci{0.60}{0.96} at 0.909, over the types whose two counts differ (41 and 20 of the 51), but the two words' mean counts differ there by less than one object: 11.9 for twelve against 12.6 for twenty-five at 0.963, and 16.8 against 17.1 at 0.909 (differences 0.65 and 0.35). From 0.909 on, both words mostly keep the start's count (0.55 of renders at 0.909, 0.74 at 0.813).

From a planted crowd, a small word moves the count toward itself at every start level; a large word hardly moves it toward its own count.

\subsection{Training the number word}
\label{sec:supp_ceiltrain}
\para{Design.} Two LoRAs of \sdm (rank 32, $\alpha{=}64$, the attention projections of every joint block; learning rate $10^{-4}$, batch 2, 1{,}500 steps, about 1.5 passes; seed 2027) were trained with the model's own flow-matching loss on 2{,}037 of its own images of U1 and U3, none of a U2 type, whose \samt count equals the request: 1{,}653 self-planted crowds of 12--30 objects (99 types; every arrangement of \cref{sec:supp_method}; $768^2$) and 384 plain images of two to four objects (111 types; $512^2$). The \emph{count} LoRA sees each image with its own prompt; the \emph{shuffled} LoRA sees the same images in the same order with the same noise, but each crowd with the number word of another crowd (a seeded permutation; 162 of 1{,}653 keep their own). Each was tested at its training scale from U2's plain prompts and noise (64 types, nine requests, seed 1, $768^2$, 40 steps, guidance 4.5; 576 images per arm). Twelve is the most common crowd count among the training images (306 of 1{,}653; 216 at twenty, 172 at thirty).

\para{Results} (\samt, 64 types). Both arms learned crowds (count $\times$1.52 \ci{1.29}{1.78} at 12--30, shuffled $\times$2.40). The count arm's order at 12--30 is 0.58 \ci{0.49}{0.67}, against 0.44 \ci{0.36}{0.52} shuffled and 0.46 plain: by \samt the count LoRA orders large crowds somewhat better than the shuffled one ($+$0.14 \ci{0.01}{0.27}). Its order at 2--8 is 0.92 (shuffled 0.91, plain 0.89). Exact by request (count arm): 0.86, 0.70, 0.53, 0.36, 0.41, 0.02, 0.12, 0.02 and 0.02 at 2--30, against 0.72, 0.44, 0.16, 0.12, 0.09, 0.03, 0, 0 and 0 plain and 0.70, 0.11, 0, 0, 0, 0, 0.02, 0.05 and 0.08 shuffled; at 12--30, 11.6\% against 2.5\% plain and 2.8\% shuffled. Medians at 12, 16, 20, 25 and 30: 12, 12, 14, 12, 10 (count), 28, 28, 28, 26, 24 (shuffled), 8, 8, 6, 6, 4 (plain). Ceiling (cap model): 5.4 \ci{4.5}{7.5} plain, 9.3 \ci{7.4}{10.8} count, 9.2 \ci{7.1}{11.6} shuffled (count over shuffled 1.01 \ci{0.83}{1.24}). Under \gdino, \owl and the consensus, count minus shuffled order at 12--30 is $+$0.08 \ci{$-$0.04}{0.20}, $+$0.10 \ci{0.00}{0.19} and $+$0.22 \ci{0.03}{0.41}, the count arm's order 0.52, 0.53 and 0.61, and its exact rate at 12--30 8.7--11.6\%.

\para{People} (fifth count). People counted 16 U2 types of both arms at every request (288 images; \cref{sec:supp_hcstudies}). The count arm's order at 2--8 is 0.98 and its mean $\log(1{+}n)$ at 12--30 exceeds the untrained model's by 0.46 \ci{0.13}{0.80}, but its order at 12--30, 0.43 \ci{0.26}{0.61}, is no higher than the shuffled arm's (count minus shuffled 0.08 \ci{$-$0.12}{0.27}): by people the count LoRA does not beat the shuffled one. Exact by request (count arm): 1.00, 0.81, 0.38, 0.44, 0.38, 0.06, 0.06, 0 and 0.06 at 2--30 (untrained 0.94, 0.38, 0, 0.13, 0.13, 0.06, 0, 0, 0; shuffled 0.81, 0.13 and then 0); at 12--30, 11\% against 4\% untrained and 0\% shuffled. Medians at 12--30: 12, 12, 12, 12, 9.5 (count), 36, 35, 33, 34.5, 17.5 (shuffled); on the same pictures \samt's medians of the count arm at 16--30 are 12, 12, 12 and 7.5.

Trained on its own counted crowds, \sdm learns counts up to about twelve, the most common crowd count among the training images, but not an order past it; trained on the same crowds with shuffled counts, it draws crowds of about 35 objects, so drawing capacity is not the limit.

\subsection{The conditional read as a classifier}
\label{sec:supp_dc}\label{tab:supp_dc}
\para{Method.} A diffusion classifier~\cite{li2023diffclassifier,clark2023zeroshot}, as Zhao~\etal use to probe numerosity~\cite{zhao2025noise}, compares, on an image $y$ of known count, the conditional denoising loss under two prompts. The image is encoded to $x_0$ (latent mode), noised to $x_k=(1-\sigma_k)x_0+\sigma_k\epsilon$ at ten steps $k$ of the model's own schedule (six draws of $\epsilon$ per image and step, shared by the prompts), and the conditional branch alone gives the velocity. An image \emph{prefers} the true count when the loss under the other count's prompt, averaged over the draws, is higher; the next type's noun at the true count checks that the classifier reads the image at all. Win rate: share of images that prefer the true prompt, with 95\% intervals over object types. The \emph{separation} of a pair is the probability that an image with the larger count favours the larger word more than an image with the smaller count does (an AUC of the loss difference; 0.5: no separation), which, unlike the win rate, ignores a shift of both counts toward one word. Early steps ($\sigma\geq0.91$ in \sdm) are those of the small word's acting windows in the timing test. A version on composites of known count (cut-outs pasted on a flat ground) did not pass its own validity check, as the conditional named the object there less reliably than on the model's own images, so we do not use it.

\para{The model's own images.} The probe uses 259 of \sdm's own U2 images whose \samt count equals the request: plain images asked for two (84 images of 52 types) or four (50 of 37), self-planted images asked for twelve (59 of 36) or twenty-five (56 of 37), and plain images asked for twelve (10). Here the classifier reads the object name: 0.97 at $\sigma{=}0.94$ and 1.00 from 0.91 on. Two against four: 0.83, 0.88, 0.93 and 0.90 at $\sigma$ 0.98, 0.96, 0.94 and 0.91 (two objects 0.81--0.92, four 0.86--0.96); at $\sigma{=}1$ both small counts lean toward ``four'' (two objects 0.49, four 0.86). Twelve against twenty-five: images of twelve prefer ``twelve'' throughout the high-noise steps (0.81--0.88), but images of twenty-five prefer ``twenty-five'' in only 0.14 \ci{0.05}{0.25} and 0.32 \ci{0.18}{0.46} at $\sigma$ 1 and 0.98, in 0.54 \ci{0.39}{0.68} and 0.48 \ci{0.33}{0.64} at 0.96 and 0.94, and in 0.71 \ci{0.61}{0.82} at 0.91, 0.64--0.89 below. Pooled over both directions, the large pair reaches 0.78 \ci{0.72}{0.85} at $\sigma{=}0.91$, the last step of the word's acting windows; above it, the pooled rate is at most 0.69, the later steps prefer the true large count more often than the early ones (difference 0.10 \ci{0.04}{0.15}), and at the noisiest steps both large counts lean toward ``twelve''. Separation: two from four 0.90, 0.98, 0.99 and 0.98 at $\sigma$ 0.98--0.91; twelve from twenty-five 0.68 \ci{0.58}{0.77}, 0.76 \ci{0.67}{0.83}, 0.76 \ci{0.68}{0.84} and 0.83 \ci{0.76}{0.90}. The ten plain images of twelve objects prefer ``twelve'' in 0.80--1.00 at high noise. At the noise levels where \sdm's layout forms, its conditional thus carries the large count, but separates it far less than the small one and leans toward the smaller word.

\para{An observer baseline.} A small convolutional network trained on noised latents of the same images (fresh noise at each level; five folds over object types, the large pair's training folds adding 356 self-planted grids of U1 and U3) and tested on the classifier's own six noise draws per image and level gives a reference for what the noisy latent itself carries. Ranking images by the loss difference, \sdm's conditional separates two objects from four almost fully from $\sigma{=}0.98$ on (area under the curve 0.90--0.99) and twelve from twenty-five partly (0.68--0.83); with the counts written in digits it separates the large pair less (0.55--0.63; digits minus words $-$0.14 \ci{$-$0.19}{$-$0.09}). The observer separates the twelve- and twenty-five-object grids almost fully from $\sigma{=}0.96$ on (0.97--1.00; 0.87 at 0.98) but the plain two- and four-object images poorly (0.65--0.68): averaged over $\sigma$ 0.98--0.94, it separates the small pair worse than the conditional does, by $-$0.29 \ci{$-$0.36}{$-$0.22}. The grids' count is thus present in the noisy latent where the conditional reads it only partly, but the observer may read the grid's periodic layout rather than its count, so the comparison is descriptive only.

\section{Showing the count}
\label{sec:supp_show}
\subsection{Saying against showing}
\label{sec:supp_saysow}

The main text's \cref{fig:saysow} shows both comparisons; this section gives every counter and the people's counts in full.

\para{Same noise, 12, 20 and 30 objects} (main text, \cref{fig:saysow}a). \sdm at $768^2$ on U2, seed 1, so every arm shares its prompt's noise with the plain image of the same cell. Saying: the number word, digits and the sentence ``$N$ $X$s arranged in $R$ rows of $C$'' that names the planted grid's rows and columns; showing: the content test's three starts at $\sigma{=}0.909$ on the 58 types with a cut-out. The said arms are given on those 58 types (as in main text, \cref{fig:saysow}a) and on all 64. Native resolution with the same seed gives 0.046 on the 58 types and 0.042 on all 64 (0.034 over both seeds; \cref{sec:supp_wording}).

\begin{table}[h]
\centering
\footnotesize
\setlength{\tabcolsep}{2.4pt}
\resizebox{\linewidth}{!}{%
\begin{tabular}{@{}lcccc@{}}
\toprule
exact at 12, 20, 30 & \samt & \gdino & \owl & consensus \\
\midrule
\multicolumn{5}{@{}l}{\emph{said, 58 types (all 64)}} \\
number word & .035 \ci{.011}{.063} (.031) & .011 (.016) & .029 (.026) & .023 (.021) \\
digits & .029 \ci{.006}{.055} (.026) & .006 (.005) & .006 (.005) & .006 (.005) \\
rows-and-columns sentence & .069 \ci{.035}{.105} (.062) & .069 (.062) & .035 (.031) & .046 (.042) \\
native resolution & .046 \ci{.017}{.083} (.042) & .035 (.031) & .052 (.047) & .029 (.026) \\
\midrule
\multicolumn{5}{@{}l}{\emph{shown, 58 types, start at $\sigma{=}0.909$}} \\
grid of ellipses & .236 \ci{.149}{.328} & .241 & .144 & .207 \\
grid of another type & .351 \ci{.260}{.446} & .305 & .172 & .264 \\
grid of the type itself & .517 \ci{.415}{.618} & .448 & .322 & .448 \\
\midrule
\multicolumn{5}{@{}l}{\emph{shown, 58 types, start at $\sigma{=}0.943$}} \\
grid of ellipses & .264 & .213 & .178 & .230 \\
grid of another type & .230 & .230 & .161 & .195 \\
grid of the type itself & .362 & .328 & .218 & .339 \\
\bottomrule
\end{tabular}}
\caption{\textbf{Saying and showing the same grid} (\sdm, U2, same noise). At $\sigma{=}0.909$ every counter puts each shown arm above each said arm and the type's own cut-out above the other two. Ratio of ellipses to the best said arm: 3.4 (\samt), 3.5 (\gdino), 2.8 (\owl, against native resolution) and 4.5 (consensus). \samt finds no requested object in 46\% of the ellipse images at $\sigma{=}0.909$.}
\label{tab:supp_saysow}
\end{table}

\para{People's counts by route} (main text, \cref{fig:saysow}b). Final human counts of both studies at the five complete-row counts (FLUX.1-dev: 12, 16, 20, 30), 16 types each. Within 10\%: $|n-N|\leq 0.1N$, that is within one object at 12 and 16, two at 20 and 25, three at 30.

\begin{table}[h]
\centering
\footnotesize
\setlength{\tabcolsep}{2.2pt}
\resizebox{\linewidth}{!}{%
\begin{tabular}{@{}lccccc@{}}
\toprule
people, 12--30 & exact & within one & within 10\% & recognisable & median $n/N$ \\
\midrule
\sdm $512^2$, word (S1) & .04 \ci{.00}{.10} & .05 & .06 & .03 & .49 \\
\sdm $768^2$, word & .04 \ci{.00}{.08} & .09 & .10 & .04 & .33 \\
\sdm, amplified word ($L{=}10$) & .04 \ci{.00}{.08} & .08 & .08 & .03 & .46 \\
\flux, word (S2) & .08 \ci{.02}{.16} & .14 & .14 & .06 & .49 \\
\sdxl, word (S2) & .04 \ci{.00}{.08} & .06 & .10 & .00 & 1.32 \\
\sdxl, Gu~\etal's first pass & .00 & .04 & .04 & .00 & .33 \\
\fluxk, word & .13 \ci{.06}{.19} & .21 & .23 & .08 & .89 \\
\midrule
\sdxl, Gu~\etal's re-layout & .11 \ci{.04}{.20} & .18 & .23 & .09 & .43 \\
\midrule
GLIGEN, boxes & .46 \ci{.31}{.60} & .50 & .50 & .18 & .92 \\
\sdm, self-planted & .59 \ci{.40}{.76} & .65 & .66 & .53 & 1.00 \\
InstanceAssemble, boxes & .68 \ci{.58}{.78} & .68 & .69 & .38 & 1.00 \\
\fluxk, planted & .93 \ci{.79}{1.00} & .94 & .94 & .68 & 1.00 \\
\bottomrule
\end{tabular}}
\caption{\textbf{People's counts by route.} U2 unless stated; 80 pictures per row (64 for \flux). Intervals: 95\% bootstrap over object types.}
\label{tab:supp_people_error}
\end{table}

\begin{table}[h]
\centering
\footnotesize
\setlength{\tabcolsep}{2.6pt}
\resizebox{\linewidth}{!}{%
\begin{tabular}{@{}lccc@{}}
\toprule
& \samt & \multicolumn{2}{c}{people, 16 types} \\
\cmidrule(lr){3-4}
exact at 12, 16, 20, 25, 30 (U2) & 64 types & exact & recognisable \\
\midrule
\multicolumn{4}{@{}l}{\emph{\sdm, same cells and noise}} \\
plain prompt & .022 & .04 \ci{.00}{.08} & .04 \ci{.00}{.08} \\
self-planting, grid & .470 & .59 \ci{.40}{.76} & .53 \ci{.35}{.70} \\
self-planting, scatter & .312 & & \\
\midrule
\multicolumn{4}{@{}l}{\emph{given the grid's boxes, from pure noise}} \\
GLIGEN (SD1.4, $512^2$) & .234 & .46 \ci{.31}{.60} & .18 \ci{.09}{.28} \\
InstanceAssemble (SD3-M, $1024^2$) & .372 & .68 \ci{.58}{.78} & .38 \ci{.24}{.51} \\
\midrule
\multicolumn{4}{@{}l}{\emph{\fluxk, same cells and noise}} \\
plain prompt & .050 & .13 \ci{.06}{.19} & .08 \ci{.03}{.13} \\
planted grid ($\sigma{=}0.941$) & .748$^a$ & .93 \ci{.79}{1.00} & .68 \ci{.50}{.84} \\
\midrule
\multicolumn{4}{@{}l}{\emph{\sdxl, Gu~\etal's pipeline~\cite{gu2026noise}, same cells and noise}} \\
first pass & .009 & .00 & .00 \\
noise re-layout & .075 & .11 \ci{.04}{.20} & .09 \ci{.03}{.16} \\
\bottomrule
\end{tabular}}
\caption{\textbf{Supplying a layout} ($^a$58 types with a cut-out). Recognisable: exact, and the objects clearly the requested one. On U2's clean grids \samt is exact on 76--79\% at each count. \samt picks self-planting's cut-outs and undercounts the layout models' images by 5--8 objects on average (self-planting: 3), so routes are compared by people's counts. People counted GLIGEN grounded at every step; by \samt the two schedules' exact rates differ by at most 0.02 overall.}
\label{tab:routes}
\end{table}

\para{Medians.} On GLIGEN's images \samt's median is 0 at every request while people's is 10, 16, 9.5, 25 and 27 (at 12, 16, 20, 25 and 30); on InstanceAssemble's, people's median equals the request at every count but 27 (26.5). People's medians (and \samt's) at 12, 16, 20, 25 and 30: \fluxk from the planted grids 12, 16, 20, 25, 30 (the same under \samt); \sdxl's first pass in Gu~\etal's pipeline 5 (1), 3.5 (1), 6.5 (2.5), 9 (2), 14.5 (3) and its re-layout 11 (8), 15 (15.5), 7.5 (1.5), 5 (1), 7 (9.5).

\begin{table}[h]
\centering
\footnotesize
\setlength{\tabcolsep}{2.4pt}
\resizebox{\linewidth}{!}{%
\begin{tabular}{@{}lccc@{}}
\toprule
16 types & count only & yes or partly & recognisable \\
\midrule
plain prompt$^\ast$ & .04 \ci{.00}{.08} & .04 \ci{.00}{.08} & .04 \ci{.00}{.08} \\
self-planting & .50 \ci{.33}{.67} & .50 \ci{.33}{.67} & .44 \ci{.28}{.60} \\
GLIGEN$^\ast$ & .46 \ci{.31}{.60} & .45 \ci{.30}{.60} & .18 \ci{.09}{.28} \\
InstanceAssemble & .65 \ci{.56}{.74} & .64 \ci{.54}{.73} & .38 \ci{.26}{.50} \\
\midrule
self-planting, complete counts & .59 \ci{.40}{.76} & .59 \ci{.40}{.76} & .53 \ci{.35}{.70} \\
self-planting, partial counts & .39 \ci{.19}{.61} & .39 \ci{.19}{.61} & .33 \ci{.16}{.52} \\
InstanceAssemble, complete & .68 \ci{.58}{.78} & .66 \ci{.55}{.76} & .38 \ci{.24}{.51} \\
InstanceAssemble, partial & .63 \ci{.50}{.73} & .61 \ci{.47}{.73} & .38 \ci{.23}{.53} \\
\midrule
self-planting $-$ plain$^\ast$ & .55 \ci{.36}{.74} & .55 \ci{.36}{.74} & .49 \ci{.30}{.66} \\
self-planting $-$ GLIGEN$^\ast$ & .13 \ci{$-$.06}{.31} & .14 \ci{$-$.06}{.33} & .35 \ci{.16}{.54} \\
self-planting $-$ InstanceAssemble & $-$.15 \ci{$-$.34}{.03} & $-$.14 \ci{$-$.32}{.04} & .06 \ci{$-$.09}{.23} \\
\midrule
\multicolumn{4}{@{}l}{\emph{second count}$^\ast$} \\
\fluxk plain & .13 \ci{.06}{.19} & .13 \ci{.06}{.19} & .08 \ci{.03}{.13} \\
\fluxk planted & .93 \ci{.79}{1.00} & .91 \ci{.76}{1.00} & .68 \ci{.50}{.84} \\
\sdxl first pass (Gu~\etal) & .00 & .00 & .00 \\
noise re-layout & .11 \ci{.04}{.20} & .11 \ci{.04}{.20} & .09 \ci{.03}{.16} \\
\sdm, word amplified & .04 \ci{.00}{.08} & .04 \ci{.00}{.08} & .03 \ci{.00}{.06} \\
\fluxk planted $-$ plain & .80 \ci{.66}{.90} & .79 \ci{.65}{.90} & .60 \ci{.41}{.76} \\
\fluxk planted $-$ self-planting & .34 \ci{.19}{.50} & .33 \ci{.18}{.49} & .15 \ci{$-$.01}{.30} \\
re-layout $-$ first pass & .11 \ci{.04}{.20} & .11 \ci{.04}{.20} & .09 \ci{.03}{.16} \\
amplified $-$ plain \sdm & .00 \ci{$-$.04}{.04} & .00 \ci{$-$.04}{.04} & $-$.01 \ci{$-$.04}{.00} \\
amplified $-$ self-planting & $-$.55 \ci{$-$.73}{$-$.38} & $-$.55 \ci{$-$.73}{$-$.38} & $-$.50 \ci{$-$.68}{$-$.33} \\
\bottomrule
\end{tabular}}
\caption{\textbf{The methods, by human count} (16 types of U2). Exact at 12--30; $^\ast$12, 16, 20, 25, 30 only, differences on the same cells. Count only: the human count equals the request; yes or partly / recognisable: and the majority look-alike answer is ``yes, clearly'' or ``partly'' / ``yes, clearly''. \samt on the same pictures: .04, .38, .18, .40. Look-alike shares where objects were marked (yes / partly / no): self-planting .72 / .27 / .01, InstanceAssemble .58 / .40 / .02, GLIGEN .41 / .57 / .02, plain .72 / .28 / .00. Second count, in the order listed: \samt .05, .71, .03, .06, .01; look-alike .84 / .16 / .00, .72 / .27 / .01, .68 / .32 / .00, .75 / .25 / .00, .64 / .36 / .00.}
\label{tab:supp_hc_method}
\end{table}

Every route that plants the grid or gives its boxes draws the asked count far more often than any way of saying it (\cref{tab:supp_people_error,tab:supp_hc_method}), and \fluxk from a planted grid most often of all.

\subsection{What the start must show}
\label{sec:supp_content}
For each of U2's 58 types with a cut-out and each count 12, 16, 20, 25 and 30 (seed 1), the self-planted grid (same centres, object size, jitter and canvas) is filled in three ways: the type's own cut-out (the set's composite, unchanged); the cut-out of a partner type, the other type with a cut-out whose crop has the closest aspect ratio, on the partner's own ground; or soft ellipses of the own cut-out's size, filled with its mean colour under the \samt mask, on the own ground. The prompt always names the type and the noise is the set's. Starts at step 9 ($\sigma{=}0.909$, self-planting's level) and step 6 (0.943); 1,740 images.

\begin{table}[h]
\centering
\footnotesize
\setlength{\tabcolsep}{3pt}
\begin{tabular}{@{}lccc@{}}
\toprule
start shows & $\sigma{=}0.909$ & $\sigma{=}0.943$ & object drawn \\
\midrule
own cut-out & .531 \ci{.43}{.63} & .393 & .81 \\
another type's cut-out & .345 \ci{.26}{.44} & .234 & .63 \\
mean-colour ellipses & .245 \ci{.16}{.33} & .276 & .54 \\
\midrule
own $-$ another & $+$.186 \ci{.08}{.29} & $+$.159 & \\
own $-$ ellipses & $+$.286 \ci{.19}{.38} & $+$.117 & \\
another $-$ ellipses & $+$.100 \ci{$-$.01}{.21} & $-$.041 & \\
\bottomrule
\end{tabular}
\caption{\textbf{What the start must show} (\samt, 290 cells per row; exact: exactly $N$ of the asked type). Object drawn: \samt finds the asked type at least once ($\sigma{=}0.909$). Under \gdino, \owl and the consensus, own minus another is $+$0.14, $+$0.16 and $+$0.21 and own minus ellipses $+$0.23, $+$0.22 and $+$0.26 at $\sigma{=}0.909$; the order own $>$ another $>$ ellipses holds under every counter.}
\label{tab:supp_content}
\end{table}

By count (own / another / ellipses, $\sigma{=}0.909$): 0.55 / 0.48 / 0.31 at 12, 0.52 / 0.38 / 0.31 at 16, 0.53 / 0.36 / 0.21 at 20, 0.59 / 0.29 / 0.21 at 25 and 0.47 / 0.21 / 0.19 at 30: the own cut-out holds across counts while the others fall. By class: objects on a plain ground 0.68 / 0.54 / 0.39, scene-like 0.47 / 0.27 / 0.13, thin 0.47 / 0.24 / 0.33. On the eight types whose mean colour barely differs from the ground, every start does poorly (0.15 / 0.10 / 0.08). At the noisier $\sigma{=}0.943$ the ellipses are redrawn as the object more often (object drawn in 79\% of images) and the three starts move closer together. A start sets the count best when it shows the asked object itself.

\subsection{The conflict test and the object test}
\label{sec:supp_conflict}
\begin{figure*}[t]
\centering
\includegraphics[width=0.80\linewidth]{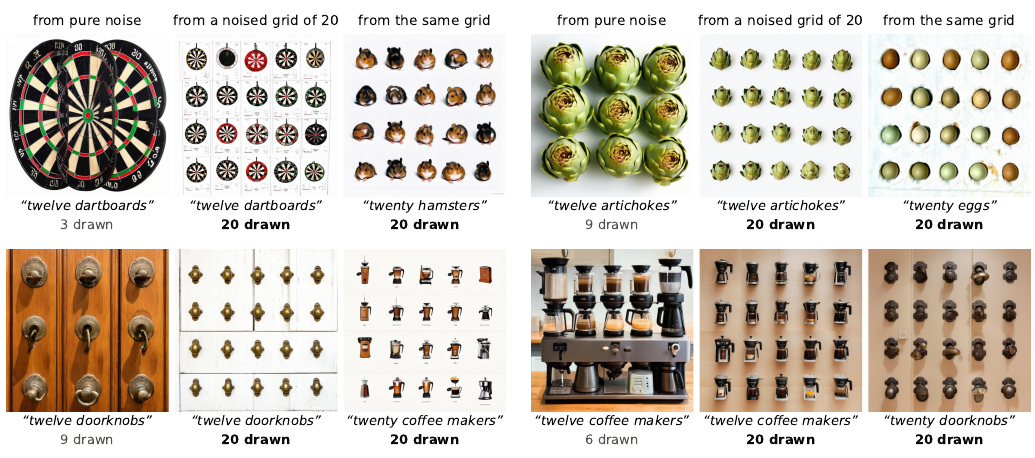}
\caption{\textbf{A planted count holds against the number word; a planted object does not hold against the object word.} Each triplet shares one starting noise (\sdm, S1). Left: asked for twelve objects from pure noise. Middle: started from a noised grid of twenty of the same objects ($\sigma{=}0.909$) and asked for twelve. Right: started from the same grid and asked for twenty of another object. Over all 48 types, the asked number is drawn in 2\% of conflict images (4\% for a grid of twenty asked for twelve), and exactly twenty of the newly named object in 27\%. Cells chosen by inspection among those that \samt counts at twenty in both right-hand images; counts by \samt.}
\label{fig:split}
\end{figure*}
In the conflict test (\cref{fig:split}, middle) a planted grid meets a different number word; in the object test (\cref{fig:split}, right) it meets a different object word. The object test's change can keep the layout, so SDEdit~\cite{meng2022sdedit} and Prompt-to-Prompt~\cite{hertz2023prompt} lead one to expect it to succeed.

\begin{table}[h]
\centering
\small
\setlength{\tabcolsep}{3.4pt}
\begin{tabular}{@{}llccccc@{}}
\toprule
& & & \multicolumn{2}{c}{planted $N$ kept} & word & shift \\
\cmidrule(lr){4-5}
Model & set & $\sigma$ & conflict & agree & wins & to word \\
\midrule
\sdm & S1 & .909 & .52 & .52 & .021 & $-$.03 \\
\sdm & S2 & .909 & .65 & .66 & .010 & $-$.02 \\
\sdxl & S2 & .881 & .59 & .57 & .007 & $-$.08 \\
\flux & S2 & .933 & .58 & .65 & .035 & $-$.02 \\
\bottomrule
\end{tabular}
\caption{\textbf{Conflict test} (screened sets). A grid of $N\in\{12,20,30\}$ objects is planted and the prompt asks for a different count $M$ (conflict) or for $N$ (agree). Word wins: the asked $M$ is drawn. Shift: paired change in log count toward $M$, conflict minus agree (95\% upper bounds $\leq$\,0.11). Plain prompts from the same noise draw $N$ in 1--2\% of these cells, so net of chance the plant is kept in 0.50 \ci{.39}{.60}, 0.64 \ci{.52}{.74}, 0.58 \ci{.46}{.69} and 0.57 \ci{.45}{.69}. 288 conflict cells per \sdm row, 144 for \sdxl and \flux; \samt.}
\label{tab:conflict}
\end{table}

The word wins at a request of twelve in 3.1\%, 1.6\%, 1.0\% and 5.2\% of conflict cells (rows of \cref{tab:conflict} in order), below the rate at which the same request hits twelve from pure noise on the same sets (7.3\%, 5.2\%, 6.2\% and 12.5\%); at a request of thirty both are zero. Kept rates by planted count in the conflict cells are 0.52, 0.55 and 0.48 (\sdm, S1), 0.70, 0.70 and 0.56 (\sdm, S2), 0.60, 0.63 and 0.54 (\sdxl) and 0.63, 0.58 and 0.52 (\flux) at 12, 20 and 30.

There is no pull toward the word either: the paired change in $\log(1{+}n)$ between a conflicting request and the agreeing request on the same plant and noise (12 asked for 30, 20 or 30 asked for 12) lies between $-$0.19 and $+$0.18 in every model and set, with every interval including zero.

\para{Start levels in \sdxl and \flux.} Start levels were calibrated on object types outside the sets, as the largest level at which a planted grid is kept in at least half of the cells, applied to nominal levels; on the families' own step grids they start at $\sigma{=}0.881$ for \sdxl (nominal 0.89, step 4 of 40) and 0.933 for \flux (nominal 0.95, step 4 of 28). For \sdxl the calibration decided between neighbouring levels; for \flux the earliest level tried already met it, so its start may be later than needed. \sdxl's noise scale is matched to these levels by signal-to-noise ratio, and both families receive \sdm's composites. \sdxl and \flux were tested at this one start level each, and at large counts only.

\para{Other requests and other counters.} With a planted 20 and requests of 6, 9, 16, 25, 30 or 36 instead, the requested number is drawn in 0.7\% of images; across these requests and a prompt with no number, the count equals that of the agreeing render in 58\% \ci{47}{69} of images. \gdino and \owl undercount large grids even on clean composites; on S2's cells whose clean composite all three counters count exactly, the net kept rate is 0.66, 0.62 and 0.43 under \samt, \gdino and \owl.

\para{Irregular arrangements.} Planted irregular arrangements in the conflict design on S1 are kept net of chance in 0.24 of cells for a hole-free scatter (0.45 for objects on a plain ground, 0.20--0.26 under the other counters; 0.12 and 0.08 for scene-like and thin ones) and 0.23 for a jittered grid, under \samt.

\para{The object test.} On S1, each type's grid of twenty is paired with a partner type by a greedy matching, improved by pairwise swaps, that keeps the median aspect ratio and size of their detected boxes close ($|\Delta\log\text{aspect}| + 0.5\,|\Delta\log\text{side}|$), and the grid is asked for twenty of the partner (``a photo of twenty hamsters'' on a grid of dartboards; $\sigma{=}0.909$, two seeds, 96 images). \samt is prompted with both names, and each detected object goes to the name with the higher score where the two names' boxes overlap (IoU ${\geq}\,0.5$). The prompt's object takes most of the detected objects in 75\% \ci{65}{84} of images, and in 77\% without two near-white images of one cell (a spoon grid asked for meatballs); this readout uses \samt alone. The count suffers: exactly twenty of the prompt's object appear in 27.1\% \ci{17.7}{37.5} of these images, against 56.2\% when the same grids are asked for their own object (paired difference 0.29 \ci{0.18}{0.41}). From pure noise, asked for twenty red or blue objects of the same set, the model makes the asked colour dominant in 95\% \ci{90}{99} of images, draws the object in 74\% and the count exactly in 1\%.

A planted count holds against the number word, while the object word repaints the planted grid as its own object.

\subsection{Self-planting details}
\label{sec:supp_method}
\begin{figure*}[h]
\centering
\includegraphics[width=0.64\linewidth]{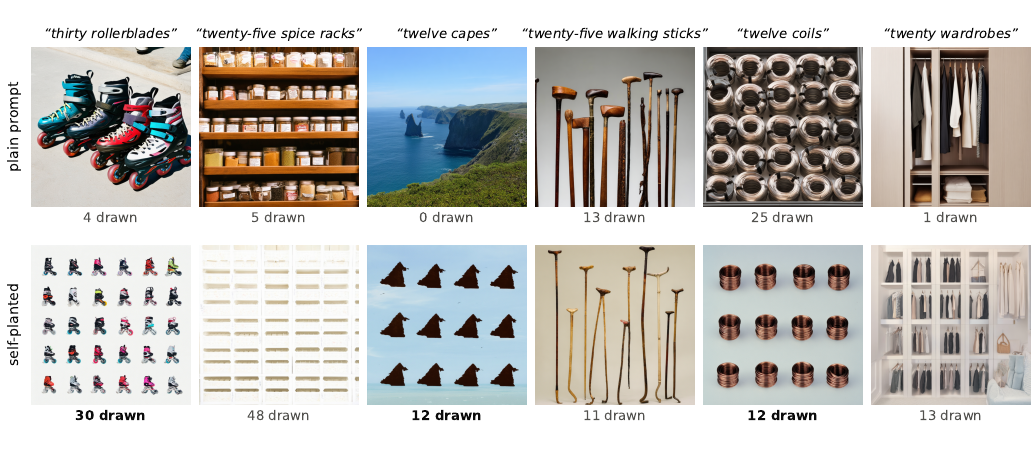}
\caption{\textbf{Self-planting on six images drawn at random} from the 80 people-counted images of U2 at the counts whose grids have only complete rows (16 types $\times$ 5 counts). Top: plain prompt; bottom: the same prompt and noise from a self-planted start. Counts by people; bold: exact. Three of the six are exact (59\% over all 80 cells). The failures show how a planted grid goes wrong: scene-like objects turn into a scene (shelving for spice racks, a closet for wardrobes), and thin objects lose their rows.}
\label{fig:random}
\end{figure*}
\begin{figure*}[h]
\centering
\includegraphics[width=0.70\linewidth]{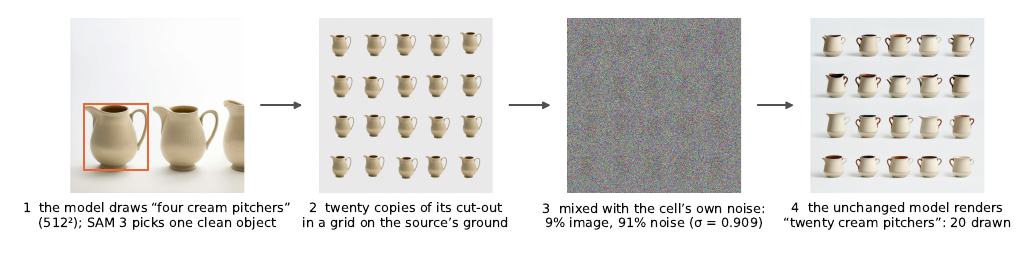}
\caption{\textbf{Self-planting.} The model's own small-count drawing supplies the object, a grid supplies the count, and the unchanged model, started at $\sigma{=}0.909$ from the cell's own noise, supplies everything else. The plain prompt from the same noise draws 10 cream pitchers. One clean instance suffices; the canvas takes the source image's border colour. Cell from U1, chosen to illustrate the steps.}
\label{fig:method}
\end{figure*}

\para{Settings.} The start level (step 9 of 40), the grid's spacing and object size and the cut-out rule were set on object types outside all sets. The canvas takes the source image's border colour, the ground on which the object was drawn. The six small drawings per type cost less than three plain $768^2$ images, and each self-planted image costs 31 of the 40 steps of one.

\para{Grid.} $N$ copies on a $768^2$ canvas filled with the source image's border colour, in $R=\mathrm{round}(\sqrt N)$ rows of $C=\lceil N/R\rceil$ (with $R$ reduced while $C(R{-}1)\geq N$) and a centred partial last row; margins of 4\% of the canvas, cells of $(W-2m)/C$ by $(W-2m)/R$; the object's long side is the cell width divided by 1.45 (capped by the object's natural size), and centres are jittered by $\mathcal{N}(0,(0.04d)^2)$ for object size $d$.

\para{Cut-out rule.} Among the six small-count draws (``two'', ``three'' and ``four'', two seeds, $512^2$), \samt detections are ranked in three tiers. Tier 1: score ${\geq}\,0.8$, overlap (IoU) ${\leq}\,0.05$ with every other detection of score ${\geq}\,0.3$, both box sides ${\geq}\,40$\,px, not touching the border. Tier 2: score ${\geq}\,0.6$, IoU ${\leq}\,0.20$, sides ${\geq}\,32$\,px, not touching the border. Tier 3: as tier 2, touching allowed. The best tier wins, then the highest score; the box is clamped to the image. On U1, 45, 6 and 6 types received tier 1, 2 and 3 cut-outs (exact 48\%, 27\% and 19\%), and 7 received none; on U2 45, 8 and 5 (none for 6), on U3 61 in all. Types without a cut-out count with their plain image in every rate.

\para{Arrangements.} All arrangements of a cell share the cut-out and its size $d$. \emph{Hole-free scatter}: Lloyd relaxation of random points with jitter, best of four starts by hole statistics, centres ${\geq}\,1.2d$ apart. \emph{Random placement}: random sequential addition with centres ${\geq}\,1.3d$ apart, $d$ shrinking by 3\% after each failed attempt. \emph{Grid with random scale and offset}: the grid scaled about the canvas centre by a factor in $[0.55,1]$ and shifted.

\begin{table}[h]
\centering
\small
\setlength{\tabcolsep}{5pt}
\begin{tabular}{@{}lccc@{}}
\toprule
U1 & \samt & consensus & $\Delta$ vs.\ grid \\
\midrule
plain prompt & .019 & .009 & $-$.359 \\
best of two plain samples & .036 & -- & $-$.343 \\
\midrule
\multicolumn{4}{@{}l}{\emph{self-planted start, same cut-out and object size}}\\
random placement & .199 & -- & $-$.180 \\
hole-free scatter & .273 & -- & $-$.105 \\
grid, random scale and offset & .343 & -- & $-$.036 \\
\textbf{grid (self-planting)} & \textbf{.378} & \textbf{.371} & \\
\bottomrule
\end{tabular}
\caption{\textbf{Self-planting on U1.} Exact rate at 12--30 objects (12, 13, 17, 20, 23, 29, 30; two seeds; 896 cells per row). $\Delta$: paired difference to the grid; 95\% intervals for the grid \ci{.291}{.468}, and for $\Delta$ \ci{$-$.445}{$-$.275}, \ci{$-$.429}{$-$.260}, \ci{$-$.233}{$-$.129}, \ci{$-$.146}{$-$.066}, \ci{$-$.065}{$-$.008}.}
\label{tab:supp_u1arms}
\end{table}

\para{Counters.} \samt also selects the cut-out, so its judgement of self-planted images is not independent of the method; the other counters and people are. Under the other counters self-planting reaches 0.27--0.36 on U2 (\cref{tab:supp_gligen}).

\para{Rows and the object type.}\label{sec:supp_rows}
Reading rows from the \samt boxes on U1, a complete last row of a planted grid comes out right in 99--100\% of the cells whose upper rows are intact, a partial one in 64--81\%; the grid beats a hole-free scatter of the same cut-outs by 0.22 \ci{0.15}{0.29} at counts with complete rows and by 0.04 \ci{0.00}{0.09} at the others (U2: 0.19 \ci{0.13}{0.26} at 12, 20 and 30). Types that would have passed the screened sets' visibility check reach 0.46--0.50 on the three unscreened sets, the others 0.19--0.33; in a variance decomposition of U1's grids the object type accounts for 52\% of the variance of the per-image exact outcome and the count for 7\%.

\para{The third unscreened set.}\label{sec:supp_u3}
On U3 (64 types, mostly furniture, appliances and animals; 61 with a cut-out; two seeds) self-planting is exact in 21--54\% of images at each count from 12 to 30, against 0--10\% for the plain prompt from the same noise; by class, in 69\% of images of objects drawn on a plain ground, 36\% of thin objects and 27\% of scene-like ones.

\para{Look.} PickScore~\cite{kirstain2023pickscore} of the self-planted image minus the plain image of the same cell: $-0.65$ \ci{$-$0.80}{$-$0.51}, $-0.57$ \ci{$-$0.72}{$-$0.43} and $-0.85$ \ci{$-$1.03}{$-$0.68} on the three unscreened sets. On U2, exact renders lose less ($-0.45$) than failed ones ($-0.67$), and the grid scores above the hole-free scatter ($+0.07$ \ci{0.03}{0.12}).

A complete grid of the model's own object sets the count best; the object type matters more than the count, and scene-like and thin objects fail most.

\subsection{Two layout models given the same grid}
\label{sec:supp_gligen}
GLIGEN~\cite{li2023gligen} with the public text-box checkpoint (\texttt{gligen-1-4-generation-text-box}, revision \texttt{d2820dc}, a \textsc{SD}~1.4 backbone) in diffusers, at $512^2$ with 50 steps and guidance 7.5. For every cell of U2 it receives the boxes of the self-planted grid: each grid centre as a square box of the grid's object size, one phrase per box (``a cream pitcher''), and the prompt ``a photo of twenty cream pitchers''. We use two grounding schedules from the same noise: the authors' scheduled sampling, which grounds the first 30\% of steps ($\beta{=}0.3$), and grounding at every step ($\beta{=}1$). The checkpoint's NSFW filter, which blanks an image when it fires, is off, as \sdm has none. Two seeds per cell, 2,304 images, counted like every other image. Where \samt finds the object, GLIGEN turns partial grids into rectangles: given the 14 boxes of a $4{\times}4$ grid with two missing, it draws 16 objects in 76\% of those images (82\% with $\beta{=}0.3$) and 14 in none.

InstanceAssemble~\cite{xiang2025instanceassemble}, \textsc{SD3}-Medium version, with the authors' released code (commit \texttt{39f98cd}, run unchanged) and weights, their pinned library versions and their default settings: 50 steps, guidance 7.0, grounding for the first 30\% of steps, layout scale 1.0, bf16, at its native $1024^2$. It receives the same boxes (scaled to $1024^2$), one phrase per box and the same prompt as GLIGEN; the generator is seeded on the GPU as in the authors' script. All layout weights load as in the authors' own loader. 1,152 images, two seeds per cell. Where \samt finds the object, InstanceAssemble is exact in 80--90\% of images at complete counts and 57--60\% at partial ones; at 14 boxes it draws 16 objects in 23\% of those images and 14 in 58\%, and it never drops a partial row.

\begin{figure}[h]
\centering
\includegraphics[width=\linewidth]{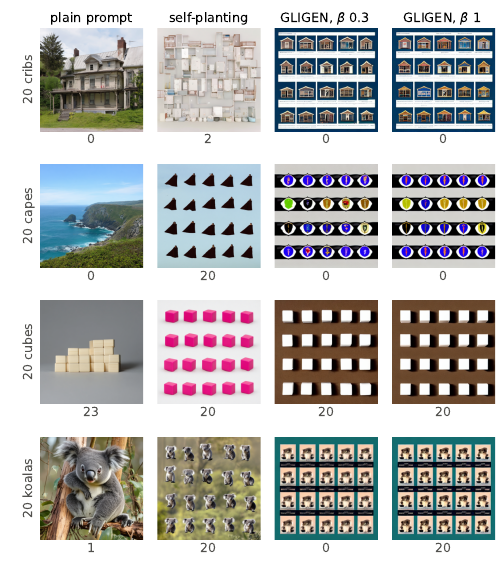}
\caption{\textbf{The same cells drawn three ways} (the first four types with a cut-out of U2, asked for 20, seed 1; \samt count under each image). GLIGEN fills each box of the grid, but often with an icon of something else: houses for cribs, ovals for capes. The koala tiles are counted in one schedule and not in the other.}
\label{fig:gligen}
\end{figure}

\begin{table}[h]
\centering
\footnotesize
\setlength{\tabcolsep}{3pt}
\resizebox{\linewidth}{!}{%
\begin{tabular}{@{}lcccc@{}}
\toprule
64 types, 12--30 & GLIGEN $\beta{=}0.3$ & GLIGEN $\beta{=}1$ & InstanceAssemble & self-planting \\
\midrule
\samt & .158 & .152 & .324 & .406 \\
\gdino & .188 & .212 & .457 & .351 \\
\owl & .122 & .118 & .139 & .273 \\
consensus & .134 & .135 & .300 & .360 \\
\midrule
complete counts (12, 16, 20, 25, 30) & .253 & .234 & .372 & .470 \\
partial counts (14, 19, 22, 27) & .039 & .049 & .264 & .326 \\
studio / thin / scene types & & .26 / .11 / .14 & .45 / .21 / .35 & .63 / .43 / .35 \\
within one of the request & .178 & .166 & .343 & .464 \\
none of the asked object & .683 & .708 & .550 & .248 \\
\bottomrule
\end{tabular}}
\caption{\textbf{Two layout models given the self-planted grid} (U2; exact rates unless stated; \samt below the first block; none of the asked object: \samt finds none). For GLIGEN the difference to self-planting, paired on type, count and seed, is 0.25 under \samt for either schedule (\ci{0.17}{0.33} and \ci{0.17}{0.34}) and 0.14--0.23 under the other counters, with every interval above zero. For InstanceAssemble it is 0.08 \ci{$-$0.00}{0.17} under \samt, 0.06 \ci{$-$0.02}{0.15} under the consensus, 0.13 \ci{0.07}{0.21} under \owl and $-$0.11 \ci{$-$0.20}{$-$0.01} under \gdino; the counters disagree on InstanceAssemble's images more than on any other method's, and we compare the methods by people's counts (\cref{tab:supp_hc_method}).}
\label{tab:supp_gligen}
\end{table}

By people, InstanceAssemble given the grid's boxes is exact as often as self-planting within the intervals, while GLIGEN fills the boxes but often with something else (\cref{tab:supp_hc_method}).

\subsection{\fluxk from a planted start}
\label{sec:supp_fluxk_plant}
\para{Design.} The composites are U2's own-cut-out grids of the content test (\cref{sec:supp_content}), that is, \sdm's cut-outs on their $768^2$ canvas, for the 58 types with a cut-out at 12, 16, 20, 25 and 30 objects (290 starts). Each is resized to $1024^2$, encoded as the pipeline encodes reference images (the VAE's mode, $2{\times}2$ patches, the VAE's normalisation statistics) and mixed with the cell's own noise at the level of step $k$ of the model's 50-step schedule (\cref{eq:start}); the unchanged pipeline then runs the remaining $50-k$ steps of that schedule with the plain prompt. Each render therefore shares prompt and noise with the plain image of the same cell in \cref{sec:supp_fluxk}.

\para{Start level.} The level was set on 24 development grids of twenty objects (U1 types, none in U2): the smallest $k\in\{16,20,24,28\}$ at which \samt counts exactly twenty in at least half of the renders. \samt kept 0.79, 0.83, 0.83 and 0.88, so $k{=}16$ ($\sigma{=}0.941$). As checks, $k{=}0$ reproduces the word test's sampler and $k{=}46$ returns the composite.

\para{Flat renders.} A check for flat images (grey standard deviation below 0.03, or mean outside $[0.03,0.97]$) flags 26 of the 290 U2 renders, which are kept. The 14 among the 16 people-counted types are low-contrast grids (pale bath mats, spice racks, white ironing boards, and desks whose cut-outs look like stools). People find the asked count in 9 of them, none of them desks (they count twice the request or none), and judge the ironing boards at 30 not to look like ironing boards.

\begin{table}[h]
\centering
\footnotesize
\setlength{\tabcolsep}{3pt}
\resizebox{\linewidth}{!}{%
\begin{tabular}{@{}lccccc@{}}
\toprule
counter & planted & plain & planted $-$ plain & order, planted & none, planted \\
\midrule
\samt & .748 \ci{.65}{.84} & .055 & $+$.69 \ci{.59}{.79} & .92 \ci{.87}{.96} & .23 \\
\gdino & .672 \ci{.58}{.76} & .031 & $+$.64 \ci{.55}{.73} & .80 \ci{.72}{.88} & .17 \\
\owl & .434 \ci{.33}{.53} & .028 & $+$.41 \ci{.31}{.51} & .84 \ci{.77}{.90} & .31 \\
consensus & .638 \ci{.53}{.74} & .028 & $+$.61 \ci{.50}{.71} & .93 \ci{.87}{.97} & .21 \\
\midrule
people (16 types) & .93 \ci{.79}{1.00} & .13 & $+$.80 \ci{.66}{.90} & & \\
\bottomrule
\end{tabular}}
\caption{\textbf{\fluxk from the planted grids} (U2, 58 types with a cut-out $\times$ 12, 16, 20, 25, 30, seed 1; exact rates unless stated; plain: the word test's image of the same cell). By count under \samt: .74, .78, .72, .78, .72, each with the request as median. People: recognisable .68 \ci{.50}{.84}; planted minus \sdm's self-planting on the same cells and grids $+$.34 \ci{.19}{.50} (recognisable $+$.15 \ci{$-$.01}{.30}).}
\label{tab:supp_fluxk_plant}
\end{table}

\samt's zeros (68 of 290 renders) concentrate in a few types: it reads every count of ironing boards, visors, cappuccinos, hairnets, coffee tables, blankets and futons as empty. Where it finds the object at all, it nearly always finds the asked number. People's median equals the request at every count, and they are exact in 88--94\% of renders at each count. The renders keep the composite's look, cut-outs on a flat ground. From a planted grid, \fluxk draws the asked count in nearly every image by people's counts.

\subsection{Noise re-layout past ten}
\label{sec:supp_gu}
\para{Method and settings.} Gu~\etal~\cite{gu2026noise} with the released code (commit \texttt{d5d49ba}, single-object pipeline, the README's library versions), \sdxl base~1.0 in fp16 with its DDPM scheduler, 50 steps, guidance 5.0, $1024^2$ and every released hyperparameter. A first pass from the cell's noise is counted by Grounding DINO, the method's gate, and an image already at the asked count is kept. Otherwise the object token's cross-attention, averaged over the first 20 steps, is clustered into the requested number of high-attention regions, the noise is re-laid around them, and a second pass guided by an attention loss over its first 20 steps renders the result. Documented differences from the release: (1) the gate runs the Hugging Face port of the same Grounding DINO weights, because the released call passes undefined names; (2) the object token is the noun's first token after the number word, which is what the released script's fixed index selects for its own prompts; (3) with fewer high-attention cells than requested, the released fallback of one cluster per cell is kept and the re-layout runs over those clusters, where the released loop indexes past its end and stops; (4) one seed per type, shared by its requests, instead of 42 for every prompt; (5) no attention-map figures; (6) self-attention runs PyTorch's fused attention, the same function, which fits the guided pass into an L4's memory at $1024^2$. A method error other than running out of memory leaves the first pass as the output. Settings were set after a probe on two types outside the sets.

\para{Cells.} U2's 64 types at the method's own range, 4 and 8, and at 12, 16, 20, 25 and 30, seed 1: 448 cells, each with the first pass and the method's output from the same noise, counted like every other image. People counted both at 12--30 for 16 types (\cref{sec:supp_human}).

\para{Results.} \begin{table}[h]
\centering
\footnotesize
\setlength{\tabcolsep}{2.4pt}
\resizebox{\linewidth}{!}{%
\begin{tabular}{@{}lcccc@{}}
\toprule
& exact, 4 and 8 & exact, 12--30 & output $-$ first & order, 12--30 \\
counter & first / output & first / output & at 12--30 & first / output \\
\midrule
\samt & .164 / .422 & .009 / .075 & .07 \ci{.03}{.10} & .63 / .58 \\
\gdino & .156 / .422 & .003 / .059 & .06 \ci{.03}{.08} & .48 / .47 \\
\owl & .125 / .258 & .016 / .037 & .02 \ci{$-$.01}{.05} & .58 / .55 \\
consensus & .109 / .383 & .000 / .056 & .06 \ci{.03}{.08} & .53 / .54 \\
\bottomrule
\end{tabular}}
\caption{\textbf{Noise re-layout on \sdxl} (U2, 64 types, seed 1; first: the method's first pass from the cell's noise; output: the method's image from the same noise). 12--30: the requests 12, 16, 20, 25 and 30. Output minus first at 4 and 8: .26 \ci{.16}{.36} / .27 \ci{.18}{.35} / .13 \ci{.06}{.20} / .27 \ci{.19}{.36} (counters in the table's order).}
\label{tab:supp_gu}
\end{table}

Of the 448 cells, the gate kept the first pass in 22 and 0 ended in a method error. The attention offered fewer clusters than requested in 0 cells, so the released fallback (difference 3) never applied. Median time per cell (first pass and re-layout) 117\,s; peak memory 16.5\,GB.

People counted both images at 12--30 for 16 types (\cref{tab:supp_hc_method}): the re-layout is exact in 11\% \ci{4}{20} of them and the first pass in none. Their median counts are 11, 15, 7.5, 5 and 7 for the outputs and 5, 3.5, 6.5, 9 and 14.5 for the first passes at 12, 16, 20, 25 and 30, and their order at 12--30 is 0.50 \ci{0.36}{0.64} for the outputs and 0.63 \ci{0.48}{0.74} for the first passes. Re-laying the noise helps a little past ten, far less than planting a grid of the object.

\subsection{Exact or within one}
\label{sec:supp_within}
A layout misses differently from a word. Of the images whose \samt count lands within one of the request, 80--94\% are exact for grids planted at $\sigma{=}0.909$ or given as boxes, against 66--77\% for irregular arrangements of the same cut-outs and 37--48\% for the plain prompt; a count that followed the request with log-normal scatter of 5\% or more would give at most 0.48 (\cref{tab:supp_within}). When a grid fails, it mostly fails by more than one object.
\begin{table}[h]
\centering
\footnotesize
\setlength{\tabcolsep}{3pt}
\resizebox{\linewidth}{!}{%
\begin{tabular}{@{}lccc@{}}
\toprule
\samt, 12--30 & exact & within one & ratio \\
\midrule
\multicolumn{4}{@{}l}{\emph{grids, planted at $\sigma{=}0.909$ or given as boxes}} \\
self-planting, U1 / U2 / U3 & .378 / .406 / .366 & .443 / .464 / .433 & .85 / .88 / .85 \\
grid, random scale and offset (U1) & .343 & .429 & .80 \\
content: own / another / ellipses (U2) & .531 / .345 / .245 & .583 / .407 / .297 & .91 / .85 / .83 \\
GLIGEN $\beta{=}0.3$ / $1$ (U2) & .158 / .152 & .178 / .166 & .89 / .92 \\
InstanceAssemble (U2) & .324 & .343 & .94 \\
\midrule
\multicolumn{4}{@{}l}{\emph{other layouts}} \\
hole-free scatter, U1 / U2 & .273 / .315 & .384 / .411 & .71 / .77 \\
random placement (U1) & .199 & .301 & .66 \\
content at $\sigma{=}0.943$: own / another / ellipses & .393 / .234 / .276 & .452 / .303 / .317 & .87 / .77 / .87 \\
\midrule
\multicolumn{4}{@{}l}{\emph{asking}} \\
plain prompt, U1 / U2 / U3 & .019 / .014 / .021 & .051 / .030 / .043 & .37 / .47 / .48 \\
\emph{reference, log-normal scatter .05 / .3} & .40 / .07 & .85 / .21 & .47 / .34 \\
\bottomrule
\end{tabular}}
\caption{\textbf{Exact or within one.} Ratio: exact rate over the rate within one of the asked count. A count that follows the request with log-normal scatter of SD 0.05 or more gives a ratio of at most 0.48 at these counts.}
\label{tab:supp_within}
\end{table}

\section{Notes}
\label{sec:supp_notes}
\para{Reproducibility.} The code, analysis scripts, object lists, generated images, automatic and human counts and the people's answers will be released. The people's studies collected no personal data (\cref{sec:supp_hcprotocol}).

\subsection{Object types}
\label{sec:supp_nouns}
Each object set is drawn in a seeded order from LVIS category names~\cite{gupta2019lvis}, kept to countable everyday objects from the names alone; a candidate that collides by name, WordNet relation~\cite{miller1995wordnet} or string overlap with any earlier type is dropped, so no type name repeats across sets or comes from development (``walking cane'' in S1 and ``walking stick'' in U2 share an LVIS definition).
\footnotesize
\para{S1.} dartboard, tea bag, figurine, laptop computer, binder, sled, globe, hamster, barrette, walking cane, pickle, snowboard, doorknob, remote control, blender, grater, spoon, microphone, kettle, toaster, scrubbing brush, ashtray, flowerpot, rubber band, notepad, bobby pin, handbag, egg, saucer, pizza, artichoke, mop, clothespin, computer keyboard, meatball, videotape, ski, wrench, skateboard, marker, zucchini, colander, coffee maker, dog collar, earring, canister, mask, bobbin.

\para{S2.} beachball, amplifier, saucepan, tennis racket, pouch, soccer ball, compass, cymbal, gravy boat, ice pack, joystick, nest, poster, mirror, bolt, business card, egg roll, shot glass, corkboard, persimmon, apricot, microscope, sewing machine, poker chip, almond, saxophone, measuring cup, gemstone, thimble, hockey stick, prune, fan, hook, radio receiver, piggy bank, thermostat, nailfile, bouquet, food processor, squirrel, green bean, painting, taco, softball, papaya, flag, milk can, projector.

\para{U1.} tobacco pipe, clasp, robe, crucifix, turban, lab coat, beanie, vest, pitchfork, dropper, cream pitcher, knee pad, skirt, crouton, duffel bag, raft, birthday card, dragonfly, tricycle, shopping bag, hair curler, swimsuit, latch, spider, lanyard, seahorse, weathervane, beanbag, stirrup, sleeping bag, scraper, wagon wheel, pendulum, wristlet, armband, diary, stool, carnation, ping-pong ball, shawl, water gun, wristband, raincoat, poncho, hinge, neckerchief, eyepatch, life jacket, boiled egg, curling iron, checkbook, propeller, tiara, safety pin, quilt, cornet, puncher, hornet, grocery bag, first-aid kit, milkshake, atomizer, ottoman, sweatband.

\para{U2.} crib, cape, cube, hatbox, koala, nightshirt, vat, apron, hammock, roller skate, candy bar, perfume, chandelier, martini, futon, heater, polo shirt, dress, dolphin, bath mat, eclair, trampoline, saddlebag, walking stick, radiator, television set, shower head, jewel, blanket, desk, coffee table, rollerblade, hairnet, cappuccino, wig, omelet, coatrack, visor, camel, spice rack, piano, blouse, microwave oven, jumpsuit, beeper, ice skate, dining table, casserole, veil, wardrobe, clip, wet suit, quesadilla, sofa, fire alarm, cigarette case, coil, cardigan, christmas tree, tapestry, brass plaque, parachute, satchel, ironing board.

\para{U3.} sink, tarp, cabinet, cougar, kitchen table, water cooler, curtain, bedpan, bathtub, alligator, stove, garden hose, polar bear, vacuum cleaner, rat, bench, air conditioner, blackboard, blazer, lawn mower, cupboard, sweatshirt, telephoto lens, bread-bin, lamb-chop, kennel, table-tennis table, ice maker, shark, bunk bed, dishwasher, manatee, shower cap, dishrag, beret, cockroach, parka, hotplate, headboard, snake, gravestone, gargoyle, bulletin board, manger, corset, tablecloth, halter top, unicycle, generator, bass horn, scarecrow, gorilla, grill, giant panda, gazelle, pool table, grizzly, water faucet, bedspread, refrigerator, subwoofer, vending machine, egg yolk, harmonium.
\normalsize

\end{document}